%% file: nsp.tex
\documentclass{article}
\PassOptionsToPackage{table}{xcolor}
\usepackage{nsp_arxiv,times}
\usepackage{amsmath,amssymb,amsfonts}
\usepackage{booktabs}
\usepackage{multirow}
\usepackage{graphicx}
\usepackage{hyperref}
\hypersetup{
  hidelinks,
  pdftitle={Neural State Prediction: Obstructing Shortcut Learning in EEG Foundation Models},
  pdfauthor={Kieren Yu, Ziyang Liu, Chang Huang, Jintai Chen, Kaishun Wu}
}
\usepackage{url}
\usepackage{caption}
\usepackage{algorithm}
\usepackage{algpseudocode}
\usepackage{float}
\usepackage{wrapfig}
\usepackage{tikz}
\usetikzlibrary{arrows.meta}

\input{math_commands.tex}

\title{Neural State Prediction: Obstructing Shortcut Learning in EEG Foundation Models}

\author{Kieren Yu\textsuperscript{1}, Ziyang Liu\textsuperscript{1}, Chang Huang\textsuperscript{1}, Jintai Chen\textsuperscript{1}, Kaishun Wu\textsuperscript{1}\\
\normalfont\small\textsuperscript{1}Hong Kong University of Science and Technology (Guangzhou)}

\begin{document}
\raggedbottom

\maketitle
 
\begin{abstract}
  
EEG foundation models increasingly use masked prediction to learn from unlabeled recordings, but optimizing this objective does not ensure transferable neural representations.
A central challenge is that stable positional cues and local correlations can make masked regions predictable without integrating distributed neural context.
To reduce this reliance on low-information prediction paths, we introduce \textbf{Neural State Prediction (NSP)}, a latent-predictive framework that constrains both the prediction target and the available context.
NSP uses a Target Encoder updated by an exponential moving average (EMA) to define latent supervision.
\textbf{Identity residualization} removes additive effects associated with channel identity and relative time from the targets, while \textbf{topology-separated context} excludes their immediate spatial and temporal neighborhood from the visible input.
We pretrain NSP on 2.2 million EEG segments from TUEG and evaluate it across 30 downstream datasets spanning clinical diagnosis, sleep staging, emotion recognition, motor imagery, event-related potentials, cognitive-state decoding, and language retrieval.
Under full-parameter multi-task fine-tuning on EEG-FM-Bench, NSP achieves \textbf{63.94 macro balanced accuracy} across 14 datasets, exceeding the strongest evaluated baseline by \textbf{2.35 percentage points}.
Controlled component ablations assess the contribution of each mechanism, while matched context controls and held-out interventions characterize the role of context geometry, signal content, and positional information.
Jointly designing latent targets and their context offers a promising direction for EEG foundation models that learn from distributed signal structure.
\end{abstract}

\section{Introduction}
\label{sec:intro}

\begin{wrapfigure}[15]{r}{0.35\textwidth}
\vspace{-10pt}
\centering
\includegraphics[width=\linewidth]{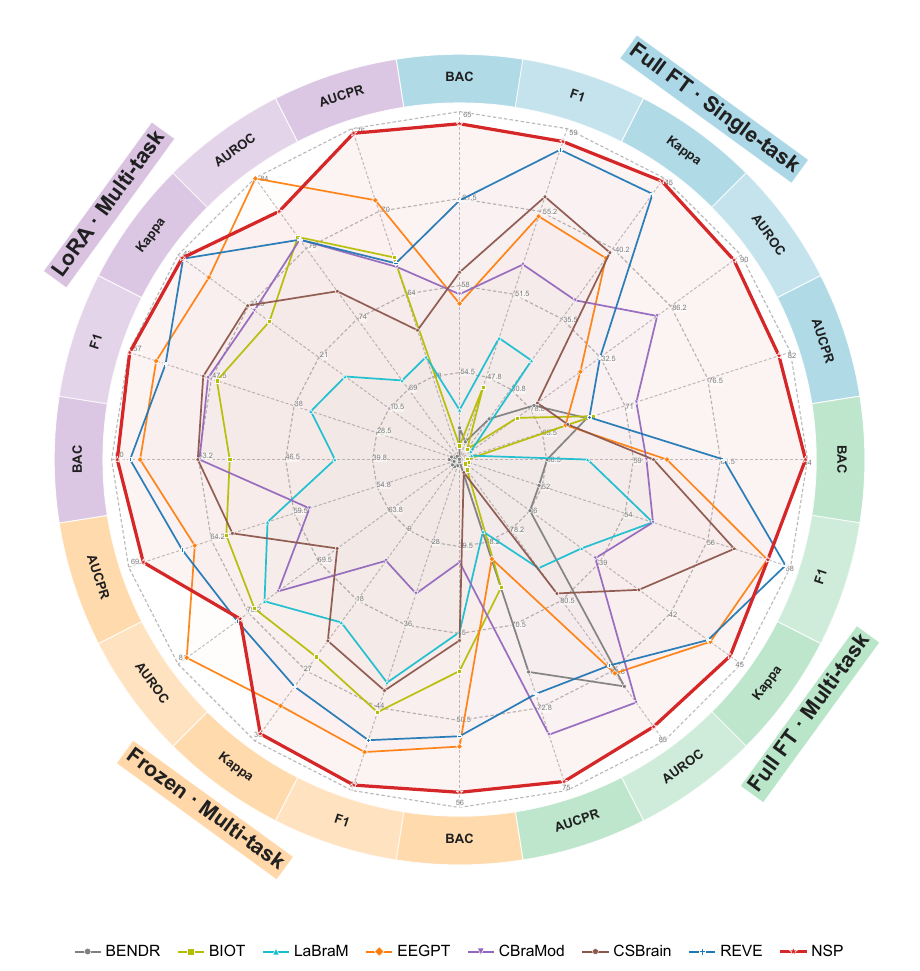}
\vspace{-6pt}
\captionsetup{font=scriptsize,skip=2pt}
\caption{\textbf{EEG-FM-Bench macro summary.}}
\label{fig:eegfmbench-macro-radar}
\vspace{-7pt}
\end{wrapfigure}

Electroencephalography (EEG) records millisecond-scale neural dynamics non-invasively and supports applications ranging from clinical monitoring to emotion recognition and brain--computer interfaces (BCIs). Many EEG systems are still developed for individual tasks and datasets from limited labeled cohorts~\citep{eegnet, eegconformer, st_transformer}. These specialized models remain useful, but often transfer imperfectly across acquisition systems, subject populations, and prediction targets. Inspired by self-supervised learning in language~\citep{brown2020language, devlin2018bert} and vision~\citep{dosovitskiy2020image, he2022masked}, EEG foundation models (EEG-FMs) instead pretrain reusable encoders on broad unlabeled corpora~\citep{bendr, labram, eegpt, cbraMod, csbrain, biot, reve, laya}. EEG-FMs have improved cross-dataset transfer, advancing general-purpose EEG representations.

\WFclear

\begin{figure}[t]
\centering
\includegraphics[width=0.85\textwidth]{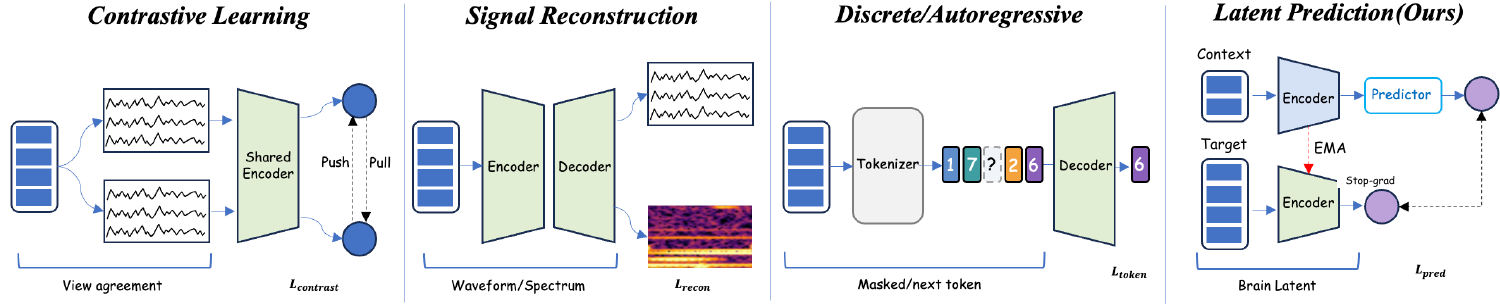}
\caption{\textbf{Comparison of EEG foundation-model pretraining paradigms.} The panels contrast view-level agreement, signal reconstruction, discrete or autoregressive token prediction, and NSP's Target--Context latent prediction.}
\label{fig:paradigms}
\end{figure}

Recent EEG-FMs use four main pretraining paradigms (Figure~\ref{fig:paradigms}): \textbf{(1)}~\textit{Contrastive learning.} These methods learn representations by bringing related EEG views closer while separating unrelated views. BIOT~\citep{biot} aligns representations across biosignal segments and modalities, whereas BENDR~\citep{bendr} contrasts masked contextual representations with their corresponding latent targets. Their success nevertheless depends on how positive views, negatives, and augmentations define invariance, and view-level agreement does not directly specify which missing neural content should be recovered. \textbf{(2)}~\textit{Signal reconstruction.} Reconstruction objectives mask part of an EEG recording and recover the omitted waveform, spectrum, or patch. EEGPT~\citep{eegpt} and REVE~\citep{reve} reconstruct masked waveforms, CBraMod~\citep{cbraMod} reconstructs EEG patches with criss-cross spatial--temporal attention, and CSBrain~\citep{csbrain} reconstructs spectrogram representations. Such objectives preserve signal detail, but may devote capacity to amplitude, noise, and other input-level variation that is not useful downstream. \textbf{(3)}~\textit{Discrete or autoregressive prediction.} Discrete approaches encode EEG as codebook tokens, whereas autoregressive approaches predict the next token or patch from preceding context. LaBraM~\citep{labram}, DeWave~\citep{duan2023dewave}, and CodeBrain~\citep{ma2025} learn or use discrete EEG tokens, while ECHO~\citep{liu2026echo}, THD-BAR~\citep{yang2025}, and KAST-BAR~\citep{wang2026kastbar} predict sequential tokens or future patches. Discretization can provide compact supervision, although its quality is bounded by the tokenizer or codebook. \textbf{(4)}~\textit{Latent prediction.} These methods predict learned representations of masked regions, typically through joint-embedding predictive architectures (JEPAs) with a context and a target encoder. EEG2Rep~\citep{foumani2024eeg2rep} combines latent self-prediction with semantic subsequence preserving (SSP) masking. Brain-JEPA~\citep{dong2024brainjepa} models latent brain dynamics, S-JEPA~\citep{guetschel2024sjepa} targets cross-dataset EEG transfer, and Laya~\citep{laya}, STST-JEPA~\citep{segal2026ststjepa}, and EEG-JEPA~\citep{li2026eegjepa} develop learned-target prediction for EEG with different target construction and masking choices. PATCHCODE~\citep{yu2026patchcode} adds discrete-code supervision from a frozen tokenizer to continuous patch-level latent prediction. These studies make latent prediction a promising alternative to reconstructing every detail of noisy electrophysiological inputs.

Despite these advances, dependence on easily predictable signal structure can hinder the development of EEG representations that generalize across tasks and recording conditions. Pretraining objectives reward agreement with a chosen target, but do not by themselves distinguish transferable neural structure from stable positional cues or local correlations. Motivated by the shared source mixtures that volume conduction induces at neighboring scalp electrodes~\citep{nunez2006electric}, we examine local interpolation alongside prediction from channel identity, electrode coordinates, and relative time. A masked objective may be reduced through either path without requiring broad neural context. Latent prediction avoids direct waveform reconstruction, but it does not automatically remove these low-information solutions: existing EEG JEPAs define learned targets and masked context, yet generally do not jointly reduce additive channel- and time-conditioned components in the target and exclude the target's nearest spatial and temporal evidence. The central objective-design question is therefore not only whether to predict a waveform, token, or representation, but also what information remains in the prediction target and what evidence is available to infer it.

To address this mismatch between pretraining success and transferable representation learning, we propose \textbf{Neural State Prediction (NSP)}, a latent-predictive framework that jointly constrains target content and available context. An EMA-updated Target Encoder first provides stable latent supervision for masked spatiotemporal regions. Identity residualization then removes the additive components associated with channel identity and relative time from each target, reducing position-conditioned predictability. In parallel, topology-separated context withholds the target's nearest-neighbor channels and adjacent time patches from the visible EEG content, requiring the Context Encoder to integrate more distant evidence. The residualized target specifies \emph{what} is predicted, whereas the topology-separated context specifies \emph{from what evidence} the prediction is formed. We pretrain NSP on 2.2~million EEG segments from TUEG~\citep{obeid2016tueg} and evaluate transfer across 30 downstream datasets spanning clinical diagnosis, motor imagery, emotion, cognitive workload, event-related potentials, language, and other brain-state decoding tasks. The evaluation comprises 16 selected NeuralBench-EEG-Core v1.0 datasets~\citep{neuralbench} (NeuralBench for short), after removing dataset--task pairs also covered by EEG-FM-Bench, together with the 14-dataset EEG-FM-Bench protocol~\citep{eegfmbench}. Figure~\ref{fig:eegfmbench-macro-radar} compares representative EEG-FMs from all four pretraining paradigms under EEG-FM-Bench; NSP achieves 63.94 macro balanced accuracy under full-parameter multi-task fine-tuning, outperforming the strongest evaluated baseline by 2.35 percentage points. Matched component and context-control ablations examine how the proposed mechanisms affect transfer, and held-out signal-content interventions and identity probes characterize sensitivity to EEG content and the positional information retained by component variants.

Our contributions are summarized as follows:
\begin{itemize}
\item We identify \textbf{position-conditioned prediction} and \textbf{local interpolation} as two low-information paths in masked EEG latent prediction, motivating joint design of targets and visible context.
\item We propose \textbf{Neural State Prediction (NSP)}, a latent-predictive \textbf{EEG-FM pretraining framework}. Its EMA Target Encoder supplies supervision, while \textbf{identity residualization} and \textbf{topology-separated context} jointly constrain target content and visible EEG evidence.
\item We evaluate NSP across \textbf{30 downstream datasets}. Under EEG-FM-Bench multi-task full fine-tuning, it achieves \textbf{63.94 macro BAC}, 2.35 percentage points above the strongest evaluated baseline. Matched ablations, signal-content interventions, and identity probes assess the components and learned representations.
\end{itemize}

\section{Neural State Prediction}
\label{sec:method}

\begin{figure}[t]
\centering
\includegraphics[width=0.98\textwidth]{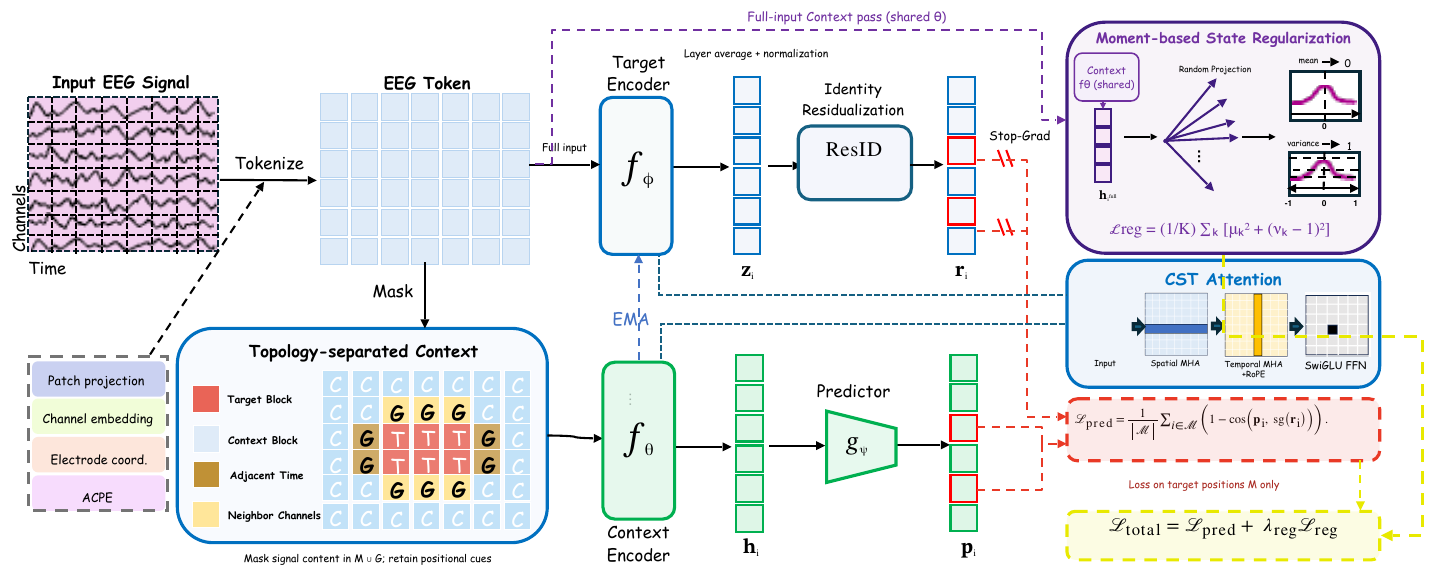}
\caption{\textbf{Neural State Prediction architecture.} Residualized, stop-gradient Target states supervise the masked Context branch through its predictor. A separate complete-input pass through the shared Context Encoder supplies moment-based state regularization. The Target Encoder follows the Context Encoder by EMA; prediction is scored only at target positions.}
\label{fig:method}
\end{figure}

Neural State Prediction (NSP) learns an EEG representation by predicting the latent states of masked spatiotemporal regions from the remaining recording. As summarized in Figure~\ref{fig:method}, a Target Encoder operating on the complete token grid defines the latent state to be predicted. Identity residualization removes batch-estimated channel and relative-time components from that state, while topology-separated context controls the evidence available to the Context Encoder. These two constructions have complementary roles: the residualized target specifies \emph{what} is predicted, and the separated context specifies \emph{from what evidence} it is inferred.

\subsection{Overview and representation encoder}
\label{sec:method-overview}

Let $\mathbf{X}\in\mathbb{R}^{B\times C\times T}$ be a batch of EEG segments, where $B$ is the batch size, $C$ is the number of channels, and $T$ is the number of samples per segment. Each channel is partitioned into $P$ temporal patches, giving the position set $\Omega=[P]\times[C]$; each position carries a $d$-dimensional state. For example $b$, $\mathcal{M}_b\subset\Omega$ denotes the target region and $\mathcal{G}_b\subset\Omega$ its spatial--temporal guard. Only EEG content outside $\mathcal{M}_b\cup\mathcal{G}_b$ is visible to the Context Encoder.

The Target and Context Encoders share the same architecture but have separate parameters, denoted by $\phi$ and $\theta$ respectively. Their tokenizers combine three sources of information: a projection of the EEG patch, a learned channel embedding, and adaptive channel position encoding (ACPE) from electrode coordinates. The patch projection is the signal-dependent term. Masking replaces this term before channel and coordinate information is added, so a masked position remains represented in the grid without exposing its EEG content. This ordering keeps masked positions addressable by channel and coordinate while withholding their EEG samples.

Both branches use a Cascaded Spatial--Temporal Transformer (CST). Each block applies spatial multi-head attention (MHA) across channels, then temporal attention across patches within each channel, followed by a feed-forward network (FFN). Temporal attention uses rotary position encoding (RoPE)~\citep{su2024roformer}. We write $\mathcal{S}=\{\mathcal{S}_b\}_{b=1}^B$ for a batch of mask sets and $\mathbf{H}_\eta^{(\ell)}(\mathbf{X};\mathcal{S})$ for the state after block $\ell$ of encoder $\eta\in\{\theta,\phi\}$ when EEG content at $\mathcal{S}_b$ is masked in example $b$. The final block is denoted by $N_{\mathrm{blk}}$. The Target branch uses empty mask sets, whereas the Context branch uses $\mathcal{S}_b=\mathcal{M}_b\cup\mathcal{G}_b$.

\subsection{Target construction: identity-residualized latent states}
\label{sec:method-target}

The Target Encoder sees the complete recording. To make its supervision less dependent on any single block, we average a set $\mathcal{I}_T$ of late Target states, apply encoder layer normalization (LN), and standardize each sample--channel slice to zero mean and unit variance over its temporal and feature dimensions:
\begin{equation}
\mathbf{z}
=\operatorname{Std}_{P\times d}\!\left[
  \operatorname{LN}\!\left(
    \frac{1}{|\mathcal{I}_T|}\sum_{\ell\in\mathcal{I}_T}
    \mathbf{H}^{(\ell)}_\phi(\mathbf{X};\varnothing)
  \right)
\right].
\label{eq:target-aggregation}
\end{equation}
This gives a target latent $\mathbf{z}_{b,p,c}\in\mathbb{R}^d$ for each position. Because this aggregation retains the full patch--channel grid, NSP predicts a state for each masked position rather than a pooled recording-level vector.

Identity residualization removes additive channel and relative-time effects estimated from the current batch. Let $\boldsymbol{\mu}$ and $\boldsymbol{\sigma}^2$ be the per-feature mean and variance of $\mathbf{z}$ over all batch, patch, and channel positions. We first place each feature on a common scale,
\begin{equation}
\mathbf{u}_{b,p,c}
=\frac{\mathbf{z}_{b,p,c}-\boldsymbol{\mu}}
{\sqrt{\boldsymbol{\sigma}^2+\varepsilon}},
\label{eq:target-standardization}
\end{equation}
where $\varepsilon$ is a numerical stabilizer. On this common scale, we estimate the additive component associated with channel identity and relative patch time:
\begin{equation}
\boldsymbol{\mu}^{\mathrm{ch}}_c
=\frac{1}{BP}\sum_{b,p}\mathbf{u}_{b,p,c},
\qquad
\boldsymbol{\mu}^{\mathrm{time}}_p
=\frac{1}{BC}\sum_{b,c}\mathbf{u}_{b,p,c}.
\label{eq:identity-means}
\end{equation}
The prediction target is the remaining state,
\begin{equation}
\mathbf{r}_{b,p,c}
=\mathbf{u}_{b,p,c}
-\boldsymbol{\mu}^{\mathrm{ch}}_c
-\boldsymbol{\mu}^{\mathrm{time}}_p.
\label{eq:identity-residual}
\end{equation}
This operation is indexed by channel and relative patch time; it does not regress directly on continuous electrode coordinates. We stop gradients through $\mathbf{r}$ before it is used as supervision. Standardization and residualization play different roles: standardization places latent dimensions on comparable scales across the batch, while the two marginal means remove the additive component shared by a channel or a relative time index, leaving residual variation not explained by those two marginals. Electrode coordinates remain available to the encoders; they are not themselves fitted or subtracted by this operation.

The Target Encoder receives no optimizer update. Following self-distillation and joint-embedding predictive learning~\citep{grill2020bootstrap,ijepa}, after each Context Encoder update its parameters follow an exponential moving average:
\begin{equation}
\phi_{s+1}
=\tau_s\phi_s+(1-\tau_s)\theta_{s+1},
\label{eq:ema-update}
\end{equation}
where $\tau_s$ is the scheduled EMA coefficient. EMA provides a slowly varying target while gradients update the Context branch and predictor.

\subsection{Context construction: topology-separated evidence}
\label{sec:method-context}

For each example, target positions $\mathcal{M}_b$ are sampled as connected spatiotemporal blocks rather than as independent tokens. This keeps the prediction target locally coherent. To prevent the Context Encoder from recovering a target by interpolation from its immediate surroundings, we expand every target block with a guard. Let $\mathcal{N}_t$ denote temporal neighborhood expansion and $\mathcal{N}_s$ spatial expansion according to electrode distance. The guard is
\begin{equation}
\mathcal{G}_b
=\left(
\mathcal{N}_t(\mathcal{M}_b)
\cup
\mathcal{N}_s(\mathcal{M}_b)
\right)\setminus\mathcal{M}_b.
\label{eq:guard-set}
\end{equation}
The prediction loss is evaluated only on $\mathcal{M}_b$; $\mathcal{G}_b$ changes the available evidence but adds no extra targets.

Let $\mathbf{q}^\theta_{b,p,c}$ denote the signal-dependent patch embedding in the Context tokenizer and $\mathbf{m}^\theta$ a learned mask embedding. With $\mathcal{S}_b=\mathcal{M}_b\cup\mathcal{G}_b$, masking is simply
\begin{equation}
\widetilde{\mathbf{q}}^{\theta}_{b,p,c}
=
\begin{cases}
\mathbf{m}^{\theta},
& (p,c)\in\mathcal{S}_b,\\
\mathbf{q}^{\theta}_{b,p,c},
& \text{otherwise}.
\end{cases}
\label{eq:masked-context}
\end{equation}
Channel and coordinate embeddings are added after this replacement, so the encoder retains position identity but not its EEG content. The final Context state is $\mathbf{h}=\mathbf{H}_\theta^{(N_{\mathrm{blk}})}(\mathbf{X};\mathcal{S})$. A shared position-wise predictor gives $\mathbf p=g_\psi(\mathbf h)$ in the Target latent space. Writing $\operatorname{nrm}$ for last-dimension $\ell_2$ normalization, the loss compares
\begin{equation}
\widehat{\mathbf{r}}_{b,p,c}
=\operatorname{nrm}\!\left(\mathbf p_{b,p,c}\right),
\qquad
\overline{\mathbf{r}}_{b,p,c}
=\operatorname{nrm}\!\left(\operatorname{sg}(\mathbf{r}_{b,p,c})\right),
\label{eq:prediction-pair}
\end{equation}
where $\operatorname{sg}$ denotes stop-gradient. The same predictor is applied at every grid position.

\subsection{Masked latent prediction}

The prediction term is the mean cosine distance between the normalized prediction and the stopped, normalized Target state:
\begin{equation}
\mathcal{L}_{\mathrm{pred}}
=\frac{1}{\sum_b|\mathcal{M}_b|}
\sum_b\sum_{(p,c)\in\mathcal{M}_b}
\left(1-\widehat{\mathbf{r}}_{b,p,c}^{\top}
\overline{\mathbf{r}}_{b,p,c}\right).
\label{eq:prediction-loss}
\end{equation}
Because the sum ranges only over $\mathcal{M}_b$, the objective asks the Context Encoder to infer target states from EEG evidence outside both the target and its guard. Each comparison remains position matched: the prediction at $(p,c)$ is paired with the residualized Target state from that same token position.

\subsection{Training objective and parameter updates}

We accompany latent prediction with a random-projection moment regularizer inspired by SIGReg~\citep{balestriero2025lejepa}. It uses a separate, unmasked Context pass $\mathbf{h}^{\mathrm{full}}=\mathbf{H}_\theta^{(N_{\mathrm{blk}})}(\mathbf{X};\varnothing)$. We flatten its token states and project them onto $K$ random unit directions $\mathbf{v}_k$. For $s_{i,k}=(\mathbf{h}_i^{\mathrm{full}})^\top\mathbf{v}_k$, let $\mu_k$ and $\nu_k$ denote the empirical mean and variance over token index $i$. Our moment-matching variant encourages zero mean and unit variance:
\begin{equation}
\mathcal{L}_{\mathrm{reg}}
=\frac{1}{K}\sum_{k=1}^K
\left[\mu_k^2+(\nu_k-1)^2\right].
\label{eq:sigreg}
\end{equation}
Unlike the distributional goodness-of-fit test used in SIGReg, this variant matches only the first two projected moments; it does not enforce a Gaussian distribution. It acts on full-input Context features, not Target latents, decoupling regularization from the sampled mask geometry.

The complete objective is
\begin{equation}
\mathcal{L}_{\mathrm{total}}
=\mathcal{L}_{\mathrm{pred}}
+\lambda_{\mathrm{reg}}\mathcal{L}_{\mathrm{reg}}.
\label{eq:total-objective}
\end{equation}
Gradient updates optimize $\theta$ and $\psi$; the subsequent EMA update in Eq.~(\ref{eq:ema-update}) updates $\phi$.

\section{Experiments}
\label{sec:experiments}

\subsection{Experimental setup}
\label{sec:protocol}

\textbf{Pretraining data and preprocessing.} We pretrain NSP on the Temple University Hospital EEG Corpus (TUEG), a heterogeneous clinical archive collected from 2002 to 2017~\citep{obeid2016tueg,cbraMod}. It contains 69,652 EEG files from 14,987 subjects and 26,846 sessions, recorded with variable montages and sampling rates. We map recordings to 19 standard 10--20 channels, resample to 200~Hz, apply 0.5--40~Hz band-pass and 50/60~Hz notch filtering, use common-average referencing, clip at $\pm500~\mu$V, and perform recording-level z-normalization. Non-overlapping 6-s windows undergo quality checks after normalization. Downstream preprocessing is benchmark specific: NeuralBench-EEG-Core follows each task's released YAML configuration, while EEG-FM-Bench follows its released event/channel selection, filtering, windowing, montage, and unit-conversion pipeline; NSP uses the task-specific input adapters; the 19-channel, 200-Hz setting specifies pretraining rather than a universal downstream input requirement.
 
\input{tables/tab_datasets}

\textbf{Downstream tasks and metrics.} We use NeuralBench-EEG-Core v1.0~\citep{neuralbench} and EEG-FM-Bench~\citep{eegfmbench} (Table~\ref{tab:datasets}). After removing dataset--task pairs shared with EEG-FM-Bench, NeuralBench contributes 16 datasets: four clinical, four BCI/sensorimotor, four event-related, and four cognitive/language tasks. Fourteen classification tasks use balanced accuracy (BAC); the ZuCo and Nieuwland2018 retrieval tasks use per-subject Top-5 accuracy and are reported separately. EEG-FM-Bench contains 14 datasets from 10 paradigms and evaluates single-task and multi-task full fine-tuning, frozen-backbone multi-task transfer, and multi-task low-rank adaptation (LoRA).

\textbf{Baselines and downstream protocols.} The benchmarks use different comparison sets. NeuralBench evaluates BENDR, BIOT, LaBraM, CBraMod, LUNA, and REVE~\citep{bendr,biot,labram,cbraMod,luna2025,reve} with its unified end-to-end recipe; NSP uses seeds 33--35. EEG-FM-Bench compares BENDR, BIOT, LaBraM, EEGPT, CBraMod, CSBrain, and REVE~\citep{bendr,biot,labram,eegpt,cbraMod,csbrain,reve} under its four released protocols~\citep{eegfmbench}. Its 30-epoch AdamW adaptation uses a 3-epoch warmup, batch size 128, gradient clipping at 1.0, weight decay 0.01, and a one-epoch encoder freeze. Multi-task full fine-tuning uses seeds 42--46; other protocols use seeds 42--44. We report dataset means with sample standard deviation (SD) and average dataset means for each macro.

\textbf{Pretraining setting.} The 42.83M-parameter Large model has eight CST blocks, width 512, eight heads, FFN width 2,048, 0.2-s patches, and a predictor with hidden width 768 and output width 512. Mixed-precision distributed data-parallel training runs on eight NVIDIA H100 GPUs with global batch size 384. AdamW uses learning rate $10^{-4}$, weight decay 0.05, gradient clipping at 1.0, 5,000 warmup steps, and cosine decay over 50,000 steps. Component studies hold data, architecture, optimization, and transfer protocol fixed; duration and capacity studies vary only the named factor. Appendix Tables~\ref{tab:app_pretraining_architecture}--\ref{tab:app_pretraining_optimization} provide the complete configuration.

\begin{figure}[H]
\centering
\captionsetup{font=small,skip=3pt}
\includegraphics[width=0.92\textwidth]{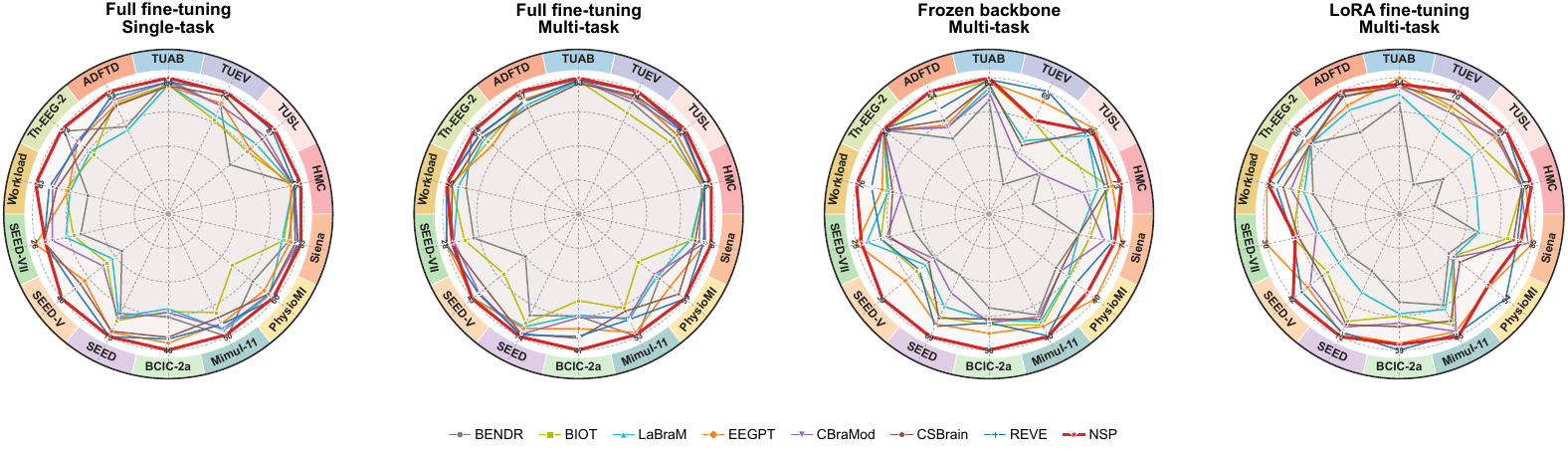}
\caption{\textbf{EEG-FM-Bench radar summary.} Dataset-wise BAC under four adaptation protocols across all 14 datasets; radial positions use per-dataset relative scaling and NSP is highlighted in red.}
\label{fig:eegfmbench-radar}
\end{figure}

\subsection{Performance Comparison with baselines}
\label{sec:benchmark}

\input{tables/tab_full_finetuning}

To compare NSP with established EEG foundation models across domains and adaptation settings, we evaluate it on 16 selected NeuralBench-EEG-Core datasets and the 14-dataset EEG-FM-Bench suite. Table~\ref{tab:full_finetuning} reports the NeuralBench-EEG-Core results under its unified end-to-end protocol, while Figure~\ref{fig:eegfmbench-radar} summarizes single-task and multi-task full fine-tuning, plus multi-task frozen and LoRA transfer on EEG-FM-Bench.

On the 14 NeuralBench classification tasks, NSP reaches 73.14 macro BAC, ahead of LaBraM (70.33), CBraMod (68.04), and REVE (67.30). It has the highest mean on nine tasks and ranks among the top two on twelve, spanning clinical, sensorimotor, event-related, and cognitive settings. Representative wins include Mumtaz2018, Singh2021, Thielen2021, Scherer2015, Wang2017, ERP CORE-N400, and Hinss2023. The margins range from near ties on Singh2021 and Wang2017 to 5.52 percentage points on Scherer2015 and 6.92 on Hinss2023, so the macro advantage aggregates both modest and substantial task-level gains. NSP also reaches $12.11\pm0.39$ and $36.27\pm0.12$ Top-5 accuracy on ZuCo and Nieuwland2018, respectively; these retrieval scores are not pooled with BAC.

EEG-FM-Bench tests NSP under several adaptation protocols. Under full-parameter multi-task fine-tuning, NSP averages 63.94 macro BAC, 2.35 percentage points above the strongest evaluated baseline. Gains are largest on several clinical and BCI datasets, including TUSL, HMC, Siena, and BCIC-2a, while emotion and motor-imagery results are more mixed; frozen and LoRA transfer are likewise less uniform than full fine-tuning. TUEV is a near tie under multi-task adaptation, and results on SEED-VII, PhysioMI, and Mimul-11 vary by task. NeuralBench measures single-task transfer across domains, whereas EEG-FM-Bench measures performance across adaptation strategies. Together, they characterize transfer across both tasks and adaptation settings.

\subsection{Component and input/context dependence analysis}
\label{sec:ablation}

\input{tables/tab_ablation}

\textbf{Target construction and representation stability.} To identify target-side contributions under limited labels, we evaluate frozen single-task transfer with 10\% labels on TUAB, TUEV, SEED, and PhysioMI (Table~\ref{tab:ablation}). Starting from block masking at 45.75 macro BAC, an EMA Target Encoder raises the score to 51.88, and adding identity residualization further improves it to 54.24. These cumulative changes support both stable latent supervision and removal of additive positional effects. Removing state regularization from Full NSP lowers macro BAC from 55.70 to 46.88, indicating that target and context design do not replace representation-stability control.

\textbf{Identity residualization and context exclusion.} To separate their contributions, matched rows fix the EMA target, regularization, architecture, and visible-token budget. Residualization adds 0.62 percentage points with random exclusion and 1.52 with topology-separated exclusion. Topology separation adds 1.06 percentage points without residualization and 1.96 with it. Both components improve the matched means, more so together.

\textbf{Signal-content and identity diagnostics.} To distinguish signal sensitivity from positional information, we zero inputs, cyclically mismatch recordings within a batch, or expand Context masks while holding Targets fixed (Figure~\ref{fig:mechanism-diagnostics}). Intervention cosine is normalized by the target-only-masked Context reference. Separate probes show that channel identity remains strongly decodable from Context states and targets without channel residualization. These component diagnostics characterize sensitivity and residual identity information, not channel invariance.

\subsection{Training duration, capacity, and label efficiency}
\label{sec:scaling}

\begin{figure}[t]
\centering
\captionsetup{font=small,skip=3pt}
\includegraphics[width=0.72\textwidth]{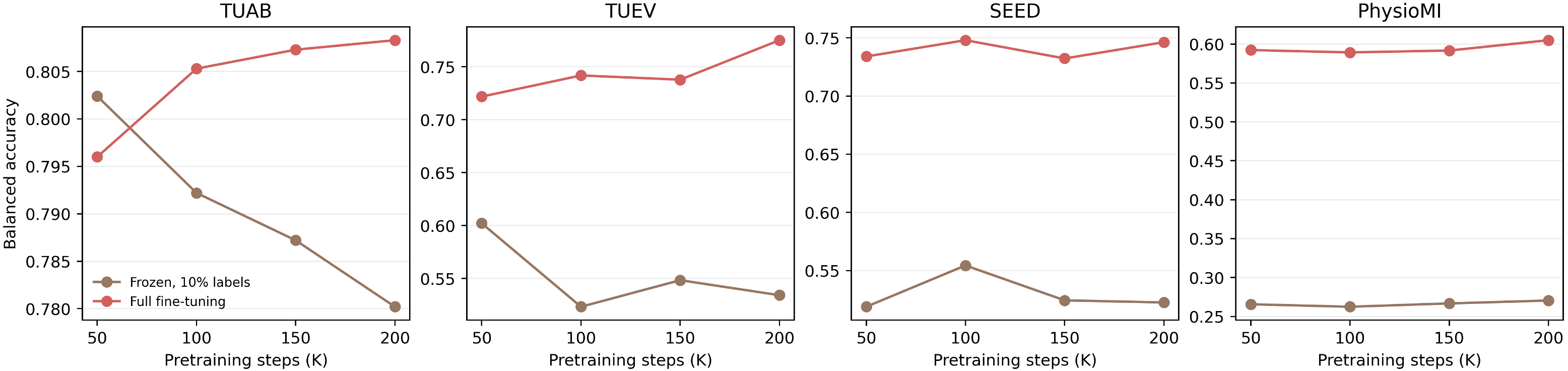}
\caption{\textbf{Task-wise transfer across pretraining duration.} BAC (0--1 scale) across checkpoints under frozen single-task transfer (10\% labels) and single-task full fine-tuning.}
\label{fig:duration}
\end{figure}

To determine how duration interacts with adaptation, we evaluate frozen and full-parameter single-task transfer at multiple checkpoints on four tasks (Figure~\ref{fig:duration}). Frozen transfer is generally strongest at the shorter budget, whereas full fine-tuning is maintained or improves on several tasks. We also pretrain Tiny and Medium NSP variants and compare them with Large under single-task full fine-tuning (Figure~\ref{fig:model-scale}). Macro BAC increases with capacity, but the preferred scale varies by task; this comparison is separate from the fixed-architecture component ablation.

\begin{figure}[t]
\centering
\captionsetup{font=small,skip=3pt}
\includegraphics[width=0.72\textwidth]{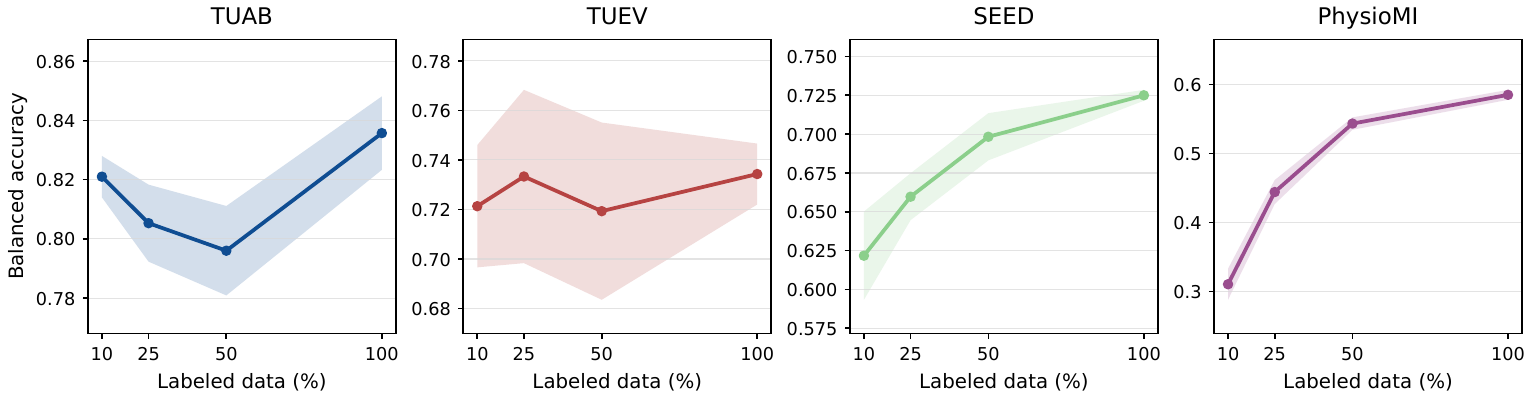}
\caption{\textbf{Label efficiency across downstream tasks.} BAC (0--1 scale) versus labeled-data fraction on TUAB, TUEV, SEED, and PhysioMI; bands show variability across evaluations.}
\label{fig:label-efficiency}
\end{figure}

To test NSP with limited downstream labels, we fine-tune it with 10\%, 25\%, 50\%, and 100\% of the available labels (Figure~\ref{fig:label-efficiency}). TUAB and TUEV change by less than 1.5 percentage points between 10\% and 100\% labels, whereas SEED improves from 62.17 to 72.50 and PhysioMI from 31.03 to 58.50. The TUAB and TUEV curves remain comparatively flat across label fractions.

\subsection{Representation geometry and task-conditioned spatial evidence}

\begin{figure}[H]
\centering
\captionsetup{font=small,skip=3pt}
\includegraphics[width=0.78\textwidth]{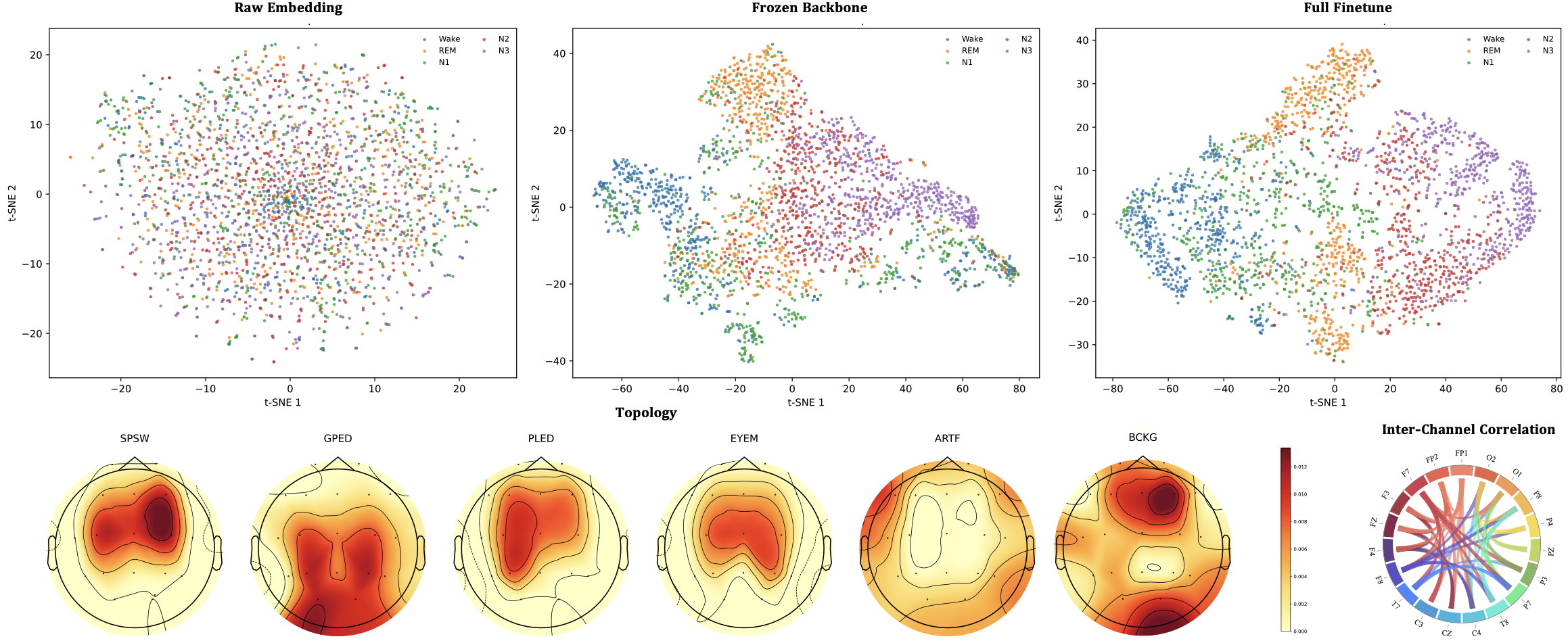}
\caption{\textbf{Representation geometry and spatial evidence.} Top: raw, frozen, and fine-tuned HMC sleep-stage representations visualized by t-SNE. Bottom: TUEV event-class Grad-CAM maps and interchannel correlations. The rows depict different datasets.}
\label{fig:representation}
\end{figure}

To examine geometry and spatial evidence, we visualize HMC embeddings with t-SNE~\citep{vandermaaten2008tsne} and TUEV attribution with Grad-CAM~\citep{selvaraju2017gradcam} (Figure~\ref{fig:representation}). HMC representations show more sleep-stage structure after pretraining and adaptation, although classes still overlap. TUEV attribution patterns differ across SPSW, GPED, PLED, EYEM, ARTF, and BCKG; chords depict selected interchannel correlations.

\section{Conclusion}
\label{sec:conclusion}

NSP couples EMA-based latent prediction with identity-residualized targets and topology-separated context to reduce additive positional effects and encourage distributed inference. Transfer across 30 datasets and component analyses support jointly designing targets and context for EEG representation learning. Future work could adapt exclusions to signal-dependent correlations. Despite broad benchmark coverage, responsible use requires prospective validation across populations and acquisition settings, together with safeguards for neural-data privacy.

\clearpage
\subsection*{AI Use Disclosure}

AI-assisted tools, including OpenAI Codex, were used to support language editing, literature search, code development and debugging, experiment scripting and monitoring, and figure preparation. The authors remain responsible for validating the code, experimental results, and references, and for all scientific claims, interpretations, and final content of this paper.

\bibliography{references}
\bibliographystyle{nsp_references}

\newpage
\appendix
\input{appendix}

\end{document}

%% file: math_commands.tex
\usepackage{amsmath,amsfonts,bm}

\def\Secref#1{Section~\ref{#1}}

\def\eqref#1{equation~\ref{#1}}

\def\1{\bm{1}}

\DeclareMathAlphabet{\mathsfit}{\encodingdefault}{\sfdefault}{m}{sl}
\SetMathAlphabet{\mathsfit}{bold}{\encodingdefault}{\sfdefault}{bx}{n}



%% file: tables/tab_datasets.tex
\begin{table*}[!t]
\centering
\caption{Overview of 16 selected NeuralBench-EEG-Core datasets and 14 EEG-FM-Bench datasets~\citep{neuralbench,eegfmbench}. Dataset--task pairs shared by the two benchmarks are excluded from the NeuralBench-EEG-Core selection. Sampling follows each benchmark's dataset-specific preprocessing; NSP uses 200\,Hz for EEG-FM-Bench. Benchmark details and complete results are provided in Appendix~\ref{app:complete-results}.}
\label{tab:datasets}
\vskip 0.05in
\begingroup
\fontsize{7.5}{8.2}\selectfont
\setlength{\tabcolsep}{3.2pt}
\renewcommand{\arraystretch}{0.94}
\resizebox{\textwidth}{!}{%
\begin{tabular}{@{}llrrrrr@{\hspace{8pt}}llrrrrr@{}}
\toprule
\textbf{Task} & \textbf{Dataset} & \textbf{Channels} & \textbf{Window (s)} & \textbf{Samples} & \textbf{Subjects} & \textbf{Classes} & \textbf{Task} & \textbf{Dataset} & \textbf{Channels} & \textbf{Window (s)} & \textbf{Samples} & \textbf{Subjects} & \textbf{Classes} \\
\midrule
Alzheimer's identification & ADFTD & 19 & 10 & 7,013 & 88 & 3 & Motor imagery & Mimul-11 & 60 & 5 & 41,347 & 11 & 3 \\
Depression diagnosis & Mumtaz2018 & 19 & 5 & 14,683 & 64 & 2 &  & PhysioMI & 64 & 4 & 9,747 & 109 & 4 \\
Parkinson's diagnosis & Singh2021 & 60 & 5 & 39,543 & 129 & 2 &  & BCIC-2a & 22 & 4 & 5,088 & 9 & 4 \\
Schizophrenia diagnosis & Albrecht2019 & 62 & 5 & 24,928 & 77 & 2 &  & Scherer2015 & 30 & 4 & 3,550 & 9 & 5 \\
Abnormal classification & TUAB & 23 & 30 & 272,320 & 2,383 & 2 & Motor execution & Srisrisawang2024 & 60 & 3 & 19,157 & 20 & 16 \\
Anomalous event detection & TUEV & 21 & 5 & 113,353 & 370 & 6 & Visual target detection & Things-EEG-2 & 63 & 5 & 41,570 & 10 & 2 \\
Slowing event classification & TUSL & 21/22 & 10 & 290 & 28 & 3 & c-VEP visual coding & Thielen2021 & 8 & 4 & 45,000 & 30 & 20 \\
Seizure detection & Siena & 29 & 10 & 50,830 & 14 & 2 & SSVEP & Wang2017 & 64 & 4 & 8,160 & 34 & 40 \\
 & CHB-MIT & 24 & 5 & 707,683 & 23 & 2 & Error-related negativity & ERP CORE-ERN & 30 & 1 & 15,972 & 40 & 2 \\
Sleep staging & HMC & 4 & 30 & 136,925 & 151 & 5 & Face-perception response & ERP CORE-N170 & 30 & 1 & 12,800 & 40 & 2 \\
Emotion recognition & SEED & 60 & 10 & 37,890 & 15 & 3 & Semantic-violation response & ERP CORE-N400 & 30 & 1 & 4,800 & 40 & 2 \\
 & SEED-V & 60 & 10 & 12,318 & 16 & 5 & Auditory P300 & Schreuder2010 & 60 & 1 & 253,800 & 21 & 2 \\
 & SEED-VII & 60 & 15 & 19,420 & 20 & 7 & Mental arithmetic & Zyma2019 & 20 & 5 & 1,659 & 35 & 2 \\
Mental workload/stress & Workload & 19 & 10 & 2,134 & 36 & 2 & Sentence retrieval & ZuCo & 128 & 3 & 25,819 & 12 & retr. \\
 & Hinss2023 & 63 & 5 & 15,399 & 29 & 3 & Word retrieval & Nieuwland2018 & 74 & 3 & 529,730 & 222 & retr. \\
\bottomrule
\end{tabular}
}
\endgroup
\end{table*}

%% file: tables/tab_full_finetuning.tex
\begin{table}[!t]
\centering
\caption{\textbf{Primary-metric performance on 16 selected NeuralBench-EEG-Core datasets~\citep{neuralbench}.} BAC (\%) is shown for 14 classification tasks and Top-5 accuracy (\%) for two retrieval tasks. Dataset-level values are mean $\pm$ sample SD; the macro row averages the 14 BAC means. Red and blue mark the best and second-best mean for each task and for the macro row.}
\label{tab:full_finetuning}
\vskip 0.04in
\fontsize{7.15}{8.0}\selectfont
\setlength{\tabcolsep}{1.50pt}
\renewcommand{\arraystretch}{1.02}
\begin{tabular}{@{}llrrrrrrr@{}}
\toprule
\textbf{Dataset} & \textbf{Metric} & \textbf{BENDR} & \textbf{BIOT} & \textbf{LaBraM} & \textbf{CBraMod} & \textbf{LUNA} & \textbf{REVE} & \textbf{NSP} \\
\midrule
\multicolumn{9}{c}{\cellcolor{blue!8}\textit{Clinical and neurological disease}} \\
Mumtaz2018 & BAC & 86.22$\pm$00.47 & 79.11$\pm$03.39 & 85.61$\pm$03.43 & 86.04$\pm$02.48 & 85.42$\pm$00.98 & \textcolor{blue}{\textbf{87.33}}$\pm$01.44 & \textcolor{red}{\textbf{90.43}}$\pm$06.40 \\
Singh2021 & BAC & 65.82$\pm$05.62 & 74.18$\pm$04.16 & \textcolor{blue}{\textbf{76.64}}$\pm$04.06 & 74.37$\pm$02.38 & 64.95$\pm$00.49 & 72.58$\pm$01.72 & \textcolor{red}{\textbf{76.67}}$\pm$01.70 \\
Albrecht2019 & BAC & 57.17$\pm$00.87 & 56.18$\pm$01.50 & 63.62$\pm$01.93 & 57.05$\pm$03.12 & \textcolor{blue}{\textbf{66.75}}$\pm$02.18 & 49.21$\pm$10.36 & \textcolor{red}{\textbf{68.43}}$\pm$02.71 \\
CHB-MIT & BAC & 78.97$\pm$01.04 & 73.58$\pm$05.20 & \textcolor{blue}{\textbf{82.09}}$\pm$00.75 & \textcolor{red}{\textbf{82.89}}$\pm$01.58 & 70.46$\pm$02.44 & 76.51$\pm$03.73 & 81.10$\pm$01.39 \\
\midrule
\multicolumn{9}{c}{\cellcolor{green!8}\textit{BCI and sensorimotor decoding}} \\
Thielen2021 & BAC & 4.96$\pm$00.24 & 4.80$\pm$00.25 & 86.39$\pm$00.74 & 86.37$\pm$01.33 & \textcolor{blue}{\textbf{87.06}}$\pm$00.28 & 80.51$\pm$01.81 & \textcolor{red}{\textbf{88.83}}$\pm$03.37 \\
Scherer2015 & BAC & 21.62$\pm$00.66 & \textcolor{blue}{\textbf{26.48}}$\pm$02.48 & 21.83$\pm$00.17 & 21.22$\pm$01.80 & 20.95$\pm$01.09 & 26.06$\pm$01.07 & \textcolor{red}{\textbf{32.00}}$\pm$00.55 \\
Srisrisawang2024 & BAC & 34.56$\pm$00.96 & 18.00$\pm$01.78 & 52.94$\pm$00.88 & 49.41$\pm$00.99 & 49.92$\pm$00.72 & \textcolor{red}{\textbf{57.90}}$\pm$01.64 & \textcolor{blue}{\textbf{54.40}}$\pm$00.95 \\
Wang2017 & BAC & 13.31$\pm$05.51 & 58.67$\pm$01.85 & \textcolor{blue}{\textbf{96.31}}$\pm$00.45 & 85.85$\pm$07.27 & 13.08$\pm$00.88 & 65.79$\pm$02.09 & \textcolor{red}{\textbf{96.83}}$\pm$00.91 \\
\midrule
\multicolumn{9}{c}{\cellcolor{orange!8}\textit{Event-related and evoked responses}} \\
ERP CORE-ERN & BAC & 78.06$\pm$04.59 & 49.40$\pm$01.12 & \textcolor{blue}{\textbf{81.96}}$\pm$01.52 & 81.32$\pm$01.04 & 81.57$\pm$03.26 & \textcolor{red}{\textbf{84.61}}$\pm$01.92 & 80.07$\pm$02.94 \\
ERP CORE-N170 & BAC & 67.03$\pm$00.61 & 49.84$\pm$00.36 & 72.14$\pm$00.55 & 71.14$\pm$00.55 & 68.74$\pm$02.04 & \textcolor{red}{\textbf{74.03}}$\pm$05.33 & \textcolor{blue}{\textbf{73.27}}$\pm$00.91 \\
ERP CORE-N400 & BAC & 60.45$\pm$01.15 & 50.07$\pm$00.64 & \textcolor{blue}{\textbf{64.62}}$\pm$01.09 & 60.42$\pm$01.13 & 64.20$\pm$01.55 & 64.51$\pm$01.97 & \textcolor{red}{\textbf{66.67}}$\pm$00.75 \\
Schreuder2010 & BAC & \textcolor{blue}{\textbf{66.70}}$\pm$00.16 & 50.25$\pm$00.21 & 66.05$\pm$00.40 & 63.11$\pm$00.62 & 62.97$\pm$00.39 & 64.70$\pm$01.00 & \textcolor{red}{\textbf{69.87}}$\pm$00.67 \\
\midrule
\multicolumn{9}{c}{\cellcolor{purple!8}\textit{Cognitive state and language decoding}} \\
Zyma2019 & BAC & 65.28$\pm$02.34 & 66.40$\pm$03.04 & 68.92$\pm$02.42 & 72.02$\pm$01.30 & 72.35$\pm$00.41 & \textcolor{red}{\textbf{72.88}}$\pm$02.81 & \textcolor{blue}{\textbf{72.50}}$\pm$05.04 \\
Hinss2023 & BAC & 45.89$\pm$00.19 & 43.18$\pm$04.73 & 65.46$\pm$03.17 & 61.37$\pm$01.70 & \textcolor{blue}{\textbf{66.01}}$\pm$00.94 & 65.57$\pm$04.49 & \textcolor{red}{\textbf{72.93}}$\pm$00.42 \\
ZuCo & Top-5 & 6.47$\pm$01.29 & 7.67$\pm$00.39 & 8.95$\pm$00.26 & 2.25$\pm$00.30 & 11.32$\pm$00.56 & \textcolor{blue}{\textbf{11.39}}$\pm$00.40 & \textcolor{red}{\textbf{12.11}}$\pm$00.39 \\
Nieuwland2018 & Top-5 & 20.73$\pm$01.69 & 5.03$\pm$00.09 & 28.34$\pm$01.27 & 29.73$\pm$00.66 & 27.99$\pm$01.37 & \textcolor{blue}{\textbf{33.73}}$\pm$00.68 & \textcolor{red}{\textbf{36.27}}$\pm$00.12 \\
\midrule
\textbf{Macro (14 BAC tasks)} &
& \multicolumn{1}{c}{53.29}
& \multicolumn{1}{c}{50.01}
& \multicolumn{1}{c}{\textcolor{blue}{\textbf{70.33}}}
& \multicolumn{1}{c}{68.04}
& \multicolumn{1}{c}{62.46}
& \multicolumn{1}{c}{67.30}
& \multicolumn{1}{c}{\textcolor{red}{\textbf{73.14}}} \\
\bottomrule
\end{tabular}
\end{table}

%% file: tables/tab_ablation.tex
\begin{table}[H]
\centering
\caption{\textbf{Component ablations: frozen single-task transfer with 10\% labels.} ResID: identity residualization; Reg.: state regularization. Block mask ``--'' denotes temporal masking, not unmasked input. Exclusion means additional context removal; Random is size matched. BAC is reported as \% (mean $\pm$ sample SD); Macro averages four tasks. \textcolor{red}{\textbf{Red}} and \textcolor{blue}{\textbf{blue}} mark the best and second-best mean per column.}
\label{tab:ablation}
\vskip 0.04in
\fontsize{6.6}{7.2}\selectfont
\setlength{\tabcolsep}{1.8pt}
\begin{tabular}{@{}lccccc@{\hspace{5pt}}ccccc@{}}
\toprule
\multicolumn{6}{c}{\textbf{Pretraining design}} & \multicolumn{5}{c}{\textbf{Frozen Single-Task BAC}} \\
\cmidrule(lr){1-6}\cmidrule(lr){7-11}
\textbf{Variant} & \textbf{Block mask} & \textbf{Target} & \textbf{ResID} & \textbf{Exclusion} & \textbf{Reg.} & \textbf{TUAB} & \textbf{TUEV} & \textbf{SEED} & \textbf{PhysioMI} & \textbf{Macro} \\
\midrule
\rowcolor{black!7}\multicolumn{11}{@{}l}{\textit{Target construction}} \\
Shared-target baseline       & -- & Shared & -- & None & $\checkmark$ & 73.27$\pm$00.35 & 38.37$\pm$02.41 & 40.53$\pm$01.78 & 25.57$\pm$01.42 & 44.44 \\
+ Block masking              & $\checkmark$ & Shared & -- & None & $\checkmark$ & 74.27$\pm$00.23 & 44.17$\pm$01.00 & 38.37$\pm$04.12 & 26.20$\pm$00.78 & 45.75 \\
+ EMA target                 & $\checkmark$ & EMA & -- & None & $\checkmark$ & 78.17$\pm$00.23 & 51.40$\pm$00.98 & \textcolor{blue}{\textbf{51.97}}$\pm$00.32 & 25.97$\pm$01.12 & 51.88 \\
+ Identity residualization   & $\checkmark$ & EMA & $\checkmark$ & None & $\checkmark$ & 78.73$\pm$00.15 & 59.77$\pm$02.17 & \textcolor{blue}{\textbf{51.97}}$\pm$01.27 & \textcolor{blue}{\textbf{26.50}}$\pm$00.35 & \textcolor{blue}{\textbf{54.24}} \\
\addlinespace[1pt]
\rowcolor{black!7}\multicolumn{11}{@{}l}{\textit{Representation-stability control}} \\
Full NSP w/o regularization  & $\checkmark$ & EMA & $\checkmark$ & Topology & -- & \textcolor{red}{\textbf{80.83}}$\pm$00.29 & 48.27$\pm$05.12 & 33.30$\pm$00.00 & 25.13$\pm$01.01 & 46.88 \\
\addlinespace[1pt]
\rowcolor{black!7}\multicolumn{11}{@{}l}{\textit{Matched residualization--context comparison}} \\
Equal-size random exclusion  & $\checkmark$ & EMA & -- & Random & $\checkmark$ & 78.92$\pm$00.48 & 58.31$\pm$01.54 & 49.87$\pm$01.12 & 25.38$\pm$00.96 & 53.12 \\
NSP with random exclusion     & $\checkmark$ & EMA & $\checkmark$ & Random & $\checkmark$ & 79.36$\pm$00.51 & 59.18$\pm$01.47 & 50.94$\pm$01.06 & 25.47$\pm$00.93 & 53.74 \\
NSP w/o identity residualization & $\checkmark$ & EMA & -- & Topology & $\checkmark$ & 79.71$\pm$00.49 & \textcolor{blue}{\textbf{59.86}}$\pm$01.55 & 51.58$\pm$00.97 & 25.55$\pm$00.98 & 54.18 \\
Full NSP                      & $\checkmark$ & EMA & $\checkmark$ & Topology & $\checkmark$ & \textcolor{blue}{\textbf{80.13}}$\pm$00.55 & \textcolor{red}{\textbf{60.70}}$\pm$01.61 & \textcolor{red}{\textbf{53.33}}$\pm$01.00 & \textcolor{red}{\textbf{28.63}}$\pm$01.37 & \textcolor{red}{\textbf{55.70}} \\
\bottomrule
\end{tabular}
\end{table}

%% file: appendix.tex

\section{Related Work}
\label{app:related-work}

\subsection{EEG Foundation Models}

EEG foundation models pretrain reusable encoders on heterogeneous, largely unlabeled recordings and adapt them across clinical, BCI, affective, and cognitive tasks. Reconstruction-based models differ in what they ask the encoder to preserve. EEGPT~\citep{eegpt} and REVE~\citep{reve} recover masked waveform content, with REVE additionally encoding electrode geometry to support variable layouts. CBraMod~\citep{cbraMod} reconstructs masked EEG patches through separate spatial and temporal attention, whereas CSBrain~\citep{csbrain} reconstructs spectrogram representations with specialized spatial--temporal modeling. LaBraM~\citep{labram} first converts EEG patches into neural tokens and predicts the masked token identities. DeWave~\citep{duan2023dewave} and CodeBrain~\citep{ma2025} likewise introduce discrete codes that replace direct waveform regression with code prediction. These choices determine whether supervision emphasizes sample-level detail, spectral structure, patch content, or a learned vocabulary.

Contrastive EEG foundation models specify transfer through invariance between related views rather than reconstruction. BENDR~\citep{bendr} masks the convolutional sequence and trains a Transformer context representation to identify its corresponding latent target among within-recording negatives. BIOT~\citep{biot} tokenizes channels into a unified biosignal sequence and aligns an original recording with a view perturbed by channel and token dropping. Contrastive objectives can suppress nuisance variation, but the learned invariance depends on positive-pair construction, negative sampling, and augmentations. This is consequential for EEG because the validity of a transformation is task dependent, and an augmentation that is benign for one paradigm may alter task-relevant temporal or spatial structure in another~\citep{lashgari2020augmentation,rommel2022augmentation}.

Autoregressive EEG foundation models instead factor a recording into an ordered prediction problem. ECHO~\citep{liu2026echo} uses contextual sequence-to-sequence modeling, THD-BAR~\citep{yang2025} organizes autoregression with topology-hierarchical structure, and KAST-BAR~\citep{wang2026kastbar} introduces knowledge-anchored semantic dynamics. This family can model long temporal dependencies and supports generative or next-token objectives, but its evidence path is inherently directional. In strongly autocorrelated EEG, next-patch prediction may therefore reward accurate local extrapolation without requiring the representation to integrate spatially distributed evidence.

\subsection{Latent Prediction and JEPA}

Joint-Embedding Predictive Architectures (JEPA) learn by predicting representations of unobserved regions in a learned embedding space rather than reconstructing the input~\citep{lecun2022path}. I-JEPA established masked latent prediction for images~\citep{ijepa}, V-JEPA extended feature prediction to video~\citep{vjepa}, and data2vec developed a related target-representation objective across speech, vision, and language~\citep{baevski2022data2vec}. LeJEPA subsequently provided a theoretically grounded and scalable formulation with an explicit representation regularizer~\citep{balestriero2025lejepa,vanassel2025jointembedding}. The same predictive-state principle now reaches world modeling: V-JEPA~2 couples large-scale latent video pretraining with an action-conditioned predictive model for physical-world planning~\citep{assran2025vjepa2}.

JEPA has also been applied to brain signals. Brain-JEPA predicts masked fMRI representations with functional-coordinate encoding and spatiotemporal masking~\citep{dong2024brainjepa}. For EEG, S-JEPA introduces spatial block masking for cross-dataset transfer~\citep{guetschel2024sjepa}; Laya applies latent prediction at larger scale~\citep{laya}; STST-JEPA combines shallow EMA targets with an auxiliary reconstruction objective~\citep{segal2026ststjepa}; and EEG-JEPA structures latent targets by content, support, and encoder depth~\citep{li2026eegjepa}. PATCHCODE~\citep{yu2026patchcode} combines continuous patch-level latent prediction with auxiliary discrete-code supervision from a frozen tokenizer; we therefore discuss it within the latent-predictive family. These methods establish latent prediction for brain signals, but they do not combine target residualization against channel/time identity with a context guard that excludes the target's immediate spatial and temporal neighborhood.

\section{Implementation Details}
\label{app:implementation}

\subsection{Architecture}

The NSP Large configuration has 42.83M trainable Context-side parameters, including its predictor. The Cascaded Spatial--Temporal Transformer (CST) contains 8 blocks, each with spatial attention ($d_{\mathrm{model}}=512$, 8 heads) across channels, temporal attention with rotary position encoding~\citep{su2024roformer} across patches, and a SwiGLU (Swish-gated linear unit) feed-forward sublayer with expansion factor 4. Patch duration is $0.2\,\mathrm{s}$ at $f_s=200\,\mathrm{Hz}$. The predictor is a two-layer multilayer perceptron (MLP) with hidden width 768, output width 512, and Gaussian error linear unit (GELU) activation, followed by LayerNorm and $\ell_2$ normalization. The Target Encoder aggregates the last three blocks ($\mathcal{I}_T=\{5,6,7\}$, zero-indexed). The EMA coefficient follows a cosine schedule from 0.996 to 0.9999 over the pretraining duration.

\subsection{Identity residualization as an additive projection}
\label{app:residualization-projection}

The residualization in Eq.~(\ref{eq:identity-residual}) has an exact least-squares interpretation. All expressions below act independently on each latent feature, and the vector notation collects those feature-wise problems. Because $\mathbf{u}$ is standardized to have zero global mean over the complete $B\times P\times C$ grid, consider the additive subspace formed by a channel term $\mathbf{a}_c\in\mathbb{R}^d$ and a relative-time term $\mathbf{t}_p\in\mathbb{R}^d$. We identify their zero levels through the constraints $\sum_c\mathbf{a}_c=\mathbf{0}$ and $\sum_p\mathbf{t}_p=\mathbf{0}$, and fit the additive component by
\begin{equation}
(\mathbf{a}^{\star},\mathbf{t}^{\star})
=\underset{\substack{\{\mathbf{a}_c\},\{\mathbf{t}_p\}\\
\sum_c\mathbf{a}_c=\mathbf{0},\;\sum_p\mathbf{t}_p=\mathbf{0}}}{\operatorname*{arg\,min}}
\sum_{b=1}^{B}\sum_{p=1}^{P}\sum_{c=1}^{C}
\left\|\mathbf{u}_{b,p,c}-\mathbf{a}_c-\mathbf{t}_p\right\|_2^2.
\label{eq:app-additive-projection}
\end{equation}
Under uniform weighting and a complete patch--channel grid, the normal equations give
\begin{equation}
\mathbf{a}^{\star}_c
=\frac{1}{BP}\sum_{b,p}\mathbf{u}_{b,p,c}
=\boldsymbol{\mu}^{\mathrm{ch}}_c,
\qquad
\mathbf{t}^{\star}_p
=\frac{1}{BC}\sum_{b,c}\mathbf{u}_{b,p,c}
=\boldsymbol{\mu}^{\mathrm{time}}_p.
\label{eq:app-additive-solution}
\end{equation}
Thus, $\mathbf{r}=\mathbf{u}-\mathbf{a}^{\star}-\mathbf{t}^{\star}$ is the orthogonal residual after projecting $\mathbf{u}$ onto this additive identity subspace. In particular, its channel and relative-time marginals vanish:
\begin{equation}
\sum_{b,p}\mathbf{r}_{b,p,c}=\mathbf{0}\quad\forall c,
\qquad
\sum_{b,c}\mathbf{r}_{b,p,c}=\mathbf{0}\quad\forall p.
\label{eq:app-zero-marginals}
\end{equation}
Consequently, $\sum_{b,p,c}\mathbf{r}_{b,p,c}^{\top}(\mathbf{a}_c+\mathbf{t}_p)=0$ for any additive channel--time function, which is the defining orthogonality condition of the projection. This result is deliberately limited: residualization removes the best additive channel and relative-time main effects under the current batch weighting, but it does not remove nonlinear identity information or channel--time interactions, and it does not imply channel invariance.

\subsection{Pretraining hyperparameters}

Pretraining uses AdamW~\citep{loshchilov2019adamw} with learning rate $1\times10^{-4}$, weight decay 0.05, global batch size 384, gradient clipping 1.0, and 5,000 linear-warmup steps followed by cosine annealing. Target blocks cover 50\% of the token grid, with temporal spans of 2--10 patches and spatial spans of 3--7 coordinate-near channels. The spatial guard includes the two nearest neighbors by Euclidean distance, and the temporal guard includes adjacent patches. State regularization uses $K=64$ random unit directions with $\lambda_{\mathrm{reg}}=0.1$.

TUEG contains 69,652 files from 14,987 subjects and 26,846 sessions recorded with heterogeneous clinical montages and sampling rates~\citep{obeid2016tueg}. We map each usable recording to 19 standard electrodes from the international 10--20 system and resample it to 200~Hz. After filtering and common-average referencing, the implementation clips amplitudes to $\pm500~\mu$V before recording-level z-normalization, then forms non-overlapping 6-s segments and applies quality checks to the resulting normalized windows. Exceeding $500~\mu$V is therefore not, by itself, a rule for rejecting a whole segment. The pretraining collection contains 2.2M examples of shape $19\times1{,}200$.

\input{tables/app_pretraining_details}

\subsection{Preprocessing protocols}

\textbf{EEG-FM-Bench}~\citep{eegfmbench}: Event selection, filtering, resampling, windowing, channel mapping, and unit conversion follow the benchmark's task/model adapter. The 19-channel, 200-Hz specification above describes NSP pretraining, not a requirement that every downstream source recording have this montage or sampling rate. Downstream tasks can supply different channel counts through their configured adapters; the source montage and the representation presented to the pretrained encoder must be distinguished.

\textbf{NeuralBench-EEG-Core}~\citep{neuralbench}: Each task uses its specified YAML configuration for event selection, filtering, resampling, windowing, and channel mapping. Tasks are not forced to 200~Hz or 19 channels; configurations vary by dataset.

\subsection{Downstream adaptation}

\textbf{EEG-FM-Bench adaptation}: 30 epochs with 3-epoch linear warmup; AdamW~\citep{loshchilov2019adamw}; batch size 128; gradient clipping 1.0; weight decay 0.01; warmup-freeze encoder 1 epoch (allows classification head to stabilize). Learning rate and pooling head hidden dimension are searched per task. Encoder learning rate is set to classifier learning rate $\times$ (1/10, 1/5, or 1/2). Seeds 42--46 for multi-task full fine-tuning; seeds 42--44 for frozen, LoRA, and single-task protocols.

\textbf{NeuralBench-EEG-Core adaptation}: Single-task per dataset; task-specific YAML configurations for epochs, batch size, learning rate, and architecture; seeds 33, 34, 35; checkpoint selection on the task-specific validation metric; test evaluation only on the selected checkpoint.

\section{Downstream Evaluation Details}
\label{app:downstream-details}

\subsection{Evaluation Metrics}

Unless stated otherwise, classification and retrieval accuracies, F1, Cohen's $\kappa$, and areas under curves are reported as percentages; differences between them are percentage points. Figures~\ref{fig:duration} and~\ref{fig:label-efficiency} use the equivalent 0--1 BAC scale. MedR is an unscaled median rank, and cosine-based diagnostics are dimensionless.

Let $K_{\mathrm{cls}}$ be the number of classes, $n_k$ the support of class $k$, and $N$ the number of evaluated examples. Balanced accuracy averages class recall,
\begin{equation}
\text{BAC} = \frac{1}{K_{\mathrm{cls}}} \sum_{k=1}^{K_{\mathrm{cls}}} \frac{\text{TP}_k}{\text{TP}_k + \text{FN}_k},
\end{equation}
where $\mathrm{TP}_k$ and $\mathrm{FN}_k$ are true-positive and false-negative counts. BAC gives each class equal weight under imbalance. With class precision $P_k$ and recall $R_k$, weighted F1 is
\begin{equation}
\text{F1}_{\text{weighted}} = \sum_{k=1}^{K_{\mathrm{cls}}} \frac{n_k}{N} \frac{2 P_k R_k}{P_k + R_k}.
\end{equation}
Cohen's $\kappa$ removes agreement expected from the empirical class marginals,
\begin{equation}
\kappa = \frac{p_o - p_e}{1 - p_e}, \qquad p_e = \sum_{k=1}^{K_{\mathrm{cls}}} \frac{n_k}{N} \frac{\hat{n}_k}{N},
\end{equation}
where $p_o$ is observed agreement and $\hat{n}_k$ is the number of predictions assigned to class $k$.

For binary tasks, the area under the receiver-operating-characteristic curve and the area under the precision-recall curve are
\begin{equation}
\text{AUROC} = \int \text{TPR}\, d\text{FPR}, \qquad \text{AUCPR} = \int \text{Precision}\, d\text{Recall}.
\end{equation}
Here TPR and FPR denote true-positive and false-positive rates. AUROC measures threshold-independent ranking over both classes, while AUCPR is particularly informative when the positive class is rare. The two language retrieval tasks use
\begin{equation}
\text{Top-5} = \frac{1}{N} \sum_{i=1}^{N} \mathbb{1}[y_i \in \text{Top5}(\hat{\mathbf{p}}_i)],
\end{equation}
which measures whether the correct item appears among the five highest-scoring candidates and is not numerically comparable to BAC.

\subsection{Baseline Models}

We compare NSP against six official NeuralBench-EEG-Core foundation-model baselines (Table~\ref{tab:full_finetuning}) under the released benchmark protocol~\citep{neuralbench}. EEG-FM-Bench uses a distinct comparison set that replaces LUNA with EEGPT and CSBrain~\citep{eegfmbench}. All dataset-level entries report a mean and sample standard deviation using the corresponding benchmark's task-specific metric definitions.

\textbf{BENDR}~\citep{bendr} adapts wav2vec-style masked contrastive prediction to EEG. At a masked time step, the Transformer context is trained to identify the corresponding unmasked convolutional representation among within-sequence negatives.

\textbf{BIOT}~\citep{biot} tokenizes each biosignal channel into fixed-length segments and rearranges them into a unified token sequence, allowing one encoder to process heterogeneous channel sets and durations. Its unsupervised objective predicts an original-signal embedding from a channel- and token-dropped view with a contrastive loss.

\textbf{LaBraM}~\citep{labram} uses neural tokenization with masked prediction on large-scale heterogeneous EEG. A learned codebook quantizes EEG patches into discrete tokens; a Transformer predicts masked tokens from visible context. Its pretraining corpus contains approximately 2,500 hours collected from about 20 datasets, and the model achieves strong transfer on medical tasks.

\textbf{CBraMod}~\citep{cbraMod} reconstructs masked EEG patches using cascaded spatial--temporal attention. Its criss-cross Transformer applies separate spatial (cross-channel) and temporal (cross-time) attention to the patch grid.

\textbf{EEGPT}~\citep{eegpt} reconstructs masked EEG content while learning electrode-aware representations through a target-side pretraining design. \textbf{CSBrain}~\citep{csbrain} operates on EEG spectrograms and reconstructs masked spectral regions with specialized spatial--temporal attention. Both models are evaluated by EEG-FM-Bench but are not part of the released NeuralBench baseline set.

\textbf{LUNA}~\citep{luna2025} maps variable electrode layouts into a fixed-size latent through learned queries and cross-attention, then applies temporal attention in that topology-agnostic space. It is pretrained on TUEG and Siena with masked-patch reconstruction.

\textbf{REVE}~\citep{reve} combines masked waveform reconstruction with a four-dimensional positional encoding that accommodates variable electrode locations and recording layouts. Its pretraining corpus aggregates recordings from 92 sources and approximately 25,000 subjects.

The public-model reproduction uses the benchmark's task-specific preprocessing and adaptation configuration with seeds 33, 34, and 35 and checkpoint selection by the task-specific validation metric. NSP follows the same task identities and metric definitions, with validation-selected task-specific readouts.

\subsection{Adaptation Protocols}

\textbf{Full fine-tuning}: All encoder parameters and classification head are updated. This is the primary protocol for NeuralBench-EEG-Core and EEG-FM-Bench Full FT single-task and multi-task evaluations.

\textbf{Frozen evaluation}: Encoder parameters are frozen; only classification heads are trained. EEG-FM-Bench uses multi-task evaluation, whereas the component ablations train a head separately for each dataset under single-task evaluation.

\textbf{LoRA}: Low-rank adaptation~\citep{hu2021lora} updates low-rank matrices injected into attention layers while freezing the base encoder. Rank is set to 8 for all experiments.

All protocols use the corresponding validation metric for checkpoint selection and report the matching test metric only on the selected checkpoint: BAC for classification and Top-5 accuracy for retrieval.

\section{Complete Benchmark Results}
\label{app:complete-results}

\subsection{NeuralBench-EEG-Core}
\label{app:neuralbench-complete}

NeuralBench-EEG-Core v1.0 provides task-specific preprocessing, splits, adaptation settings, and evaluation metrics for heterogeneous EEG datasets~\citep{neuralbench}. We retain 16 datasets after removing dataset--task pairs also evaluated by EEG-FM-Bench, yielding four clinical and neurological-disease tasks, four BCI and sensorimotor tasks, four event-related or evoked-response tasks, and four cognitive-state or language-decoding tasks. The following four subsections first describe the datasets in each group and then compare model performance. Fourteen classification datasets use BAC as the primary metric, whereas ZuCo and Nieuwland2018 use subject-aggregated Top-5 retrieval accuracy. Every table additionally reports two task-appropriate metrics and marks the best and second-best mean for each metric.

\input{tables/app_complete_neuralbench_results}

\subsection{EEG-FM-Bench}
\label{app:eegfmbench}

The EEG-FM-Bench protocol~\citep{eegfmbench} evaluates 14 datasets across 10 EEG paradigms under four adaptation strategies: single-task full fine-tuning, multi-task full fine-tuning, frozen-backbone multi-task transfer, and multi-task LoRA. Each dataset paragraph introduces the task and recording scale before summarizing NSP across the four strategies. The tables report BAC together with F1 and Cohen's $\kappa$ for multiclass tasks, or AUROC and AUCPR for binary tasks. Across all 14 datasets, NSP obtains 63.94 macro BAC with multi-task full fine-tuning.

\input{tables/app_complete_eegfm_results}

\section{Additional Ablations}
\label{app:additional-ablations}

The pretraining-component ablations use frozen single-task evaluation: the encoder is fixed and a head is trained separately on each of TUAB, TUEV, SEED, and PhysioMI. Unless 100\% is specified, heads use 10\% of training labels. This differs from EEG-FM-Bench's frozen multi-task protocol. The final study instead varies the head under full fine-tuning.

\subsection{Cumulative component ablation}

To determine how transfer changes as prediction targets and available context are constrained, we compare the cumulative configurations and the state-regularization control in Table~\ref{tab:ablation}. Macro BAC rises from 44.44 for the shared-target baseline to 45.75 with block masking, 51.88 with EMA, 54.24 with identity residualization, and 55.70 for Full NSP. Task-wise changes are not uniformly positive, and removing regularization lowers the macro to 46.88. The matched context comparisons in Table~\ref{tab:ablation} separately assess exclusion geometry at a fixed visible-token budget.

\subsection{Guard design}

\begin{figure}[H]
\centering
\includegraphics[width=0.75\textwidth]{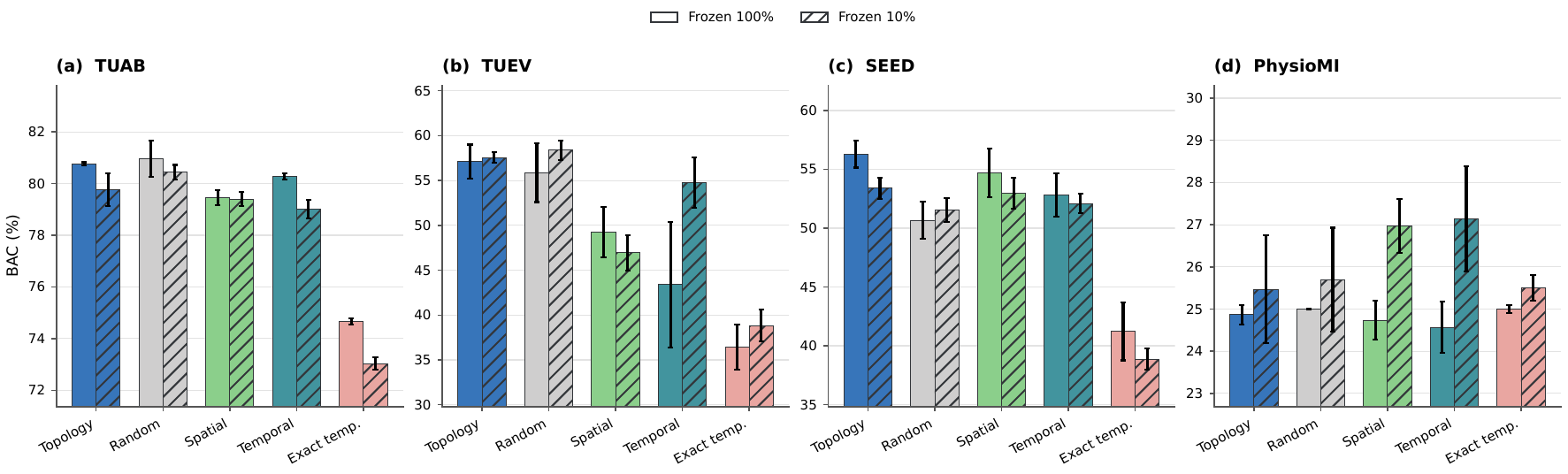}
\caption{\textbf{Context exclusion and temporal-masking controls.} Frozen single-task BAC with 10\% and 100\% labels. Topology denotes NSP without identity residualization, whereas Random denotes NSP with random exclusion and residualization. Spatial-only and temporal-only exclusion modify available context; the exact-temporal condition is a separate temporal-only baseline without EMA or residualization. These controls are not all isolated changes in guard geometry.}
\label{fig:guard-design}
\end{figure}

To examine how masking and context exclusion affect transfer, we compare random, spatial-only, and temporal-only exclusion with the complete topology-separated design on four tasks (Figure~\ref{fig:guard-design}). The exact-temporal condition is a separate temporal-only masking baseline without EMA or residualization, not an isolated guard variant. Table~\ref{tab:ablation} provides matched visible-token controls for the exclusion-geometry comparison.

\subsection{Residualization design}

\begin{figure}[H]
\centering
\includegraphics[width=0.75\textwidth]{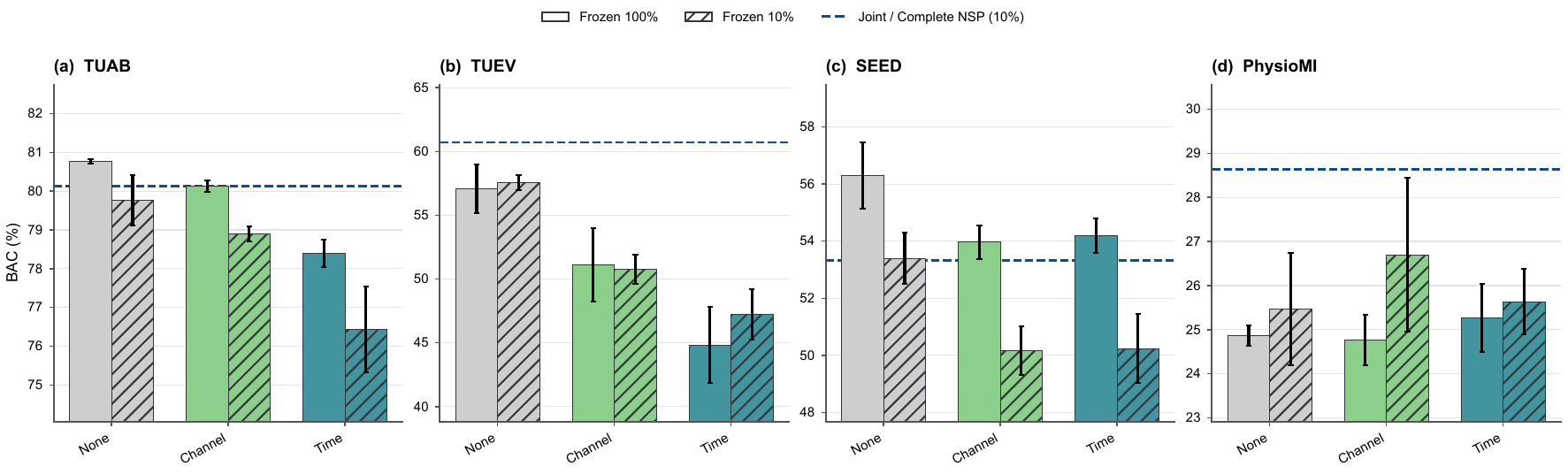}
\caption{\textbf{Partial identity-residualization variants.} Frozen single-task BAC without residualization and with channel-only or time-only residualization using 10\% and 100\% labels. Dashed lines show complete joint residualization with 10\% labels.}
\label{fig:residualization}
\end{figure}

To isolate which positional marginal is removed by the target construction, we compare no residualization with channel-only, time-only, and joint residualization under matched frozen single-task protocols (Figure~\ref{fig:residualization}). Channel and time removal produce different task-wise changes, and neither partial variant reproduces the complete joint configuration uniformly.

\subsection{State regularization coefficient}

\begin{figure}[H]
\centering
\includegraphics[width=0.98\textwidth]{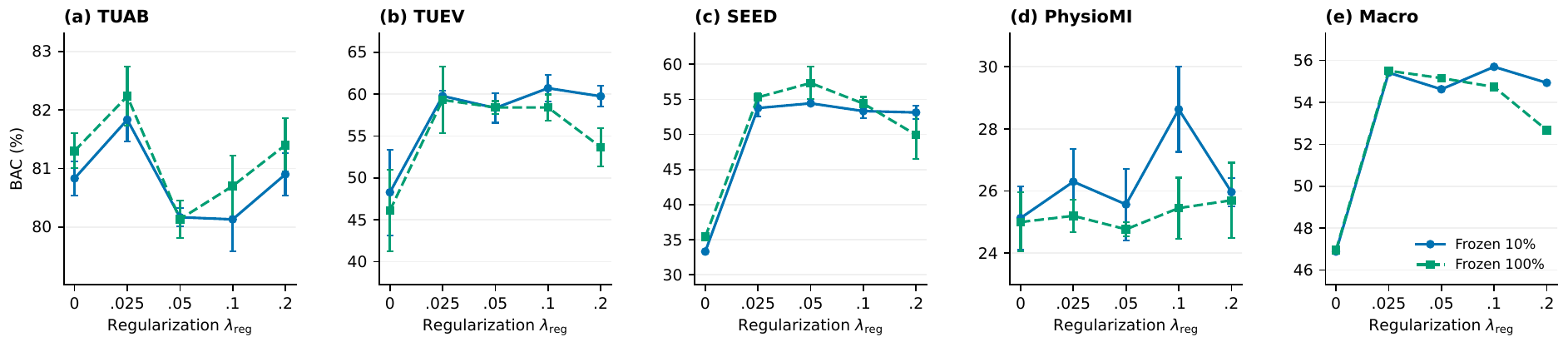}
\caption{\textbf{Sensitivity to moment-based state regularization.} Single-task heads are trained with 10\% or 100\% of labels while the encoder remains frozen. Task panels show BAC with sample SD, and Macro is the unweighted mean of the four task means. Macro uncertainty is not inferred from task-wise SDs. The highest observed macro occurs at $\lambda_{\mathrm{reg}}=0.025$ with 100\% labels and at $\lambda_{\mathrm{reg}}=0.1$ with 10\% labels.}
\label{fig:sigreg-sensitivity}
\end{figure}

To test sensitivity to moment-based state regularization, we vary $\lambda_{\mathrm{reg}}$ and evaluate the frozen encoder with single-task heads trained on either 10\% or 100\% of the available labels (Figure~\ref{fig:sigreg-sensitivity}); these are not full-parameter fine-tuning runs. Under Frozen 100\%, the four-task macro is 46.95 without regularization and 55.51, 55.16, 54.74, and 52.68 at $\lambda_{\mathrm{reg}}=0.025$, $0.05$, $0.1$, and $0.2$, respectively. All tested nonzero weights improve on the unregularized setting in both protocols. The highest observed macro occurs at $\lambda_{\mathrm{reg}}=0.025$ with 100\% labels and at $\lambda_{\mathrm{reg}}=0.1$ with 10\% labels; task-wise responses differ, and these means do not establish a statistically unique optimum.

\subsection{Target layer selection}

\begin{figure}[H]
\centering
\includegraphics[width=0.75\textwidth]{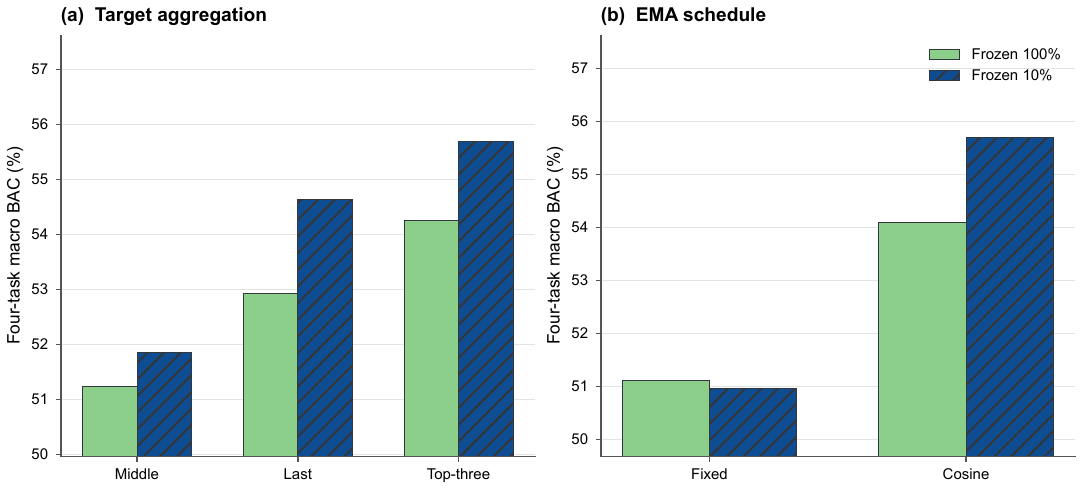}
\caption{\textbf{Target aggregation and EMA schedule.} Aggregate balanced accuracy for middle-layer, last-layer, and top-three-layer targets (left), and fixed versus cosine EMA schedules (right). Bars compare frozen single-task evaluation with 100\% and 10\% labels.}
\label{fig:target-ema}
\end{figure}

To identify which encoder depth supplies the most transferable latent target, we compare middle-layer, last-layer, and top-three-layer aggregation under both frozen single-task label budgets (Figure~\ref{fig:target-ema}, left). Under Frozen 100\%, the top-three target obtains 54.25 macro BAC, compared with 52.93 for the last-layer target and 51.24 for the middle-layer target.

\subsection{EMA schedule}

To test whether a progressively slower Target Encoder improves supervision, we compare fixed and cosine EMA schedules with the remaining configuration held constant (Figure~\ref{fig:target-ema}, right). The cosine schedule is stronger under both evaluations: 54.10 versus 51.12 macro BAC with Frozen 100\%, and 55.70 versus 50.96 with Frozen 10\%.

\subsection{Task-wise comparison of EMA target and Full NSP}

\begin{figure}[H]
\centering
\includegraphics[width=0.75\textwidth]{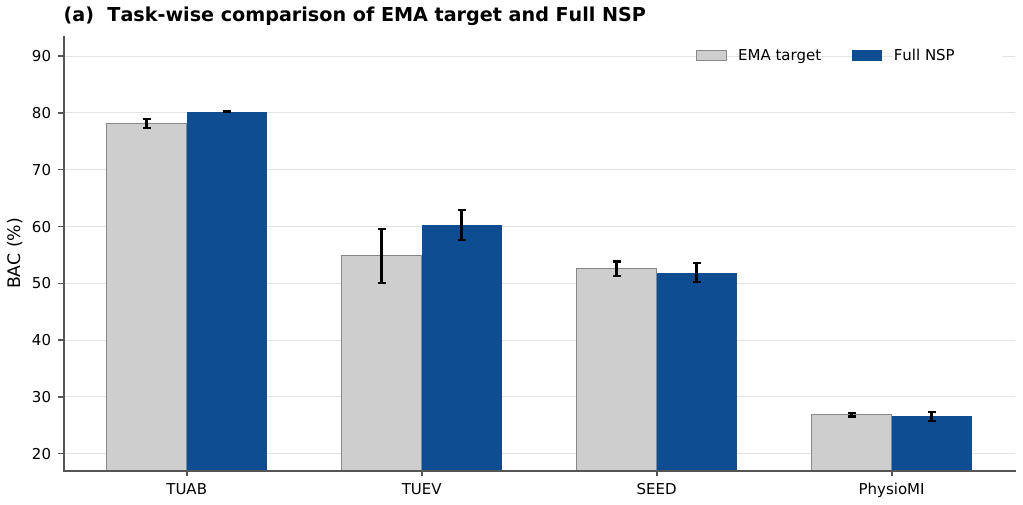}
\caption{\textbf{Task-wise comparison of EMA-target and Full NSP configurations.} Frozen single-task BAC with 10\% labels on TUAB, TUEV, SEED, and PhysioMI. The complete method improves some tasks while leaving others similar; error bars summarize variability across repeated evaluations.}
\label{fig:pretraining-seed}
\end{figure}

To verify that the aggregate component difference is not driven by a single task, we compare the EMA-target configuration with Full NSP on TUAB, TUEV, SEED, and PhysioMI (Figure~\ref{fig:pretraining-seed}). The task-wise view shows where the complete context design changes transfer rather than compressing the comparison into one macro score.

\subsection{Fine-tuning head design}

\begin{figure}[H]
\centering
\includegraphics[width=0.98\textwidth]{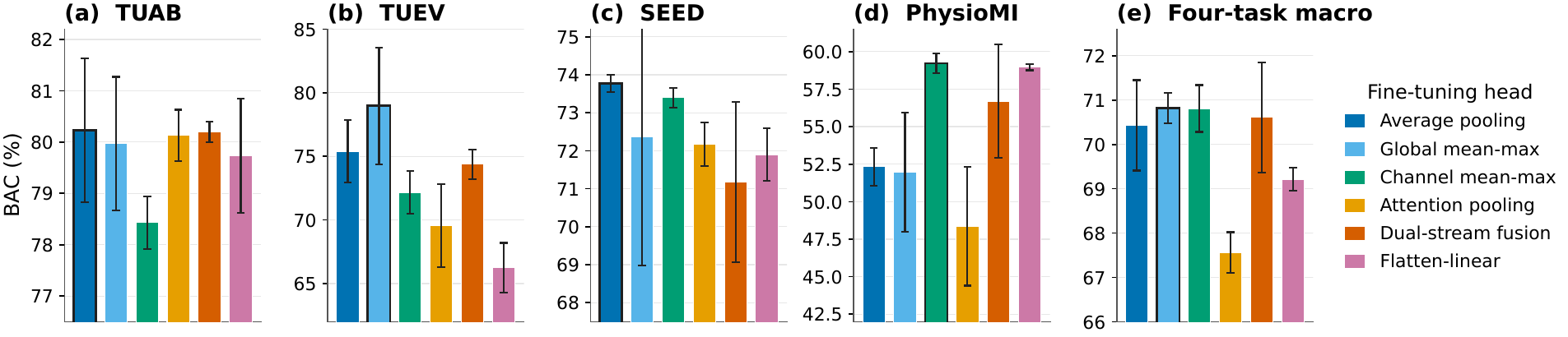}
\caption{\textbf{Effect of fine-tuning head design.} Mean balanced accuracy on TUAB, TUEV, SEED, and PhysioMI and their four-task macro. Error bars show sample standard deviations; colors distinguish six head designs, and dark outlines mark the highest mean in each panel.}
\label{fig:finetuning-heads}
\end{figure}

To determine whether downstream conclusions depend on one readout design, we compare six classification heads while holding fine-tuning and model selection fixed (Figure~\ref{fig:finetuning-heads}). Global mean--max gives the highest four-task macro BAC (70.82), closely followed by channel mean--max (70.81), dual-stream fusion (70.61), and average pooling (70.43). The task-wise ordering is less uniform: average pooling has the highest mean on TUAB and SEED, global mean--max on TUEV, and channel mean--max on PhysioMI. The results therefore do not identify a uniformly dominant head; small aggregate differences should be interpreted together with the reported variability.

\section{Model Scale and Control Studies}
\label{app:scale}

\subsection{Model scale}

\begin{figure}[H]
\centering
\includegraphics[width=0.8\textwidth]{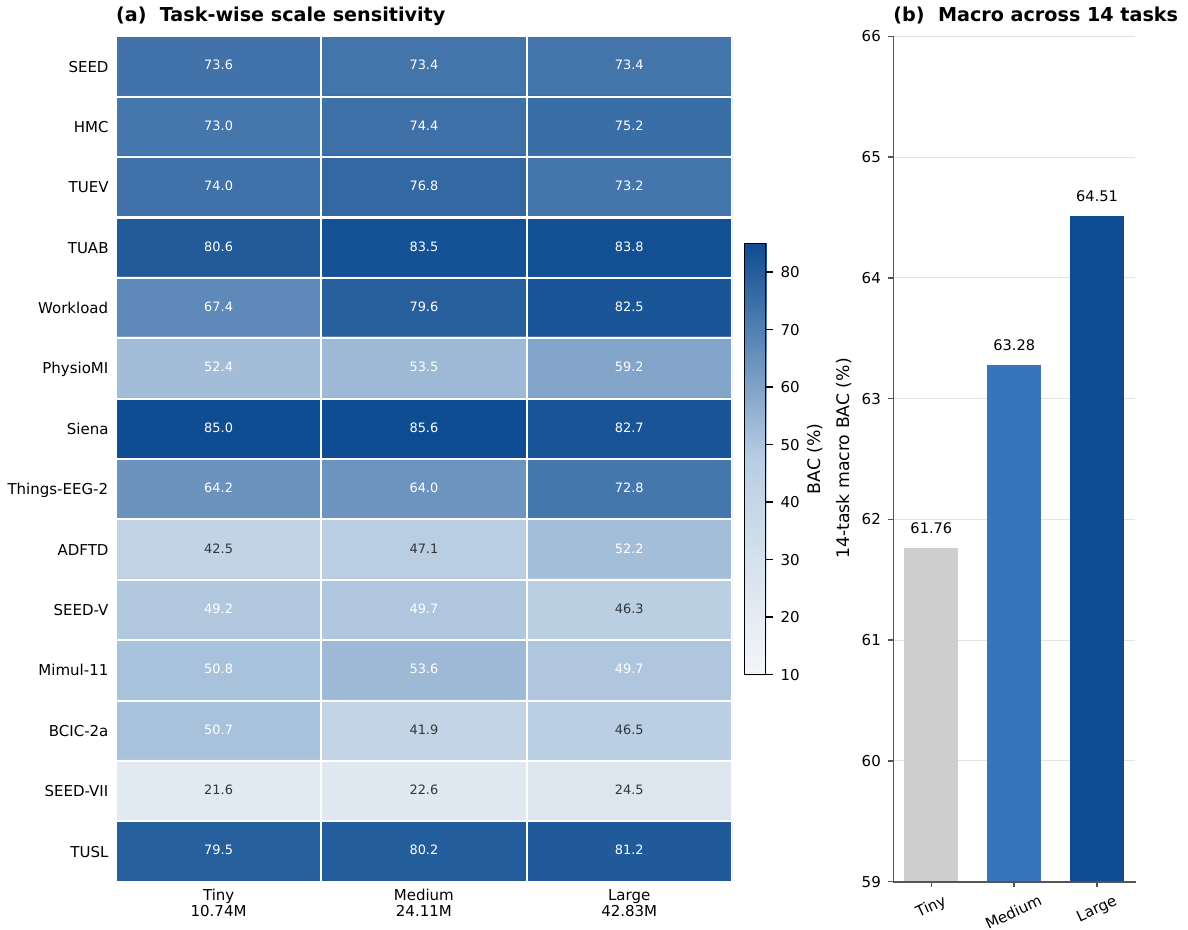}
\caption{\textbf{Model scale produces task-dependent changes.} The heatmap compares balanced accuracy for Tiny, Medium, and Large encoders across the displayed downstream tasks; the bar chart summarizes the accompanying aggregate trend. Larger capacity improves the aggregate score, but the best encoder size varies by dataset.}
\label{fig:model-scale}
\end{figure}

To examine the effect of encoder capacity, we pretrain Tiny (10.74M parameters) and Medium (24.11M) NSP models in addition to Large (42.83M), and evaluate the three scales with average pooling under single-task full fine-tuning (Figure~\ref{fig:model-scale}). The heatmap covers all 14 EEG-FM-Bench datasets, and the adjacent panel averages their BAC means.

\textbf{Aggregate trend favors capacity.} The reported macro BAC rises from 61.76 for Tiny through 63.28 for Medium to 64.51 for Large. The increase is modest relative to the change in parameter count.

\textbf{Task-level effects are heterogeneous.} Figure~\ref{fig:model-scale} shows that Large performs best on HMC, TUAB, Workload, PhysioMI, Things-EEG-2, ADFTD, SEED-VII, and TUSL. Medium is strongest on TUEV, Siena, SEED-V, and Mimul-11, while Tiny leads on SEED and BCIC-2a. The aggregate increase therefore does not imply a uniform task-wise ordering.

\textbf{Scope of the scale comparison.} These independently pretrained models examine capacity under the scale-study protocol. Their full-fine-tuning results are not directly comparable with the frozen, limited-label component ablation in Table~\ref{tab:ablation}. The increase in macro BAC therefore characterizes this scale comparison, rather than quantifying a trade-off between model size and objective design.

\subsection{Initialization controls}

\begin{figure}[H]
\centering
\includegraphics[width=0.75\textwidth]{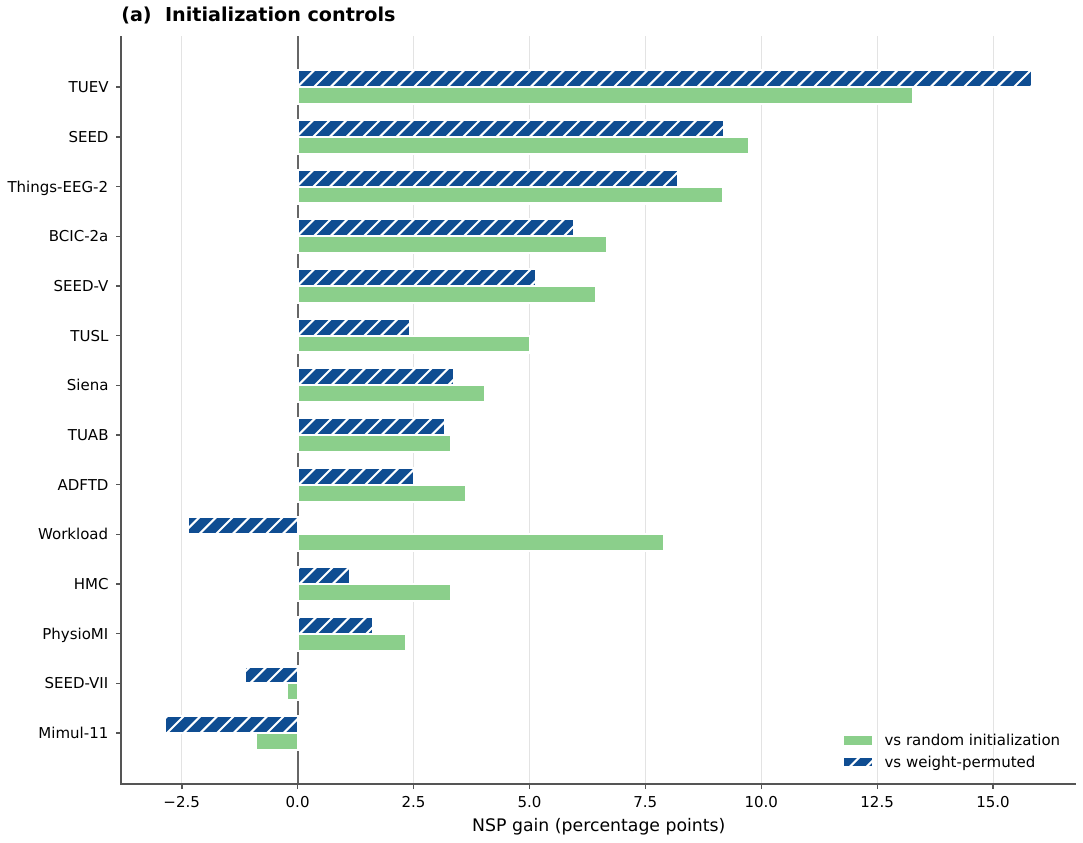}
\caption{\textbf{Initialization and parameter-organization controls.} Dataset-wise NSP gain relative to a randomly initialized encoder and to a weight-permuted encoder. Positive values indicate an advantage for the learned NSP parameters; differences between the two controls show that both parameter values and their arrangement can affect transfer.}
\label{fig:initialization-controls}
\end{figure}

Figure~\ref{fig:initialization-controls} compares the downstream gain of NSP with two controls that disrupt learned parameters: random initialization and weight permutation. Most datasets retain a positive gain over both controls, indicating that transfer depends on learned parameter values and, for many tasks, on their organization within the encoder. The magnitude varies substantially by dataset, and a small number of tasks do not improve over both controls.

\subsection{Effect of pretraining recording exclusion}
\label{app:recording-disjoint-control}

To assess how excluding downstream recordings changes transfer, we compare pretraining with the original TUEG manifest and with a recording-disjoint manifest that removes the evaluated downstream recordings (Table~\ref{tab:recording-disjoint-control}). Under the same downstream protocol, BAC changes from 83.83 to 82.87 on TUAB, 73.17 to 71.83 on TUEV, and 81.17 to 79.88 on TUSL. The respective decreases of 0.96, 1.34, and 1.29 percentage points indicate some sensitivity to recording exposure, while much of the observed transfer performance is retained after exclusion. This comparison is defined by recording-level membership in the pretraining manifest.

\begin{table}[H]
\centering
\caption{\textbf{Recording-disjoint TUEG pretraining control.} Balanced accuracy (\%, mean $\pm$ sample standard deviation) on the three TUH-derived downstream datasets. The recording-disjoint manifest excludes the evaluated downstream recordings from pretraining.}
\label{tab:recording-disjoint-control}
\small
\setlength{\tabcolsep}{8pt}
\renewcommand{\arraystretch}{1.08}
\begin{tabular}{lccc}
\toprule
Pretraining manifest & TUAB & TUEV & TUSL \\
\midrule
Original TUEG & 83.83$\pm$00.50 & 73.17$\pm$01.70 & 81.17$\pm$01.12 \\
Recording-disjoint TUEG & 82.87$\pm$00.94 & 71.83$\pm$01.53 & 79.88$\pm$00.76 \\
\bottomrule
\end{tabular}
\end{table}

\section{Mechanism Diagnostics}
\label{app:mechanism-diagnostics}

To complement downstream transfer results, we evaluate sensitivity to signal content and available context, and probe positional information retained by component variants. The interventions hold Target states fixed while changing Context input. They characterize prediction sensitivity and identity decodability, rather than demonstrate the elimination of position-only prediction or local interpolation.

\subsection{Context interventions and identity probes}

\begin{figure}[H]
\centering
\includegraphics[width=\textwidth]{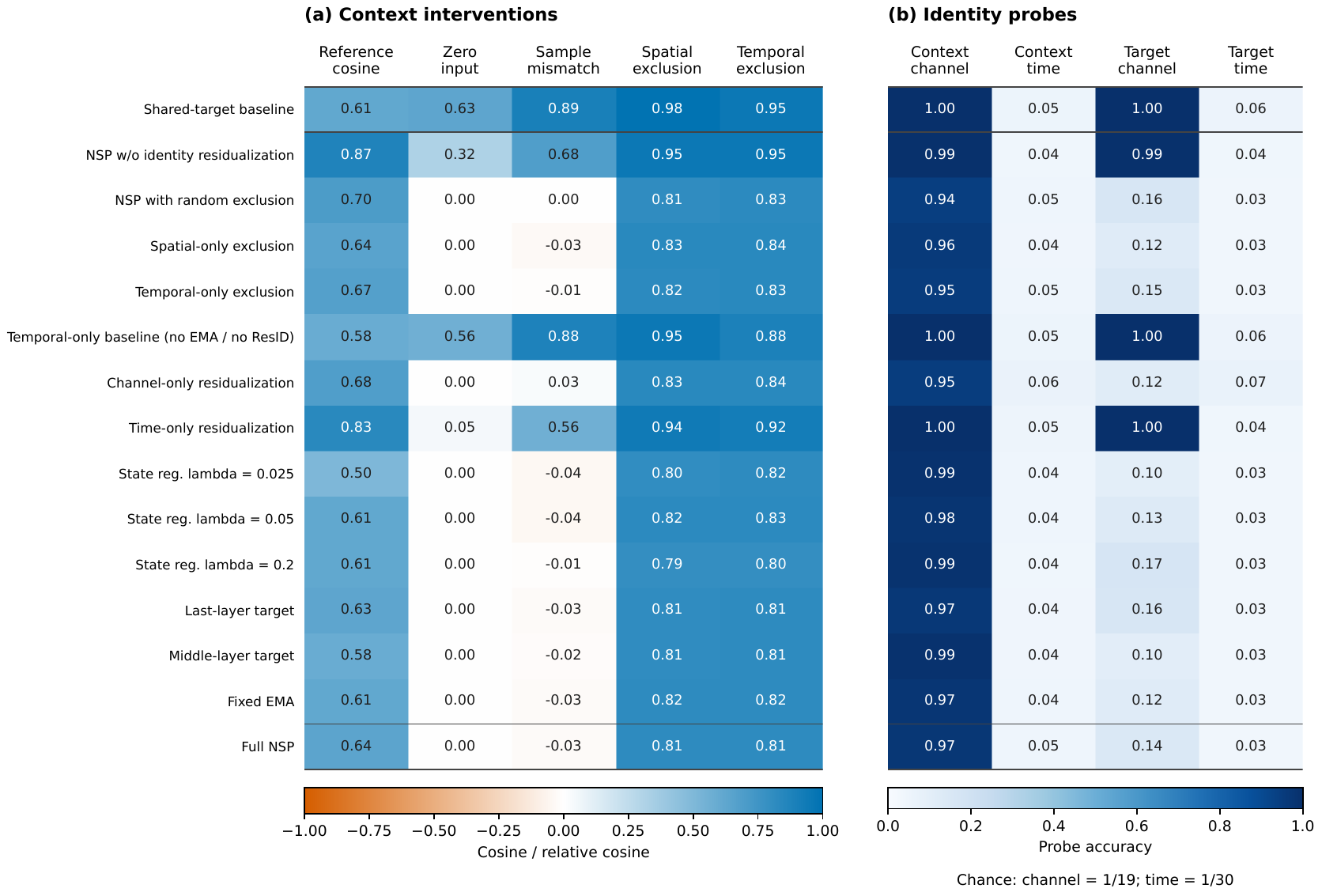}
\caption{\textbf{Context interventions and identity probes from the shared-target baseline to Full NSP.} The first and last rows show the baseline and Full NSP, with component variants between them. (a) Reference cosine uses target-only masking; intervention columns show relative cosine agreement with unchanged Targets. Sample mismatch operates across recordings in a batch. (b) Channel/time probe accuracy for Context and constructed Target states; chance is $1/19$ for channel and $1/30$ for time. Random exclusion includes residualization.}
\label{fig:mechanism-diagnostics}
\end{figure}

To distinguish signal sensitivity from identity decodability, we evaluate each displayed component variant with Context interventions and linear identity probes (Figure~\ref{fig:mechanism-diagnostics}). The reference Context masks target positions only; it is not an entirely unmasked input. For intervention $a$, relative cosine agreement is $R_a=C_a/C_{\mathrm{ref}}$, where $C_a$ averages prediction--target cosine over the masked target positions and $C_{\mathrm{ref}}$ is the same statistic for the reference Context. Targets are computed from the original recordings and held fixed. Zeroing removes EEG content, while sample mismatch cyclically shifts recordings along the batch dimension without permuting channels or time within a recording. Spatial and temporal interventions expand the Context mask around the same targets. Negative ratios indicate negative intervention cosine, not a negative fraction of retained information.

The probe panel reports channel and relative-time classification accuracy from full-input Context states and constructed Target states, with chance levels $1/19$ and $1/30$, respectively. Context channel accuracy remains high because residualization acts on Targets. Time-only residualization likewise leaves channel identity largely decodable; the temporal-only baseline has neither EMA nor residualization. The random-exclusion diagnostic includes residualization and corresponds to \emph{NSP with random exclusion}, not the no-residualization random control in Table~\ref{tab:ablation}. The shared-target baseline and Full NSP endpoints are evaluated using the same diagnostic protocol: 512 evaluation segments for interventions and separate sets of 256 segments for probe training and testing, with 64 sampled tokens per segment. Sample-mismatch relative cosine changes from 0.894 to $-0.028$, while Target channel-probe accuracy changes from 100.00\% to 14.43\%. The latter remains above chance, supporting reduced, not eliminated, identity decodability. These are single-checkpoint diagnostics rather than estimates of variability across pretraining runs.

\subsection{Component-transfer overview}

\begin{figure}[H]
\centering
\includegraphics[width=\textwidth]{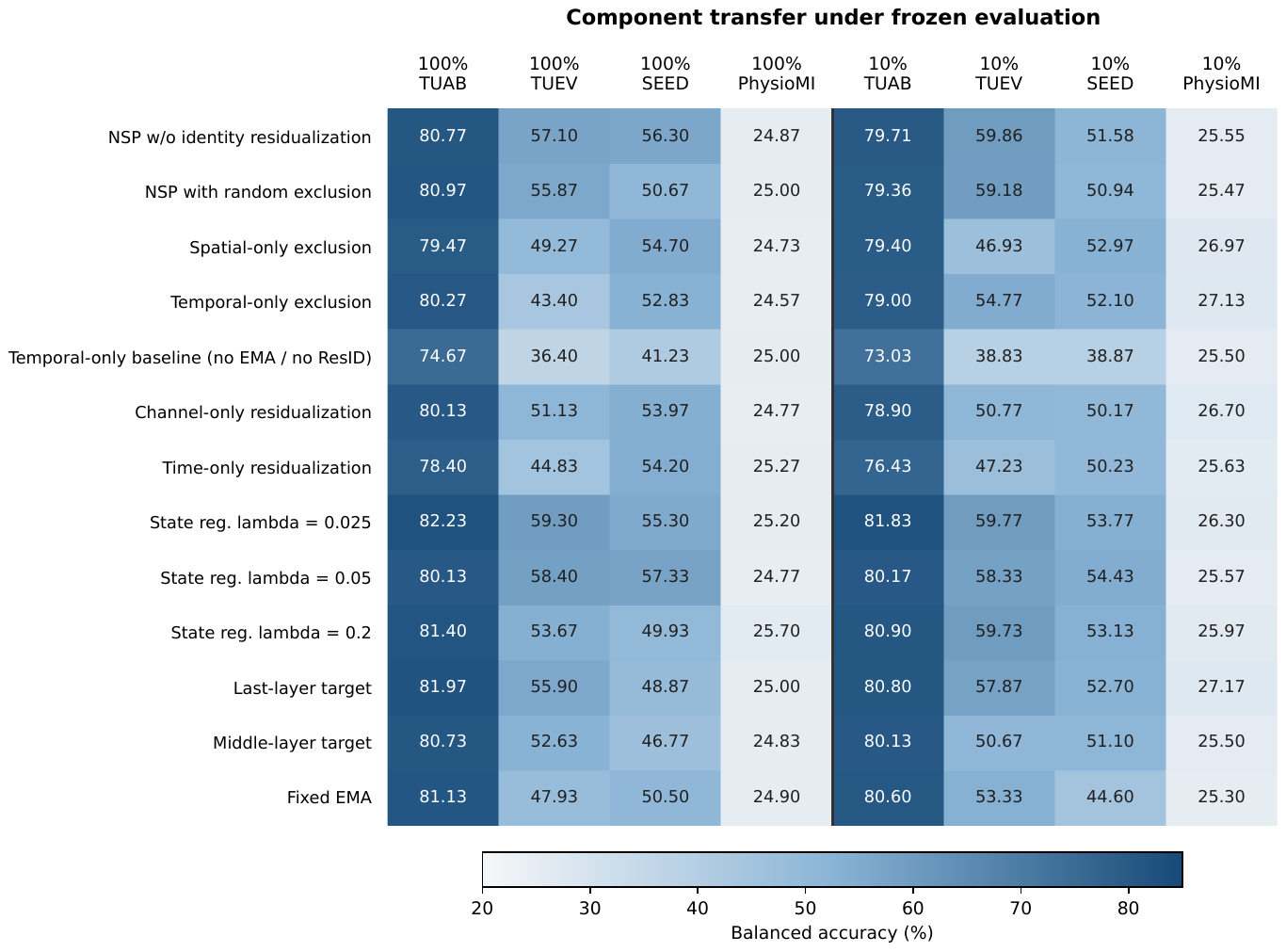}
\caption{\textbf{Complete component-transfer overview.} Frozen single-task BAC for guard, residualization, regularization, target-layer, and EMA variants on TUAB, TUEV, SEED, and PhysioMI. The two column groups use 100\% and 10\% labels, respectively, and show strongly task-dependent component effects.}
\label{fig:component-transfer}
\end{figure}

Figure~\ref{fig:component-transfer} reports frozen single-task BAC with 10\% and 100\% labels for guard, residualization, regularization, target-layer, and EMA variants on TUAB, TUEV, SEED, and PhysioMI. The side-by-side label budgets show the magnitude and direction of each component's effect by dataset.

\section{Additional Visualizations}
\label{app:visualizations}

\subsection{t-SNE on additional tasks}

\begin{figure}[H]
\centering
\includegraphics[width=\textwidth]{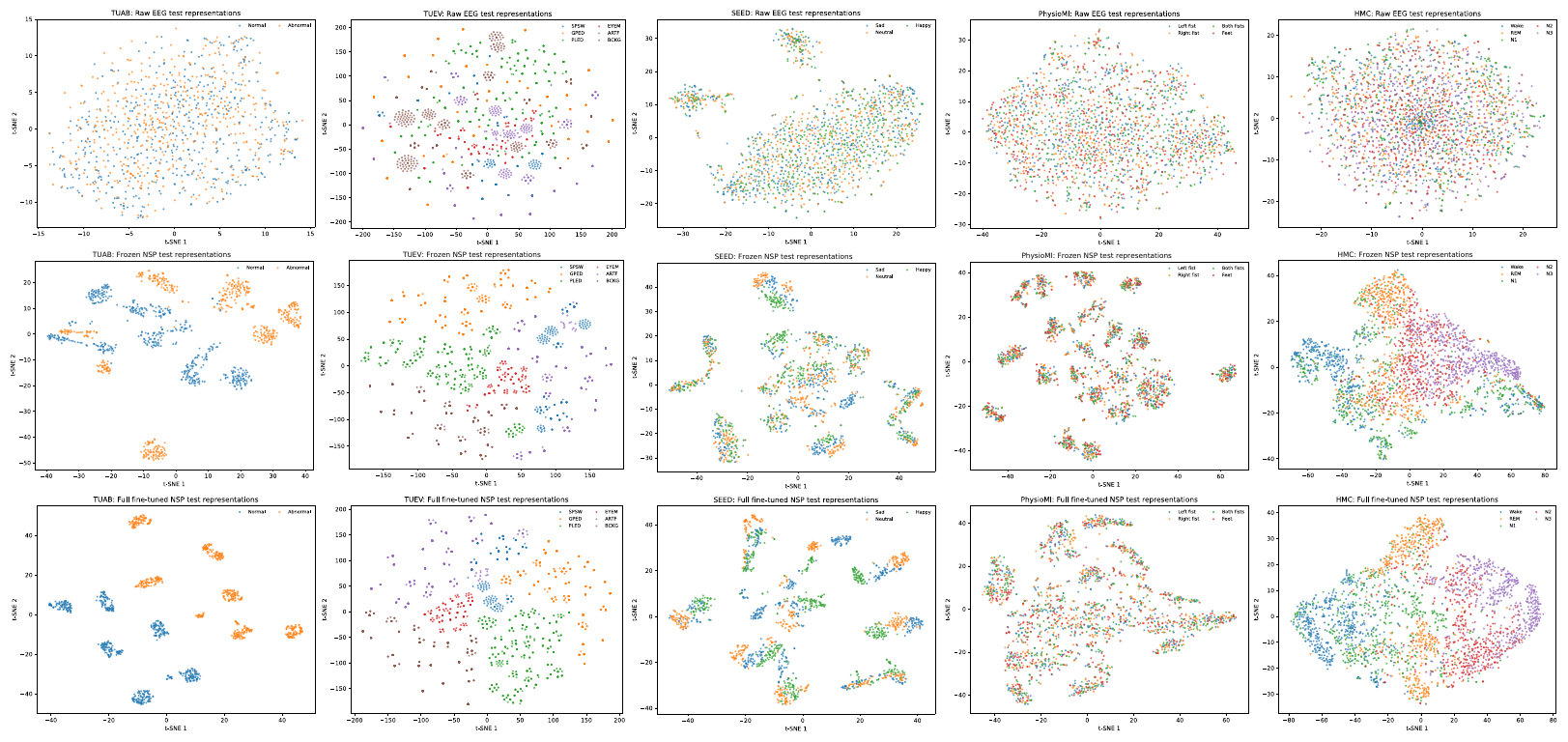}
\caption{\textbf{Representation geometry across five downstream tasks.} Columns show TUAB, TUEV, SEED, PhysioMI, and HMC; rows show raw EEG, frozen NSP, and fully adapted NSP representations.}
\label{fig:tsne-additional}
\end{figure}

\begin{figure}[H]
\centering
\includegraphics[width=\textwidth]{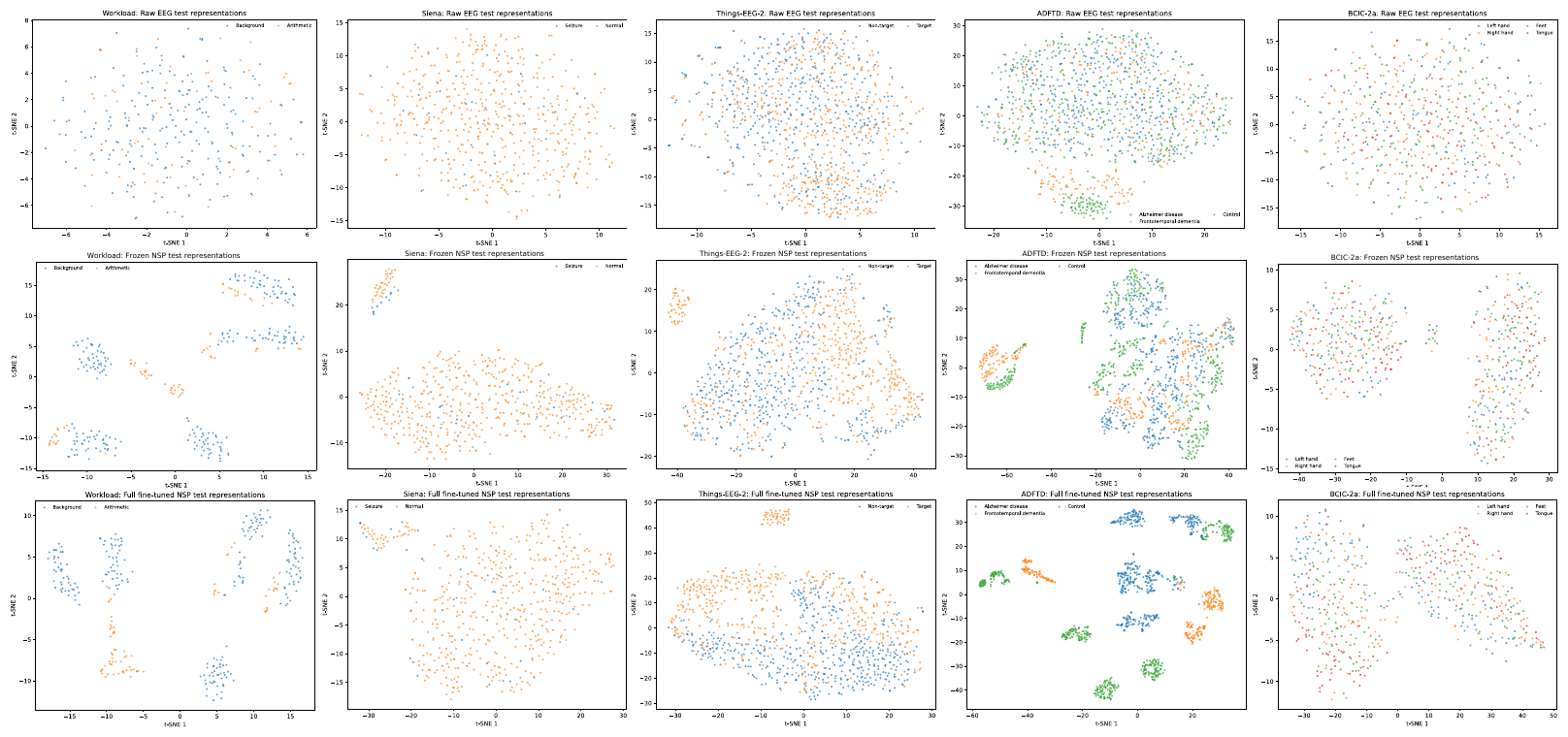}
\caption{\textbf{Representation geometry across additional paradigms.} Columns show Workload, Siena, Things-EEG-2, ADFTD, and BCIC-2a; rows show raw EEG, frozen NSP, and fully adapted NSP representations. Each projection is fit independently and illustrates task-specific changes in class geometry.}
\label{fig:tsne-additional2}
\end{figure}

Beyond the HMC example in the main paper, Figures~\ref{fig:tsne-additional} and~\ref{fig:tsne-additional2} compare raw EEG, frozen NSP, and fully adapted NSP representations across ten downstream tasks using t-SNE~\citep{vandermaaten2008tsne}.

\subsection{Grad-CAM topology maps}

\begin{figure}[H]
\centering
\includegraphics[width=0.95\textwidth]{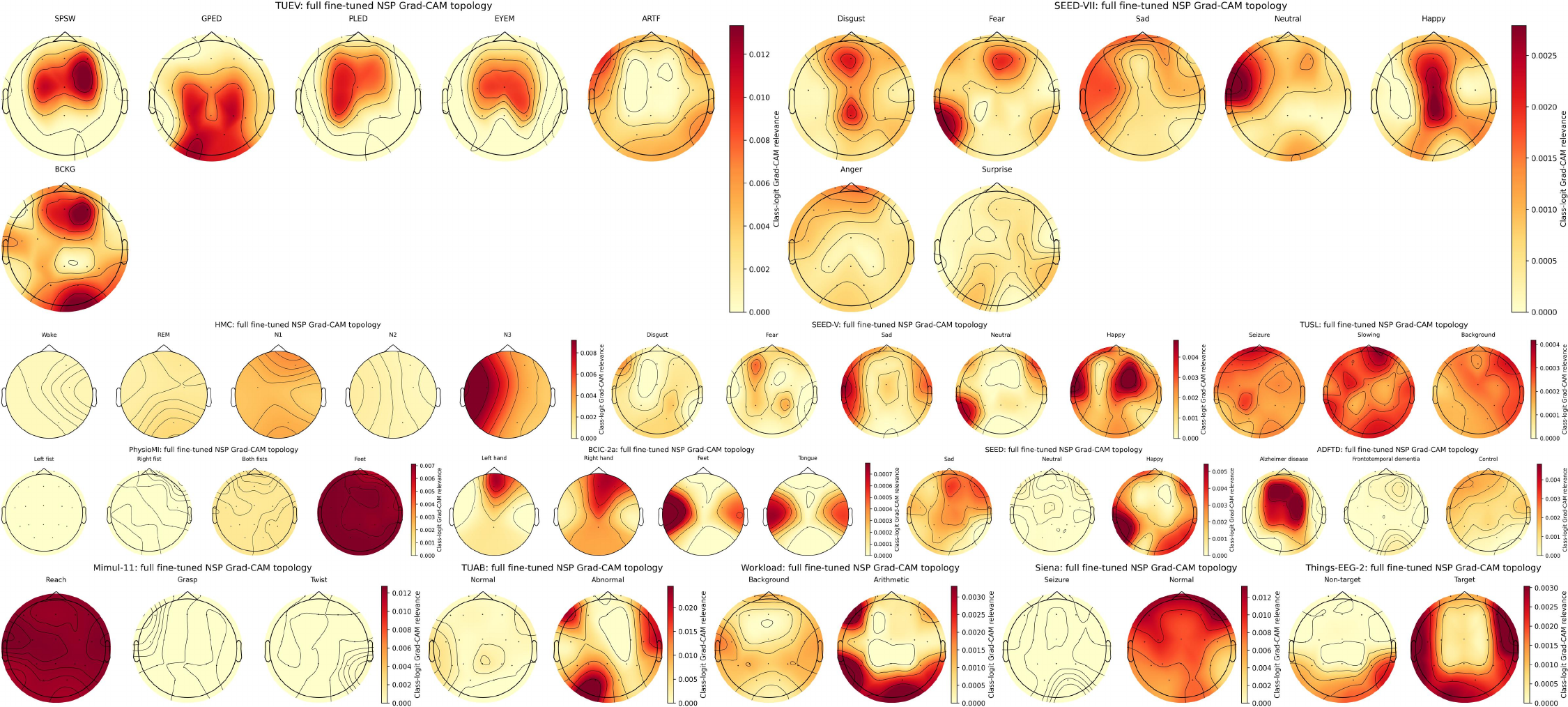}
\caption{\textbf{Class-conditioned Grad-CAM topology maps across 14 EEG-FM-Bench datasets.} Each dataset group contains one topographic attribution map per class, obtained after summing relevance over time. The maps are descriptive and task conditioned; quantitative transfer claims rest on the controlled benchmark evaluations.}
\label{fig:gradcam-topology}
\end{figure}

We compute class-conditioned Grad-CAM attribution~\citep{selvaraju2017gradcam} for the 14 EEG-FM-Bench datasets, each summed over time to a per-channel map. Figure~\ref{fig:gradcam-topology} shows spatially structured, task-conditioned patterns spanning clinical (TUAB, TUEV, TUSL), sleep (HMC), seizure (Siena), motor-imagery (PhysioMI, BCIC-2a), emotion (SEED), workload (Workload), and visual (Things-EEG-2) tasks. These maps are descriptive and do not identify physiological generators or functional connectivity.

\subsection{Interchannel correlation structure}

\begin{figure}[H]
\centering
\includegraphics[width=\textwidth]{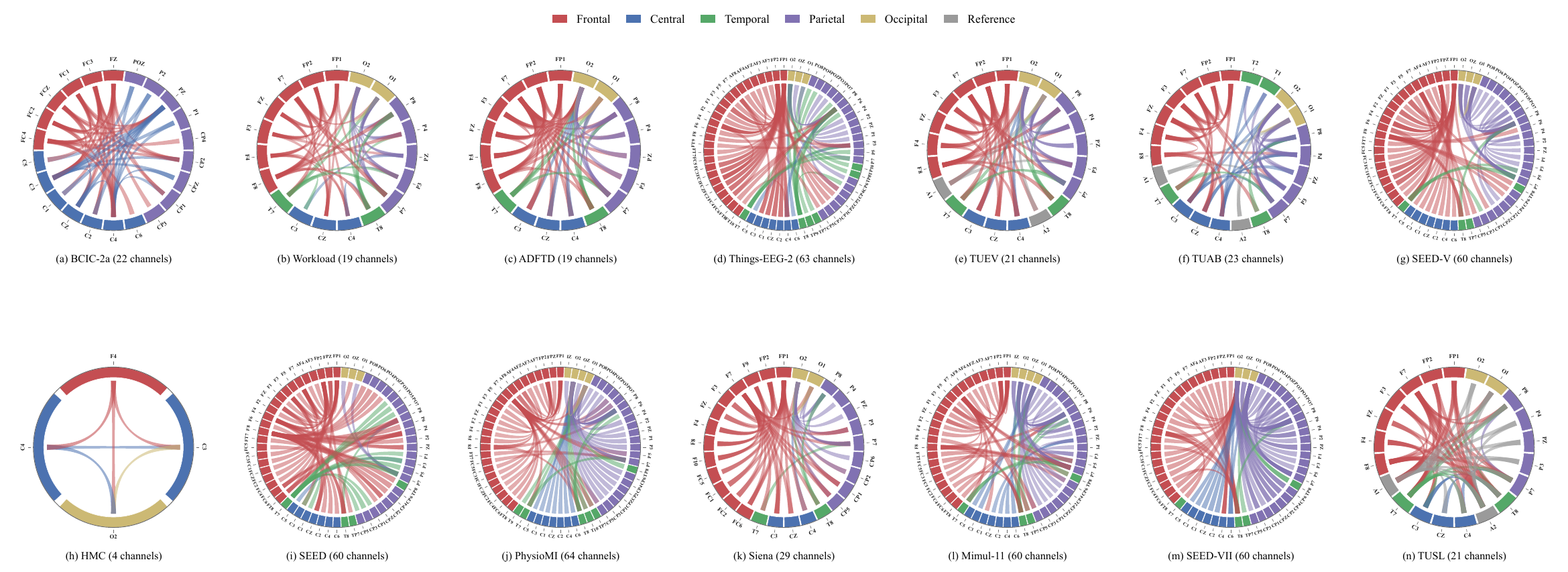}
\caption{\textbf{Interchannel correlation diagrams across 14 EEG-FM-Bench datasets.} Source-montage channel labels are associated with model channels through the benchmark adapter mapping. Ribbon width encodes selected attention-derived interchannel correlation scores, and color groups channels by anatomical region.}
\label{fig:interchannel-chord}
\end{figure}

To examine task-conditioned interchannel correlations, we display attention-derived correlation scores from task-adapted NSP as chord diagrams across the 14 EEG-FM-Bench datasets (Figure~\ref{fig:interchannel-chord}). Channel labels follow each dataset's source montage and the benchmark adapter's channel mapping. Ribbon width represents the selected interchannel correlation score, and channels in the same anatomical region share a color.

%% file: tables/app_pretraining_details.tex
\section{NSP Pretraining Details}
\label{app:pretraining_details}

\subsection{Objective Justification and Training Algorithm}

Predicting latent targets avoids forcing the encoder to reproduce every
high-variance detail of the EEG waveform, but it does not determine which
information remains in the target. On scalp EEG, channel and relative-time
identity provide stable position cues, while volume-conducted neighboring
electrodes provide a short path for local interpolation. NSP therefore modifies
both sides of the prediction problem: identity residualization changes
\emph{what} is predicted, topology-separated context changes \emph{which
evidence} is visible, and the state regularizer prevents the Context Encoder
from satisfying the objective with a degenerate representation. The full
mathematical construction is given in \Secref{sec:method}; Algorithm
\ref{alg:nsp_pretraining} records the executable order of operations.
Here, \textsc{TargetConstruct} denotes selected-layer aggregation,
standardization, and identity residualization in \Secref{sec:method};
$\operatorname{nrm}$ is normalization to unit $\ell_2$ norm along the feature
dimension. We use $i=(b,p,c)$ for a flattened token index and
$\mathcal M=\{(b,p,c):(p,c)\in\mathcal M_b\}$, so
$|\mathcal M|=\sum_b|\mathcal M_b|$. The labels $\mathbf z_i$, $\mathbf r_i$,
$\mathbf h_i$, and $\mathbf p_i$ in Figure~\ref{fig:method} use this convention.
Here $f_\eta(\mathbf X;\mathcal S)=\mathbf H_\eta^{(N_{\mathrm{blk}})}(\mathbf X;\mathcal S)$;
Target construction instead uses the selected intermediate states of $f_\phi$.
Masking replaces signal-dependent patch content before positional embeddings are added.

\begin{algorithm}[H]
\caption{Neural State Prediction pretraining.}
\label{alg:nsp_pretraining}
\small
\begin{algorithmic}[1]
\Require EEG minibatch $\mathbf X$; Context Encoder $f_\theta$; Target Encoder
$f_\phi$; predictor $g_\psi$
\State $\{\mathcal M_b\}_{b=1}^B \gets \Call{TopologyMask}{\mathbf X}$
\State $\mathcal G_b \gets \Call{ExpandGuard}{\mathcal M_b}$; $\mathcal S_b\gets\mathcal M_b\cup\mathcal G_b$ for each $b$
\State $\mathbf h \gets f_\theta(\mathbf X;\mathcal S)$ \Comment{Mask patch content, not positional cues}
\State $\mathbf r \gets \operatorname{sg}\!\left(
  \Call{TargetConstruct}{\{\mathbf H_\phi^{(\ell)}(\mathbf X;\varnothing)\}_{\ell\in\mathcal I_T}}\right)$
\State $\mathbf p \gets g_\psi(\mathbf h)$ \Comment{Predictor on Context states only}
\State $\displaystyle \mathcal L_{\mathrm{pred}} \gets
  \frac{1}{|\mathcal M|}\sum_{i\in\mathcal M}
  \left[1-\left\langle
  \operatorname{nrm}(\mathbf p_i),
  \operatorname{nrm}(\mathbf r_i)\right\rangle\right]$
\State $\mathbf h^{\mathrm{full}} \gets f_\theta(\mathbf X;\varnothing)$ \Comment{Shared Context parameters}
\State $\mathcal L_{\mathrm{reg}} \gets
  \Call{MomentRegularizer}{\mathbf h^{\mathrm{full}}}$ \Comment{Eq.~(\ref{eq:sigreg})}
\State $\mathcal L_{\mathrm{total}} \gets \mathcal L_{\mathrm{pred}} +
  \lambda_{\mathrm{reg}}\mathcal L_{\mathrm{reg}}$
\State $(\theta,\psi) \gets
  \Call{AdamWStep}{\theta,\psi,\nabla_{\theta,\psi}\mathcal L_{\mathrm{total}}}$
\State $\phi \gets \tau_s\phi+(1-\tau_s)\theta$ \Comment{EMA after the Context update}
\end{algorithmic}
\end{algorithm}

\subsection{Hyperparameter Configuration}

\begin{table}[H]
\centering
\caption{Default architecture and pretraining configuration of NSP.}
\label{tab:app_pretraining_architecture}
\footnotesize
\setlength{\tabcolsep}{4pt}
\renewcommand{\arraystretch}{1.04}
\begin{tabular}{@{}ll@{}}
\toprule
Hyperparameter & Configuration \\
\midrule
\multicolumn{2}{@{}l}{\textbf{Input \& Tokenization}} \\
\addlinespace[1pt]
Segment duration & 6 s \\
Sampling rate & 200 Hz \\
EEG channels & 19 \\
Patch duration & 0.2 s \\
\addlinespace[2pt]
\multicolumn{2}{@{}l}{\textbf{Context Encoder \& Predictor}} \\
\addlinespace[1pt]
Encoder depth & 8 \\
Hidden dimension & 512 \\
Attention heads & 8 \\
FFN dimension & 2,048 \\
Dropout & 0.1 \\
Temporal attention & Bidirectional \\
Predictor hidden dimension & 768 \\
Predictor output dimension & 512 \\
\addlinespace[2pt]
\multicolumn{2}{@{}l}{\textbf{Target Encoder}} \\
\addlinespace[1pt]
Target layers & $\{5,6,7\}$ (zero-based) \\
EMA momentum & $0.996 \rightarrow 0.9999$ \\
EMA schedule & Cosine \\
\addlinespace[2pt]
\multicolumn{2}{@{}l}{\textbf{Target Masking and Context Exclusion}} \\
\addlinespace[1pt]
Target ratio & 0.50 \\
Temporal target span & 2--10 patches (0.4--2.0 s) \\
Spatial target span & 3--7 coordinate-near channels \\
Temporal guard radius & 1 patch \\
Spatial guard & 2 nearest channels per target channel \\
\addlinespace[2pt]
\multicolumn{2}{@{}l}{\textbf{Moment-based State Regularization}} \\
\addlinespace[1pt]
Projection directions & 64 \\
Regularization weight & $\lambda_{\mathrm{reg}}=0.1$ \\
\addlinespace[2pt]
\multicolumn{2}{@{}l}{\textbf{Model}} \\
\addlinespace[1pt]
Trainable Context-side parameters & 42.83M \\
\bottomrule
\end{tabular}
\end{table}

\begin{table}[H]
\centering
\caption{Pretraining and optimization hyperparameters of NSP.}
\label{tab:app_pretraining_optimization}
\footnotesize
\setlength{\tabcolsep}{4pt}
\renewcommand{\arraystretch}{1.04}
\begin{tabular}{@{}ll@{}}
\toprule
Hyperparameter & Configuration \\
\midrule
Pretraining corpus & TUEG~\citep{obeid2016tueg} \\
Pretraining examples & 2.2M segmented examples \\
Optimizer & AdamW~\citep{loshchilov2019adamw} \\
Learning rate & $1\times10^{-4}$ \\
Adam $(\beta_1,\beta_2)$ & $(0.9,0.999)$ \\
Weight decay & 0.05 \\
Training steps & 50,000 \\
Warmup steps & 5,000 \\
Warmup initial factor & 0.1 \\
Learning-rate schedule & Linear warmup followed by cosine annealing \\
Batch size per GPU & 48 \\
Number of GPUs & 8 NVIDIA H100 GPUs \\
Global batch size & 384 \\
Gradient accumulation & 1 \\
Gradient clipping & Global norm 1.0 \\
Mixed precision & Enabled \\
Distributed backend & NVIDIA Collective Communications Library (NCCL) \\
Distributed sampling & Rank-sharded, shuffled, drop-last \\
Pretraining random seeds & 45, 46, 47 \\
\bottomrule
\end{tabular}
\end{table}

\subsection{Training Diagnostics}

\subsubsection{Diagnostic Experiments}

We monitor optimization and representation health jointly. The diagnostic
sequence compares a temporal shared-target predictor, a spatiotemporal
shared-target predictor, an EMA Target Encoder, identity-residualized targets,
Full NSP, and NSP without state regularization. These names describe the
actual intervention and are used consistently throughout the paper. All
configurations share the Large encoder, 50,000 optimization steps, and the same
distributed input stream; differences in raw prediction loss are interpreted
only when their target and input-mask definitions match.

The control without state regularization reaches a Context-state mean cosine
of 0.9422 by step 4,000 and 0.9979 at the endpoint, while its effective rank
falls to 113.6. Full NSP instead retains a rank of 470.9 with near-zero
mean cosine. In this control, lower prediction loss coincides with highly
aligned Context states and reduced effective rank. We therefore assess
pretraining with downstream transfer and representation-health measures
alongside loss.

\begin{figure}[H]
  \centering
  \includegraphics[width=0.92\textwidth]{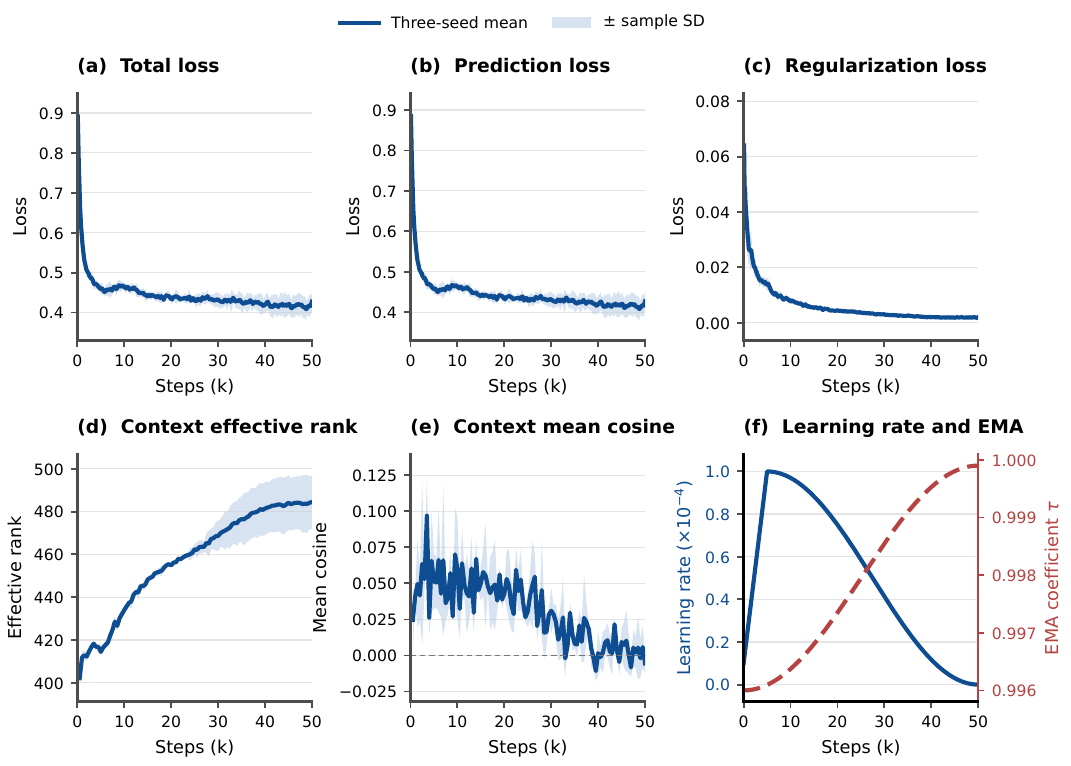}
  \caption{\textbf{NSP pretraining diagnostics.} Lines show the across-run mean and shaded regions show sample SD for optimization losses and representation-health statistics. The final panel displays the learning-rate and EMA schedules.}
  \label{fig:app_pretraining_curves}
\end{figure}

Figure~\ref{fig:app_pretraining_curves} reports the optimization trajectory of
Full NSP across three independent pretraining runs. Total and prediction
loss decrease rapidly during warmup and then continue to decline gradually,
while the regularization term approaches zero. At the same time, Context
effective rank rises and mean cosine remains near zero, so the lower objective
is accompanied by a non-collapsed representation rather than constant states.

\subsection{Downstream Benchmark Requirements}

NeuralBench-EEG-Core~\citep{neuralbench} specifies task-level preprocessing, data splits, adaptation settings, and validation metrics. We evaluate 16 selected datasets with single-task adaptation: 14 classification tasks contribute to the BAC macro, while two language tasks use Top-5 retrieval accuracy and are reported separately. EEG-FM-Bench~\citep{eegfmbench} specifies preprocessing and model adapters for 14 datasets and compares single-task full fine-tuning, multi-task full fine-tuning, frozen-backbone transfer, and LoRA. We follow each benchmark's released task definitions and report its results within its own protocol.

%% file: tables/app_complete_neuralbench_results.tex

\subsubsection{Clinical and Neurological Disease}

Mumtaz2018 contains 14,683 five-second windows from 64 participants for major-depression diagnosis~\citep{mumtaz2018mdd}; Singh2021 contains 39,543 windows from 129 participants for Parkinson's-disease classification~\citep{singh2021parkinsons}; and Albrecht2019 contains 24,928 windows from 77 participants for schizophrenia classification~\citep{albrecht2019schizo}. CHB-MIT (the Dan2023 benchmark task) is the larger seizure-detection task, with 707,683 windows from 23 participants~\citep{dan2023chbmit}. All four tasks are binary and report BAC, AUROC, and AUCPR. NSP obtains the highest BAC on the three diagnostic datasets and remains competitive on CHB-MIT, where CBraMod leads BAC while NSP ranks second in AUROC and AUCPR. On Mumtaz2018 and Albrecht2019, NSP also has the highest AUROC; its AUCPR ranks first and second, respectively.

\begin{table}[H]
\centering
\caption{NeuralBench clinical and neurological-disease results (\%, mean $\pm$ sample SD). Red and blue mark the best and second-best mean for each metric.}
\label{tab:neuralbench_clinical_complete}
\vskip -0.04in
\fontsize{7.0}{7.6}\selectfont
\setlength{\tabcolsep}{3.0pt}
\renewcommand{\arraystretch}{1.04}
\rowcolors{3}{black!3}{white}
\begin{tabular}{@{}lccc@{\hspace{8pt}}lccc@{}}
\toprule
\multicolumn{4}{c}{\textbf{Mumtaz2018}} & \multicolumn{4}{c}{\textbf{Singh2021}} \\
\cmidrule(lr){1-4}\cmidrule(lr){5-8}
Model & BAC & AUROC & AUCPR & Model & BAC & AUROC & AUCPR \\
\midrule
BENDR   & 86.22$\pm$00.47 & 92.20$\pm$00.26 & 91.17$\pm$00.27 & BENDR   & 65.82$\pm$05.62 & 70.83$\pm$06.33 & 85.33$\pm$03.73 \\
BIOT    & 79.11$\pm$03.39 & 86.20$\pm$01.13 & 82.88$\pm$05.00 & BIOT    & 74.18$\pm$04.16 & 81.60$\pm$04.45 & 88.86$\pm$02.51 \\
    LaBraM  & 85.61$\pm$03.43 & 88.33$\pm$05.14 & 87.57$\pm$03.82 & LaBraM  & \textcolor{blue}{\textbf{76.64$\pm$04.06}} & \textcolor{red}{\textbf{87.32$\pm$04.69}} & \textcolor{red}{\textbf{94.19$\pm$02.42}} \\
CBraMod & 86.04$\pm$02.48 & 92.57$\pm$02.82 & 93.63$\pm$01.06 & CBraMod & 74.37$\pm$02.38 & 83.18$\pm$04.65 & 91.94$\pm$02.19 \\
    LUNA    & 85.42$\pm$00.98 & \textcolor{blue}{\textbf{94.04$\pm$00.69}} & \textcolor{blue}{\textbf{94.43$\pm$01.02}} & LUNA    & 64.95$\pm$00.49 & 73.50$\pm$01.11 & 87.02$\pm$00.59 \\
    REVE    & \textcolor{blue}{\textbf{87.33$\pm$01.44}} & 90.72$\pm$01.45 & 86.62$\pm$03.03 & REVE & 72.58$\pm$01.72 & 82.46$\pm$03.04 & \textcolor{blue}{\textbf{92.56$\pm$01.81}} \\
    NSP     & \textcolor{red}{\textbf{90.43$\pm$06.40}} & \textcolor{red}{\textbf{96.12$\pm$06.78}} & \textcolor{red}{\textbf{94.81$\pm$06.54}} & NSP & \textcolor{red}{\textbf{76.67$\pm$01.70}} & \textcolor{blue}{\textbf{84.21$\pm$01.87}} & 91.26$\pm$02.03 \\
\midrule
\multicolumn{4}{c}{\textbf{Albrecht2019}} & \multicolumn{4}{c}{\textbf{CHB-MIT}} \\
\cmidrule(lr){1-4}\cmidrule(lr){5-8}
Model & BAC & AUROC & AUCPR & Model & BAC & AUROC & AUCPR \\
\midrule
BENDR   & 57.17$\pm$00.87 & 60.31$\pm$01.40 & 69.42$\pm$00.23 & BENDR   & 78.97$\pm$01.04 & 88.53$\pm$01.77 & 99.94$\pm$00.01 \\
BIOT    & 56.18$\pm$01.50 & 57.52$\pm$01.38 & 64.19$\pm$02.16 & BIOT    & 73.58$\pm$05.20 & 80.41$\pm$02.16 & 99.81$\pm$00.04 \\
    LaBraM  & 63.62$\pm$01.93 & 66.26$\pm$04.42 & 75.04$\pm$07.83 & LaBraM  & \textcolor{blue}{\textbf{82.09$\pm$00.75}} & \textcolor{red}{\textbf{93.07$\pm$00.52}} & \textcolor{red}{\textbf{99.97$\pm$00.00}} \\
CBraMod & 57.05$\pm$03.12 & 60.03$\pm$04.59 & 67.79$\pm$04.87 & CBraMod & \textcolor{red}{\textbf{82.89$\pm$01.58}} & 90.84$\pm$01.72 & 99.93$\pm$00.03 \\
    LUNA    & \textcolor{blue}{\textbf{66.75$\pm$02.18}} & \textcolor{blue}{\textbf{71.42$\pm$02.22}} & \textcolor{red}{\textbf{81.81$\pm$02.04}} & LUNA & 70.46$\pm$02.44 & 80.50$\pm$06.88 & 99.87$\pm$00.08 \\
REVE    & 49.21$\pm$10.36 & 50.00$\pm$16.05 & 63.24$\pm$12.36 & REVE & 76.51$\pm$03.73 & 82.74$\pm$07.09 & 99.83$\pm$00.11 \\
    NSP     & \textcolor{red}{\textbf{68.43$\pm$02.71}} & \textcolor{red}{\textbf{73.06$\pm$02.90}} & \textcolor{blue}{\textbf{79.44$\pm$03.15}} & NSP & 81.10$\pm$01.39 & \textcolor{blue}{\textbf{91.64$\pm$01.57}} & \textcolor{blue}{\textbf{99.95$\pm$00.02}} \\
\bottomrule
\end{tabular}
\rowcolors{2}{}{}
\end{table}

\subsubsection{BCI and Sensorimotor Decoding}

This group spans four multiclass BCI paradigms. Thielen2021 provides 45,000 c-VEP windows from 30 participants~\citep{thielen2021cvep}; Scherer2015 contains 3,550 five-class motor-imagery windows from nine participants~\citep{scherer2015imagery}; Srisrisawang2024 contains 19,157 sixteen-class motor-execution windows from 20 participants~\citep{srisrisawang2024reaching}; and Wang2017 contains 8,160 forty-class SSVEP windows from 34 participants~\citep{wang2017ssvep}. NSP leads BAC, weighted F1, and $\kappa$ on Thielen2021, Scherer2015, and Wang2017. On Srisrisawang2024, REVE is strongest and NSP is second across all three metrics.

\begin{table}[H]
\centering
\caption{NeuralBench BCI and sensorimotor results (\%, mean $\pm$ sample SD). Red and blue mark the best and second-best mean for each metric.}
\label{tab:neuralbench_bci_complete}
\vskip -0.04in
\fontsize{7.0}{7.6}\selectfont
\setlength{\tabcolsep}{3.0pt}
\renewcommand{\arraystretch}{1.04}
\rowcolors{3}{black!3}{white}
\begin{tabular}{@{}lccc@{\hspace{8pt}}lccc@{}}
\toprule
\multicolumn{4}{c}{\textbf{Thielen2021}} & \multicolumn{4}{c}{\textbf{Scherer2015}} \\
\cmidrule(lr){1-4}\cmidrule(lr){5-8}
Model & BAC & F1 & $\kappa$ & Model & BAC & F1 & $\kappa$ \\
\midrule
BENDR   & 4.96$\pm$00.24 & 1.64$\pm$00.37 & -0.04$\pm$00.26 & BENDR & 21.62$\pm$00.66 & 19.11$\pm$01.61 & 2.02$\pm$00.83 \\
    BIOT    & 4.80$\pm$00.25 & 3.84$\pm$00.04 & -0.21$\pm$00.26 & BIOT & \textcolor{blue}{\textbf{26.48$\pm$02.48}} & \textcolor{blue}{\textbf{25.69$\pm$03.03}} & \textcolor{blue}{\textbf{8.10$\pm$03.10}} \\
LaBraM  & 86.39$\pm$00.74 & 86.48$\pm$00.78 & 85.67$\pm$00.78 & LaBraM & 21.83$\pm$00.17 & 20.34$\pm$00.83 & 2.29$\pm$00.21 \\
CBraMod & 86.37$\pm$01.33 & 86.38$\pm$01.31 & 85.66$\pm$01.40 & CBraMod & 21.22$\pm$01.80 & 15.89$\pm$05.21 & 1.52$\pm$02.25 \\
    LUNA    & \textcolor{blue}{\textbf{87.06$\pm$00.28}} & \textcolor{blue}{\textbf{87.05$\pm$00.28}} & \textcolor{blue}{\textbf{86.38$\pm$00.30}} & LUNA & 20.95$\pm$01.09 & 14.21$\pm$06.01 & 1.19$\pm$01.37 \\
REVE    & 80.51$\pm$01.81 & 80.55$\pm$01.86 & 79.49$\pm$01.91 & REVE & 26.06$\pm$01.07 & 24.02$\pm$01.88 & 7.57$\pm$01.33 \\
    NSP     & \textcolor{red}{\textbf{88.83$\pm$03.37}} & \textcolor{red}{\textbf{88.74$\pm$03.36}} & \textcolor{red}{\textbf{88.05$\pm$03.33}} & NSP & \textcolor{red}{\textbf{32.00$\pm$00.55}} & \textcolor{red}{\textbf{30.13$\pm$00.52}} & \textcolor{red}{\textbf{9.74$\pm$00.79}} \\
\midrule
\multicolumn{4}{c}{\textbf{Srisrisawang2024}} & \multicolumn{4}{c}{\textbf{Wang2017}} \\
\cmidrule(lr){1-4}\cmidrule(lr){5-8}
Model & BAC & F1 & $\kappa$ & Model & BAC & F1 & $\kappa$ \\
\midrule
BENDR   & 34.56$\pm$00.96 & 32.79$\pm$01.28 & 30.21$\pm$01.01 & BENDR & 13.31$\pm$05.51 & 7.13$\pm$04.10 & 11.09$\pm$05.66 \\
BIOT    & 18.00$\pm$01.78 & 16.24$\pm$02.13 & 12.53$\pm$01.89 & BIOT & 58.67$\pm$01.85 & 58.33$\pm$02.05 & 57.61$\pm$01.89 \\
    LaBraM  & 52.94$\pm$00.88 & 53.16$\pm$00.70 & 49.87$\pm$00.92 & LaBraM & \textcolor{blue}{\textbf{96.31$\pm$00.45}} & \textcolor{blue}{\textbf{96.31$\pm$00.44}} & \textcolor{blue}{\textbf{96.21$\pm$00.46}} \\
CBraMod & 49.41$\pm$00.99 & 48.82$\pm$01.23 & 46.08$\pm$01.06 & CBraMod & 85.85$\pm$07.27 & 85.62$\pm$07.59 & 85.49$\pm$07.45 \\
LUNA    & 49.92$\pm$00.72 & 49.86$\pm$01.02 & 46.63$\pm$00.77 & LUNA & 13.08$\pm$00.88 & 11.22$\pm$00.57 & 10.85$\pm$00.90 \\
    REVE    & \textcolor{red}{\textbf{57.90$\pm$01.64}} & \textcolor{red}{\textbf{57.86$\pm$01.70}} & \textcolor{red}{\textbf{55.18$\pm$01.73}} & REVE & 65.79$\pm$02.09 & 65.21$\pm$03.10 & 64.92$\pm$03.42 \\
    NSP     & \textcolor{blue}{\textbf{54.40$\pm$00.95}} & \textcolor{blue}{\textbf{53.75$\pm$00.94}} & \textcolor{blue}{\textbf{50.72$\pm$00.89}} & NSP & \textcolor{red}{\textbf{96.83$\pm$00.91}} & \textcolor{red}{\textbf{96.72$\pm$00.91}} & \textcolor{red}{\textbf{96.61$\pm$00.90}} \\
\bottomrule
\end{tabular}
\rowcolors{2}{}{}
\end{table}

\subsubsection{Event-Related and Evoked Responses}

ERP CORE-ERN, ERP CORE-N170, and ERP CORE-N400 isolate error-related, face-perception, and semantic-violation responses in one-second windows, with 15,972, 12,800, and 4,800 examples, respectively~\citep{kappenman2021erpcore}. Schreuder2010 provides a larger auditory-P300 task with 253,800 windows from 21 participants~\citep{schreuder2010p300}. All four tasks are binary and report BAC, AUROC, and AUCPR. NSP leads all three metrics on ERP CORE-N400 and Schreuder2010, ranks second on ERP CORE-N170, and is below the strongest baselines on ERP CORE-ERN.

\begin{table}[H]
\centering
\caption{NeuralBench event-related and evoked-response results (\%, mean $\pm$ sample SD). Red and blue mark the best and second-best mean for each metric.}
\label{tab:neuralbench_erp_complete}
\vskip -0.04in
\fontsize{7.0}{7.6}\selectfont
\setlength{\tabcolsep}{3.0pt}
\renewcommand{\arraystretch}{1.04}
\rowcolors{3}{black!3}{white}
\begin{tabular}{@{}lccc@{\hspace{8pt}}lccc@{}}
\toprule
\multicolumn{4}{c}{\textbf{ERP CORE-ERN}} & \multicolumn{4}{c}{\textbf{ERP CORE-N170}} \\
\cmidrule(lr){1-4}\cmidrule(lr){5-8}
Model & BAC & AUROC & AUCPR & Model & BAC & AUROC & AUCPR \\
\midrule
    BENDR   & 78.06$\pm$04.59 & 86.11$\pm$04.33 & 96.37$\pm$01.35 & BENDR & 67.03$\pm$00.61 & 71.94$\pm$01.25 & 47.25$\pm$02.01 \\
BIOT    & 49.40$\pm$01.12 & 50.54$\pm$06.31 & 82.92$\pm$04.01 & BIOT & 49.84$\pm$00.36 & 50.20$\pm$00.46 & 25.55$\pm$00.46 \\
    LaBraM  & \textcolor{blue}{\textbf{81.96$\pm$01.52}} & \textcolor{blue}{\textbf{90.10$\pm$01.67}} & 97.18$\pm$00.75 & LaBraM & 72.14$\pm$00.55 & 78.75$\pm$01.01 & 57.33$\pm$02.71 \\
    CBraMod & 81.32$\pm$01.04 & 89.43$\pm$00.77 & \textcolor{blue}{\textbf{97.31$\pm$00.29}} & CBraMod & 71.14$\pm$00.55 & 78.72$\pm$01.37 & 57.09$\pm$03.52 \\
LUNA    & 81.57$\pm$03.26 & 89.74$\pm$02.40 & 97.09$\pm$01.05 & LUNA & 68.74$\pm$02.04 & 76.63$\pm$01.34 & 54.25$\pm$01.05 \\
    REVE    & \textcolor{red}{\textbf{84.61$\pm$01.92}} & \textcolor{red}{\textbf{91.73$\pm$01.41}} & \textcolor{red}{\textbf{97.76$\pm$00.59}} & REVE & \textcolor{red}{\textbf{74.03$\pm$05.33}} & \textcolor{red}{\textbf{80.88$\pm$03.75}} & \textcolor{red}{\textbf{59.25$\pm$05.32}} \\
    NSP     & 80.07$\pm$02.94 & 88.35$\pm$03.24 & 96.86$\pm$02.44 & NSP & \textcolor{blue}{\textbf{73.27$\pm$00.91}} & \textcolor{blue}{\textbf{79.88$\pm$00.99}} & \textcolor{blue}{\textbf{58.41$\pm$02.85}} \\
\midrule
\multicolumn{4}{c}{\textbf{ERP CORE-N400}} & \multicolumn{4}{c}{\textbf{Schreuder2010}} \\
\cmidrule(lr){1-4}\cmidrule(lr){5-8}
Model & BAC & AUROC & AUCPR & Model & BAC & AUROC & AUCPR \\
\midrule
    BENDR   & 60.45$\pm$01.15 & 65.56$\pm$01.11 & 68.95$\pm$01.46 & BENDR & \textcolor{blue}{\textbf{66.70$\pm$00.16}} & \textcolor{blue}{\textbf{72.33$\pm$00.12}} & \textcolor{blue}{\textbf{91.73$\pm$00.05}} \\
BIOT    & 50.07$\pm$00.64 & 49.29$\pm$00.43 & 49.44$\pm$01.14 & BIOT & 50.25$\pm$00.21 & 50.80$\pm$00.63 & 83.76$\pm$00.26 \\
LaBraM  & \textcolor{blue}{\textbf{64.62$\pm$01.09}} & 70.43$\pm$01.22 & 72.77$\pm$01.87 & LaBraM & 66.05$\pm$00.40 & 71.99$\pm$00.46 & 91.72$\pm$00.22 \\
CBraMod & 60.42$\pm$01.13 & 66.54$\pm$01.35 & 66.62$\pm$01.33 & CBraMod & 63.11$\pm$00.62 & 68.28$\pm$00.32 & 90.56$\pm$00.11 \\
LUNA    & 64.20$\pm$01.55 & 70.16$\pm$01.71 & 72.42$\pm$02.26 & LUNA & 62.97$\pm$00.39 & 68.12$\pm$00.32 & 90.38$\pm$00.13 \\
    REVE    & 64.51$\pm$01.97 & \textcolor{blue}{\textbf{71.31$\pm$01.06}} & \textcolor{blue}{\textbf{73.03$\pm$00.81}} & REVE & 64.70$\pm$01.00 & 70.81$\pm$00.49 & 91.15$\pm$00.26 \\
    NSP     & \textcolor{red}{\textbf{66.67$\pm$00.75}} & \textcolor{red}{\textbf{72.35$\pm$00.81}} & \textcolor{red}{\textbf{74.11$\pm$01.91}} & NSP & \textcolor{red}{\textbf{69.87$\pm$00.67}} & \textcolor{red}{\textbf{75.71$\pm$00.73}} & \textcolor{red}{\textbf{92.44$\pm$00.23}} \\
\bottomrule
\end{tabular}
\rowcolors{2}{}{}
\end{table}

\subsubsection{Cognitive State and Language Decoding}

Zyma2019 evaluates binary mental-arithmetic decoding using 1,659 windows from 35 participants~\citep{zyma2019}, while Hinss2023 evaluates three-way cognitive-state decoding using 15,399 windows from 29 participants~\citep{hinss2023cogbci}. ZuCo contains 25,819 sentence-reading examples from 12 participants~\citep{hollenstein2018zuco}, and Nieuwland2018 provides 529,730 word-level examples from 222 participants~\citep{nieuwland2018language}; these two tasks are evaluated as retrieval rather than classification. NSP ranks second on Zyma2019 BAC, leads all three Hinss2023 metrics, and obtains the best Top-1, Top-5, and median-rank results on both retrieval datasets. Retrieval values are interpreted within their own metric family and are not pooled into the 14-task BAC macro.

\begin{table}[H]
\centering
\caption{NeuralBench cognitive-state and language results (mean $\pm$ sample SD). All metrics except median rank (MedR) are percentages. Retrieval uses Top-1, Top-5, and MedR; MedR is an unscaled rank and lower is better. Red and blue mark the best and second-best mean for each metric.}
\label{tab:neuralbench_cognitive_complete}
\vskip -0.04in
\fontsize{7.0}{7.6}\selectfont
\setlength{\tabcolsep}{3.0pt}
\renewcommand{\arraystretch}{1.04}
\rowcolors{3}{black!3}{white}
\begin{tabular}{@{}lccc@{\hspace{8pt}}lccc@{}}
\toprule
\multicolumn{4}{c}{\textbf{Zyma2019}} & \multicolumn{4}{c}{\textbf{Hinss2023}} \\
\cmidrule(lr){1-4}\cmidrule(lr){5-8}
Model & BAC & AUROC & AUCPR & Model & BAC & F1 & $\kappa$ \\
\midrule
BENDR   & 65.28$\pm$02.34 & 74.69$\pm$05.12 & 89.33$\pm$03.31 & BENDR & 45.89$\pm$00.19 & 45.01$\pm$01.05 & 18.83$\pm$00.28 \\
BIOT    & 66.40$\pm$03.04 & 72.30$\pm$05.30 & 89.05$\pm$01.27 & BIOT & 43.18$\pm$04.73 & 42.35$\pm$05.03 & 14.77$\pm$07.09 \\
LaBraM  & 68.92$\pm$02.42 & 80.90$\pm$03.78 & 92.18$\pm$02.89 & LaBraM & 65.46$\pm$03.17 & 63.16$\pm$04.69 & 48.20$\pm$04.76 \\
CBraMod & 72.02$\pm$01.30 & 77.04$\pm$00.39 & 91.57$\pm$00.49 & CBraMod & 61.37$\pm$01.70 & 59.32$\pm$02.71 & 42.06$\pm$02.54 \\
    LUNA    & 72.35$\pm$00.41 & 76.85$\pm$00.69 & 90.16$\pm$02.05 & LUNA & \textcolor{blue}{\textbf{66.01$\pm$00.94}} & 64.59$\pm$01.16 & \textcolor{blue}{\textbf{49.01$\pm$01.40}} \\
    REVE    & \textcolor{red}{\textbf{72.88$\pm$02.81}} & \textcolor{red}{\textbf{85.54$\pm$01.30}} & \textcolor{red}{\textbf{94.12$\pm$00.86}} & REVE & 65.57$\pm$04.49 & \textcolor{blue}{\textbf{65.02$\pm$04.48}} & 48.35$\pm$06.73 \\
    NSP     & \textcolor{blue}{\textbf{72.50$\pm$05.04}} & \textcolor{blue}{\textbf{81.43$\pm$05.67}} & \textcolor{blue}{\textbf{92.27$\pm$03.42}} & NSP & \textcolor{red}{\textbf{72.93$\pm$00.42}} & \textcolor{red}{\textbf{71.26$\pm$00.41}} & \textcolor{red}{\textbf{55.14$\pm$00.47}} \\
\midrule
\multicolumn{4}{c}{\textbf{ZuCo}} & \multicolumn{4}{c}{\textbf{Nieuwland2018}} \\
\cmidrule(lr){1-4}\cmidrule(lr){5-8}
Model & Top-1 & Top-5 & MedR$\downarrow$ & Model & Top-1 & Top-5 & MedR$\downarrow$ \\
\midrule
BENDR   & 1.38$\pm$00.09 & 6.47$\pm$01.29 & 49.33$\pm$01.53 & BENDR & 7.48$\pm$00.85 & 20.73$\pm$01.69 & 26.00$\pm$02.65 \\
BIOT    & 1.91$\pm$00.28 & 7.67$\pm$00.39 & 49.67$\pm$02.31 & BIOT & 1.14$\pm$00.08 & 5.03$\pm$00.09 & 71.00$\pm$01.00 \\
LaBraM  & 2.24$\pm$00.24 & 8.95$\pm$00.26 & 45.00$\pm$01.00 & LaBraM & 11.26$\pm$00.86 & 28.34$\pm$01.27 & 17.33$\pm$01.53 \\
CBraMod & 0.54$\pm$00.17 & 2.25$\pm$00.30 & 118.33$\pm$03.79 & CBraMod & 11.61$\pm$00.22 & 29.73$\pm$00.66 & 15.33$\pm$00.58 \\
    LUNA    & \textcolor{blue}{\textbf{3.15$\pm$00.15}} & 11.32$\pm$00.56 & 40.67$\pm$01.53 & LUNA & 10.79$\pm$00.67 & 27.99$\pm$01.37 & 17.00$\pm$01.73 \\
    REVE    & 2.95$\pm$00.21 & \textcolor{blue}{\textbf{11.39$\pm$00.40}} & \textcolor{blue}{\textbf{38.67$\pm$01.53}} & REVE & \textcolor{blue}{\textbf{14.44$\pm$00.22}} & \textcolor{blue}{\textbf{33.73$\pm$00.68}} & \textcolor{blue}{\textbf{12.67$\pm$01.16}} \\
    NSP     & \textcolor{red}{\textbf{3.36$\pm$00.11}} & \textcolor{red}{\textbf{12.11$\pm$00.39}} & \textcolor{red}{\textbf{37.22$\pm$01.40}} & NSP & \textcolor{red}{\textbf{15.01$\pm$00.03}} & \textcolor{red}{\textbf{36.27$\pm$00.12}} & \textcolor{red}{\textbf{11.31$\pm$00.88}} \\
\bottomrule
\end{tabular}
\rowcolors{2}{}{}
\end{table}

%% file: tables/app_complete_eegfm_results.tex

\subsubsection{Emotion Recognition}
This domain evaluates affective decoding at progressively finer label granularity. SEED, SEED-V, and SEED-VII move from three to five and seven emotion classes while retaining dense multichannel recordings, allowing the comparison to expose how adaptation quality changes as inter-class boundaries become more subtle.

\paragraph{SEED.} SEED evaluates three-class emotion recognition from 60-channel, 10-s EEG windows, comprising 37,890 samples from 15 participants~\citep{zheng2015seed}. NSP obtains $73.40\pm00.26$ BAC with single-task full fine-tuning, $73.62\pm00.97$ with multi-task full fine-tuning, $68.93\pm00.60$ with frozen transfer, and $71.67\pm00.67$ with LoRA. Its best setting is multi-task full fine-tuning, which exceeds the strongest baseline under that setting by 1.00 point; the table reports the corresponding F1 and $\kappa$ values.

\begin{table}[H]
\centering
\caption{Complete SEED results across four adaptation settings (\%, mean $\pm$ sample SD). \textcolor{red}{\textbf{Bold red}} and \textcolor{blue}{\textbf{bold blue}} mark the best and second-best mean for each metric and protocol.}
\label{tab:eegfm_seed}
\vskip -0.04in
\fontsize{7.2}{7.8}\selectfont
\setlength{\tabcolsep}{3.2pt}
\renewcommand{\arraystretch}{1.05}
\rowcolors{3}{black!3}{white}
\begin{tabular}{@{}lccc@{\hspace{10pt}}lccc@{}}
\toprule
\multicolumn{4}{c}{\textbf{Full Fine-Tuning}} & \multicolumn{4}{c}{\textbf{Parameter-Efficient Multi-Task}} \\
\cmidrule(lr){1-4}\cmidrule(lr){5-8}
Model & \multicolumn{3}{c}{Single-task} & Model & \multicolumn{3}{c}{Frozen} \\
 & BAC & F1 & $\kappa$ & & BAC & F1 & $\kappa$ \\
\midrule
BENDR & 59.50$\pm$00.42 & 58.88$\pm$00.98 & 39.42$\pm$00.64 & BENDR & 34.12$\pm$00.24 & 24.48$\pm$02.82 & 01.10$\pm$00.32 \\
BIOT & 63.87$\pm$01.77 & 63.12$\pm$01.94 & 45.99$\pm$02.67 & BIOT & 57.85$\pm$00.19 & 58.17$\pm$00.21 & 37.17$\pm$00.29 \\
LaBraM & 61.59$\pm$01.71 & 60.30$\pm$01.60 & 43.28$\pm$02.48 & LaBraM & 52.13$\pm$00.05 & 51.23$\pm$00.34 & 28.33$\pm$00.12 \\
EEGPT & 70.83$\pm$00.28 & 70.48$\pm$00.16 & 56.41$\pm$00.42 & EEGPT & 61.80$\pm$00.65 & 60.41$\pm$01.07 & 42.89$\pm$00.97 \\
CBraMod & 62.59$\pm$02.40 & 61.67$\pm$02.66 & 44.36$\pm$03.58 & CBraMod & 44.32$\pm$00.28 & 44.50$\pm$00.30 & 16.76$\pm$00.45 \\
CSBrain & 70.23$\pm$00.25 & 70.70$\pm$00.22 & 55.70$\pm$00.28 & CSBrain & 58.57$\pm$00.05 & 58.53$\pm$00.05 & 38.00$\pm$00.14 \\
REVE & \textcolor{blue}{\textbf{73.37$\pm$00.30}} & \textcolor{blue}{\textbf{73.32$\pm$00.41}} & \textcolor{red}{\textbf{62.77$\pm$00.44}} & REVE & \textcolor{blue}{\textbf{63.22$\pm$00.13}} & \textcolor{blue}{\textbf{63.35$\pm$00.08}} & \textcolor{blue}{\textbf{45.20$\pm$00.32}} \\
NSP & \textcolor{red}{\textbf{73.40$\pm$00.26}} & \textcolor{red}{\textbf{73.77$\pm$00.31}} & \textcolor{blue}{\textbf{60.43$\pm$00.45}} & NSP & \textcolor{red}{\textbf{68.93$\pm$00.60}} & \textcolor{red}{\textbf{69.33$\pm$00.60}} & \textcolor{red}{\textbf{53.73$\pm$00.96}} \\
\midrule
Model & \multicolumn{3}{c}{Multi-task} & Model & \multicolumn{3}{c}{LoRA} \\
 & BAC & F1 & $\kappa$ & & BAC & F1 & $\kappa$ \\
\midrule
BENDR & 60.82$\pm$00.19 & 60.84$\pm$00.20 & 41.59$\pm$00.16 & BENDR & 33.30$\pm$00.00 & 17.00$\pm$00.00 & 00.00$\pm$00.00 \\
BIOT & 65.62$\pm$00.92 & 65.57$\pm$00.81 & 50.11$\pm$00.99 & BIOT & 62.70$\pm$00.14 & 62.60$\pm$00.08 & 44.20$\pm$00.22 \\
LaBraM & 67.71$\pm$00.55 & 68.11$\pm$00.66 & 52.22$\pm$00.83 & LaBraM & 45.90$\pm$00.22 & 43.07$\pm$02.04 & 18.90$\pm$00.29 \\
EEGPT & 68.75$\pm$00.40 & 67.42$\pm$00.78 & 52.37$\pm$00.28 & EEGPT & \textcolor{blue}{\textbf{69.97$\pm$00.05}} & \textcolor{blue}{\textbf{69.27$\pm$00.61}} & \textcolor{blue}{\textbf{55.10$\pm$00.08}} \\
CBraMod & 72.25$\pm$01.90 & \textcolor{blue}{\textbf{72.90$\pm$01.63}} & 58.68$\pm$02.91 & CBraMod & 66.07$\pm$00.21 & 65.57$\pm$00.38 & 49.27$\pm$00.29 \\
CSBrain & 71.92$\pm$00.39 & 71.84$\pm$00.20 & 58.02$\pm$00.59 & CSBrain & 64.67$\pm$00.05 & 64.07$\pm$00.17 & 47.20$\pm$00.08 \\
REVE & \textcolor{blue}{\textbf{72.62$\pm$00.92}} & 72.75$\pm$00.76 & \textcolor{red}{\textbf{62.57$\pm$00.94}} & REVE & 69.23$\pm$00.21 & 68.97$\pm$00.46 & 54.03$\pm$00.34 \\
NSP & \textcolor{red}{\textbf{73.62$\pm$00.97}} & \textcolor{red}{\textbf{74.02$\pm$00.97}} & \textcolor{blue}{\textbf{60.94$\pm$01.51}} & NSP & \textcolor{red}{\textbf{71.67$\pm$00.67}} & \textcolor{red}{\textbf{72.10$\pm$00.61}} & \textcolor{red}{\textbf{57.83$\pm$00.98}} \\
\bottomrule
\end{tabular}
\rowcolors{2}{}{}
\renewcommand{\arraystretch}{1}
\end{table}

\paragraph{SEED-V.} SEED-V increases emotion granularity to five classes and provides 60-channel, 10-s windows, with 12,318 samples from 16 participants~\citep{liu2022seedv}. NSP reaches $46.30\pm01.31$ BAC with single-task full fine-tuning, $43.42\pm00.77$ with multi-task full fine-tuning, $38.77\pm00.55$ with frozen transfer, and $41.77\pm00.21$ with LoRA. Single-task full fine-tuning is strongest and leads the best listed baseline by 5.65 points.

\begin{table}[H]
\centering
\caption{Complete SEED-V results across four adaptation settings (\%, mean $\pm$ sample SD). \textcolor{red}{\textbf{Bold red}} and \textcolor{blue}{\textbf{bold blue}} mark the best and second-best mean for each metric and protocol.}
\label{tab:eegfm_seed_v}
\vskip -0.04in
\fontsize{7.2}{7.8}\selectfont
\setlength{\tabcolsep}{3.2pt}
\renewcommand{\arraystretch}{1.05}
\rowcolors{3}{black!3}{white}
\begin{tabular}{@{}lccc@{\hspace{10pt}}lccc@{}}
\toprule
\multicolumn{4}{c}{\textbf{Full Fine-Tuning}} & \multicolumn{4}{c}{\textbf{Parameter-Efficient Multi-Task}} \\
\cmidrule(lr){1-4}\cmidrule(lr){5-8}
Model & \multicolumn{3}{c}{Single-task} & Model & \multicolumn{3}{c}{Frozen} \\
 & BAC & F1 & $\kappa$ & & BAC & F1 & $\kappa$ \\
\midrule
BENDR & 20.20$\pm$00.13 & 10.72$\pm$01.57 & 00.21$\pm$00.13 & BENDR & 20.00$\pm$00.00 & 06.00$\pm$01.98 & 00.00$\pm$00.00 \\
BIOT & 26.66$\pm$00.13 & 25.85$\pm$00.13 & 09.53$\pm$00.08 & BIOT & 26.27$\pm$00.47 & 24.77$\pm$00.90 & 09.17$\pm$00.54 \\
LaBraM & 24.35$\pm$00.55 & 23.64$\pm$01.21 & 05.61$\pm$01.13 & LaBraM & 23.23$\pm$00.48 & 19.63$\pm$00.91 & 04.43$\pm$00.68 \\
EEGPT & 36.43$\pm$01.33 & 33.00$\pm$01.94 & 18.52$\pm$00.67 & EEGPT & \textcolor{blue}{\textbf{30.48$\pm$00.73}} & \textcolor{blue}{\textbf{28.71$\pm$01.45}} & \textcolor{blue}{\textbf{12.90$\pm$00.57}} \\
CBraMod & 28.34$\pm$00.17 & 27.48$\pm$00.89 & 10.55$\pm$00.20 & CBraMod & 20.28$\pm$00.02 & 10.93$\pm$00.03 & 00.45$\pm$00.03 \\
CSBrain & 38.23$\pm$00.24 & 39.07$\pm$00.34 & 22.70$\pm$00.29 & CSBrain & 21.97$\pm$00.12 & 14.43$\pm$00.62 & 02.50$\pm$00.16 \\
REVE & \textcolor{blue}{\textbf{40.65$\pm$00.51}} & \textcolor{blue}{\textbf{40.35$\pm$01.63}} & \textcolor{blue}{\textbf{25.82$\pm$00.72}} & REVE & 25.43$\pm$00.33 & 24.07$\pm$00.41 & 07.07$\pm$00.37 \\
NSP & \textcolor{red}{\textbf{46.30$\pm$01.31}} & \textcolor{red}{\textbf{44.70$\pm$01.65}} & \textcolor{red}{\textbf{29.93$\pm$01.94}} & NSP & \textcolor{red}{\textbf{38.77$\pm$00.55}} & \textcolor{red}{\textbf{39.50$\pm$00.44}} & \textcolor{red}{\textbf{23.50$\pm$00.70}} \\
\midrule
Model & \multicolumn{3}{c}{Multi-task} & Model & \multicolumn{3}{c}{LoRA} \\
 & BAC & F1 & $\kappa$ & & BAC & F1 & $\kappa$ \\
\midrule
BENDR & 21.75$\pm$00.66 & 19.94$\pm$02.58 & 02.90$\pm$00.68 & BENDR & 20.00$\pm$00.00 & 08.80$\pm$00.00 & 00.00$\pm$00.00 \\
BIOT & 30.54$\pm$00.18 & 30.97$\pm$00.26 & 14.14$\pm$00.21 & BIOT & 28.37$\pm$00.12 & 27.87$\pm$00.33 & 11.33$\pm$00.19 \\
LaBraM & \textcolor{blue}{\textbf{43.26$\pm$00.29}} & \textcolor{blue}{\textbf{43.20$\pm$00.81}} & \textcolor{blue}{\textbf{28.93$\pm$00.50}} & LaBraM & 23.93$\pm$00.53 & 22.10$\pm$00.80 & 05.17$\pm$00.69 \\
EEGPT & 43.07$\pm$00.60 & 42.99$\pm$00.54 & 28.62$\pm$00.86 & EEGPT & 36.07$\pm$00.85 & 35.77$\pm$01.77 & 20.10$\pm$00.75 \\
CBraMod & 40.56$\pm$00.77 & 41.72$\pm$00.72 & 26.22$\pm$00.97 & CBraMod & 35.03$\pm$00.27 & 33.22$\pm$00.54 & 18.88$\pm$00.33 \\
CSBrain & 38.02$\pm$00.23 & 38.07$\pm$01.03 & 22.48$\pm$00.42 & CSBrain & 30.43$\pm$00.17 & 28.20$\pm$00.65 & 13.17$\pm$00.29 \\
REVE & 40.84$\pm$00.98 & 40.58$\pm$02.00 & 25.59$\pm$01.48 & REVE & \textcolor{blue}{\textbf{38.37$\pm$00.21}} & \textcolor{blue}{\textbf{38.87$\pm$00.42}} & \textcolor{blue}{\textbf{22.70$\pm$00.49}} \\
NSP & \textcolor{red}{\textbf{43.42$\pm$00.77}} & \textcolor{red}{\textbf{43.44$\pm$00.85}} & \textcolor{red}{\textbf{28.96$\pm$01.05}} & NSP & \textcolor{red}{\textbf{41.77$\pm$00.21}} & \textcolor{red}{\textbf{41.47$\pm$00.31}} & \textcolor{red}{\textbf{27.03$\pm$00.35}} \\
\bottomrule
\end{tabular}
\rowcolors{2}{}{}
\renewcommand{\arraystretch}{1}
\end{table}

\paragraph{SEED-VII.} SEED-VII is a seven-class emotion-recognition task with 60-channel, 15-s windows, totaling 19,420 samples from 20 participants~\citep{jiang2024seedvii}. NSP records $24.50\pm00.98$ BAC with single-task full fine-tuning, $26.50\pm01.49$ with multi-task full fine-tuning, $24.23\pm00.80$ with frozen transfer, and $23.40\pm01.22$ with LoRA. Multi-task full fine-tuning is the strongest NSP setting but remains 1.28 points below the best baseline under the same protocol.

\begin{table}[H]
\centering
\caption{Complete SEED-VII results across four adaptation settings (\%, mean $\pm$ sample SD). \textcolor{red}{\textbf{Bold red}} and \textcolor{blue}{\textbf{bold blue}} mark the best and second-best mean for each metric and protocol.}
\label{tab:eegfm_seed_vii}
\vskip -0.04in
\fontsize{7.2}{7.8}\selectfont
\setlength{\tabcolsep}{3.2pt}
\renewcommand{\arraystretch}{1.05}
\rowcolors{3}{black!3}{white}
\begin{tabular}{@{}lccc@{\hspace{10pt}}lccc@{}}
\toprule
\multicolumn{4}{c}{\textbf{Full Fine-Tuning}} & \multicolumn{4}{c}{\textbf{Parameter-Efficient Multi-Task}} \\
\cmidrule(lr){1-4}\cmidrule(lr){5-8}
Model & \multicolumn{3}{c}{Single-task} & Model & \multicolumn{3}{c}{Frozen} \\
 & BAC & F1 & $\kappa$ & & BAC & F1 & $\kappa$ \\
\midrule
BENDR & 17.43$\pm$00.60 & 14.20$\pm$00.79 & 03.98$\pm$00.69 & BENDR & 14.30$\pm$00.00 & 03.63$\pm$00.61 & 00.00$\pm$00.00 \\
BIOT & 18.89$\pm$00.43 & 16.84$\pm$01.12 & 05.80$\pm$00.48 & BIOT & 20.60$\pm$00.45 & 16.60$\pm$00.43 & 07.50$\pm$00.42 \\
LaBraM & 20.39$\pm$00.41 & 19.55$\pm$03.82 & 09.42$\pm$02.64 & LaBraM & 23.23$\pm$00.48 & 19.63$\pm$00.91 & 04.43$\pm$00.68 \\
EEGPT & \textcolor{red}{\textbf{26.42$\pm$00.55}} & \textcolor{red}{\textbf{23.89$\pm$01.29}} & \textcolor{red}{\textbf{14.05$\pm$00.99}} & EEGPT & \textcolor{red}{\textbf{25.20$\pm$00.08}} & \textcolor{red}{\textbf{24.17$\pm$00.12}} & \textcolor{red}{\textbf{12.70$\pm$00.08}} \\
CBraMod & 23.10$\pm$00.93 & 21.82$\pm$02.00 & 09.70$\pm$02.62 & CBraMod & 19.43$\pm$00.72 & 16.76$\pm$00.29 & 06.97$\pm$00.95 \\
CSBrain & 23.73$\pm$00.17 & 23.43$\pm$00.12 & 10.40$\pm$00.22 & CSBrain & 18.90$\pm$00.29 & 16.60$\pm$00.50 & 06.27$\pm$00.41 \\
REVE & 22.77$\pm$00.17 & \textcolor{blue}{\textbf{23.67$\pm$00.09}} & \textcolor{blue}{\textbf{10.60$\pm$00.08}} & REVE & 20.57$\pm$00.17 & \textcolor{blue}{\textbf{20.50$\pm$00.33}} & 08.50$\pm$00.29 \\
NSP & \textcolor{blue}{\textbf{24.50$\pm$00.98}} & 22.27$\pm$01.27 & 10.30$\pm$01.35 & NSP & \textcolor{blue}{\textbf{24.23$\pm$00.80}} & 19.13$\pm$01.14 & \textcolor{blue}{\textbf{10.07$\pm$01.10}} \\
\midrule
Model & \multicolumn{3}{c}{Multi-task} & Model & \multicolumn{3}{c}{LoRA} \\
 & BAC & F1 & $\kappa$ & & BAC & F1 & $\kappa$ \\
\midrule
BENDR & 21.98$\pm$00.66 & 19.31$\pm$00.87 & 09.49$\pm$01.03 & BENDR & 14.30$\pm$00.00 & 04.77$\pm$00.33 & 00.00$\pm$00.00 \\
BIOT & 23.93$\pm$00.39 & 22.13$\pm$00.52 & 11.38$\pm$00.38 & BIOT & 23.85$\pm$00.08 & 21.95$\pm$00.12 & 11.37$\pm$00.11 \\
LaBraM & 26.13$\pm$00.92 & 24.74$\pm$00.57 & 13.07$\pm$00.93 & LaBraM & 18.13$\pm$00.71 & 13.30$\pm$01.36 & 04.37$\pm$01.23 \\
EEGPT & \textcolor{red}{\textbf{27.78$\pm$00.47}} & 25.40$\pm$00.12 & \textcolor{red}{\textbf{15.80$\pm$01.12}} & EEGPT & \textcolor{red}{\textbf{29.83$\pm$00.53}} & \textcolor{red}{\textbf{27.30$\pm$01.55}} & \textcolor{red}{\textbf{17.73$\pm$01.03}} \\
CBraMod & 26.05$\pm$00.72 & \textcolor{blue}{\textbf{25.74$\pm$01.22}} & 13.09$\pm$00.86 & CBraMod & 18.73$\pm$00.19 & 14.87$\pm$01.00 & 05.00$\pm$00.36 \\
CSBrain & 26.67$\pm$01.37 & 22.68$\pm$02.69 & 11.40$\pm$02.04 & CSBrain & 22.73$\pm$00.09 & 22.27$\pm$00.24 & 10.80$\pm$00.08 \\
REVE & \textcolor{blue}{\textbf{27.62$\pm$00.61}} & 23.93$\pm$01.08 & 14.78$\pm$00.44 & REVE & \textcolor{blue}{\textbf{23.93$\pm$00.58}} & 21.67$\pm$01.76 & \textcolor{blue}{\textbf{11.48$\pm$00.88}} \\
NSP & 26.50$\pm$01.49 & \textcolor{red}{\textbf{26.30$\pm$01.82}} & \textcolor{blue}{\textbf{14.96$\pm$01.77}} & NSP & 23.40$\pm$01.22 & \textcolor{blue}{\textbf{23.30$\pm$01.42}} & 10.10$\pm$01.65 \\
\bottomrule
\end{tabular}
\rowcolors{2}{}{}
\renewcommand{\arraystretch}{1}
\end{table}

\subsubsection{Motor Imagery}
This domain covers imagined-movement decoding across heterogeneous cohorts, montages, and action taxonomies. PhysioMI provides a large-participant four-class setting, Mimul-11 targets three upper-limb actions, and BCIC-2a supplies the standard four-class BCI protocol.

\paragraph{PhysioMI.} PhysioMI evaluates four-class motor imagery from 64-channel, 4-s trials, comprising 9,747 samples from 109 participants~\citep{schalk2004bci2000}. NSP achieves $59.23\pm00.67$ BAC with single-task full fine-tuning, $58.68\pm00.21$ with multi-task full fine-tuning, $36.90\pm00.25$ with frozen transfer, and $45.13\pm00.95$ with LoRA. Single-task full fine-tuning performs best for NSP and is 0.80 points below the strongest baseline in that setting.

\begin{table}[H]
\centering
\caption{Complete PhysioMI results across four adaptation settings (\%, mean $\pm$ sample SD). \textcolor{red}{\textbf{Bold red}} and \textcolor{blue}{\textbf{bold blue}} mark the best and second-best mean for each metric and protocol.}
\label{tab:eegfm_physiomi}
\vskip -0.04in
\fontsize{7.2}{7.8}\selectfont
\setlength{\tabcolsep}{3.2pt}
\renewcommand{\arraystretch}{1.05}
\rowcolors{3}{black!3}{white}
\begin{tabular}{@{}lccc@{\hspace{10pt}}lccc@{}}
\toprule
\multicolumn{4}{c}{\textbf{Full Fine-Tuning}} & \multicolumn{4}{c}{\textbf{Parameter-Efficient Multi-Task}} \\
\cmidrule(lr){1-4}\cmidrule(lr){5-8}
Model & \multicolumn{3}{c}{Single-task} & Model & \multicolumn{3}{c}{Frozen} \\
 & BAC & F1 & $\kappa$ & & BAC & F1 & $\kappa$ \\
\midrule
BENDR & 47.78$\pm$00.28 & 42.72$\pm$08.93 & 36.48$\pm$08.47 & BENDR & 25.22$\pm$00.24 & 12.55$\pm$02.53 & 00.27$\pm$00.31 \\
BIOT & 36.22$\pm$00.10 & 35.65$\pm$00.18 & 14.95$\pm$00.15 & BIOT & 29.80$\pm$00.33 & 29.90$\pm$00.43 & 06.40$\pm$00.41 \\
LaBraM & 57.27$\pm$00.26 & 57.29$\pm$00.25 & 43.00$\pm$00.35 & LaBraM & 29.63$\pm$00.34 & 28.97$\pm$00.77 & 06.23$\pm$00.42 \\
EEGPT & 54.16$\pm$00.18 & 53.27$\pm$00.92 & 38.04$\pm$01.08 & EEGPT & \textcolor{red}{\textbf{39.90$\pm$01.39}} & \textcolor{red}{\textbf{39.51$\pm$01.92}} & \textcolor{red}{\textbf{19.87$\pm$01.84}} \\
CBraMod & 56.74$\pm$00.36 & 56.74$\pm$00.39 & 42.31$\pm$00.48 & CBraMod & 26.90$\pm$00.24 & 22.76$\pm$00.17 & 02.54$\pm$00.32 \\
CSBrain & 56.57$\pm$00.45 & 56.57$\pm$00.53 & 42.07$\pm$00.59 & CSBrain & 26.80$\pm$00.29 & 24.77$\pm$01.25 & 02.40$\pm$00.37 \\
REVE & \textcolor{red}{\textbf{60.03$\pm$00.58}} & \textcolor{red}{\textbf{60.08$\pm$00.62}} & \textcolor{red}{\textbf{46.65$\pm$00.80}} & REVE & 32.55$\pm$00.18 & 32.12$\pm$00.40 & 10.08$\pm$00.23 \\
NSP & \textcolor{blue}{\textbf{59.23$\pm$00.67}} & \textcolor{blue}{\textbf{59.00$\pm$00.78}} & \textcolor{blue}{\textbf{45.63$\pm$00.93}} & NSP & \textcolor{blue}{\textbf{36.90$\pm$00.25}} & \textcolor{blue}{\textbf{36.47$\pm$00.65}} & \textcolor{blue}{\textbf{15.83$\pm$00.67}} \\
\midrule
Model & \multicolumn{3}{c}{Multi-task} & Model & \multicolumn{3}{c}{LoRA} \\
 & BAC & F1 & $\kappa$ & & BAC & F1 & $\kappa$ \\
\midrule
BENDR & 44.77$\pm$00.31 & 44.63$\pm$00.03 & 26.33$\pm$00.40 & BENDR & 25.10$\pm$00.08 & 11.90$\pm$02.55 & 00.13$\pm$00.12 \\
BIOT & 33.02$\pm$00.29 & 32.75$\pm$00.36 & 10.67$\pm$00.40 & BIOT & 26.93$\pm$00.05 & 26.17$\pm$00.48 & 02.57$\pm$00.09 \\
LaBraM & 43.19$\pm$01.03 & 42.38$\pm$01.15 & 24.25$\pm$01.37 & LaBraM & 27.40$\pm$00.16 & 20.47$\pm$01.13 & 03.20$\pm$00.24 \\
EEGPT & 50.52$\pm$00.80 & 50.26$\pm$00.76 & 34.00$\pm$01.06 & EEGPT & \textcolor{blue}{\textbf{45.83$\pm$00.85}} & \textcolor{blue}{\textbf{45.33$\pm$00.74}} & \textcolor{blue}{\textbf{27.80$\pm$01.10}} \\
CBraMod & 41.80$\pm$00.69 & 40.35$\pm$00.87 & 22.42$\pm$00.92 & CBraMod & 27.60$\pm$00.22 & 24.50$\pm$02.03 & 03.50$\pm$00.29 \\
CSBrain & 55.43$\pm$00.37 & 55.10$\pm$00.63 & 40.48$\pm$00.48 & CSBrain & 30.68$\pm$00.28 & 29.92$\pm$00.61 & 07.57$\pm$00.38 \\
REVE & \textcolor{red}{\textbf{58.82$\pm$00.45}} & \textcolor{red}{\textbf{58.77$\pm$00.56}} & \textcolor{red}{\textbf{45.10$\pm$00.60}} & REVE & \textcolor{red}{\textbf{53.95$\pm$00.33}} & \textcolor{red}{\textbf{54.18$\pm$00.31}} & \textcolor{red}{\textbf{38.62$\pm$00.46}} \\
NSP & \textcolor{blue}{\textbf{58.68$\pm$00.21}} & \textcolor{blue}{\textbf{57.84$\pm$00.89}} & \textcolor{blue}{\textbf{41.54$\pm$00.81}} & NSP & 45.13$\pm$00.95 & 44.97$\pm$01.98 & 26.87$\pm$00.60 \\
\bottomrule
\end{tabular}
\rowcolors{2}{}{}
\renewcommand{\arraystretch}{1}
\end{table}

\paragraph{Mimul-11.} Mimul-11 is a three-class upper-limb motor-imagery task with 60-channel, 5-s trials, containing 41,347 samples from 11 participants~\citep{jeong2020mimul}. NSP obtains $49.70\pm00.91$ BAC with single-task full fine-tuning, $51.66\pm00.60$ with multi-task full fine-tuning, $40.47\pm00.26$ with frozen transfer, and $44.83\pm00.80$ with LoRA. Multi-task full fine-tuning is its strongest setting and trails the leading baseline by 0.89 points.

\begin{table}[H]
\centering
\caption{Complete Mimul-11 results across four adaptation settings (\%, mean $\pm$ sample SD). \textcolor{red}{\textbf{Bold red}} and \textcolor{blue}{\textbf{bold blue}} mark the best and second-best mean for each metric and protocol.}
\label{tab:eegfm_mimul_11}
\vskip -0.04in
\fontsize{7.2}{7.8}\selectfont
\setlength{\tabcolsep}{3.2pt}
\renewcommand{\arraystretch}{1.05}
\rowcolors{3}{black!3}{white}
\begin{tabular}{@{}lccc@{\hspace{10pt}}lccc@{}}
\toprule
\multicolumn{4}{c}{\textbf{Full Fine-Tuning}} & \multicolumn{4}{c}{\textbf{Parameter-Efficient Multi-Task}} \\
\cmidrule(lr){1-4}\cmidrule(lr){5-8}
Model & \multicolumn{3}{c}{Single-task} & Model & \multicolumn{3}{c}{Frozen} \\
 & BAC & F1 & $\kappa$ & & BAC & F1 & $\kappa$ \\
\midrule
BENDR & \textcolor{red}{\textbf{50.44$\pm$02.30}} & \textcolor{red}{\textbf{59.09$\pm$01.23}} & \textcolor{red}{\textbf{36.10$\pm$03.49}} & BENDR & 33.30$\pm$00.00 & 37.90$\pm$00.00 & 00.00$\pm$00.00 \\
BIOT & 40.55$\pm$01.14 & 49.68$\pm$01.29 & 14.45$\pm$02.10 & BIOT & 36.65$\pm$00.13 & 43.92$\pm$00.38 & 06.90$\pm$00.27 \\
LaBraM & 47.80$\pm$01.68 & \textcolor{blue}{\textbf{57.43$\pm$01.27}} & \textcolor{blue}{\textbf{30.71$\pm$03.14}} & LaBraM & 35.50$\pm$00.43 & 42.97$\pm$00.88 & 04.60$\pm$00.93 \\
EEGPT & 45.19$\pm$01.69 & 54.66$\pm$01.38 & 23.68$\pm$02.30 & EEGPT & 37.11$\pm$01.97 & \textcolor{blue}{\textbf{44.95$\pm$03.07}} & 10.06$\pm$03.66 \\
CBraMod & 47.07$\pm$00.09 & 55.10$\pm$00.36 & 25.13$\pm$00.69 & CBraMod & 33.95$\pm$00.11 & 40.09$\pm$00.18 & 01.51$\pm$00.25 \\
CSBrain & 42.67$\pm$00.77 & 51.93$\pm$00.87 & 18.13$\pm$02.05 & CSBrain & 34.80$\pm$00.59 & 42.73$\pm$01.05 & 03.10$\pm$01.39 \\
REVE & 45.41$\pm$01.15 & 50.63$\pm$00.71 & 18.93$\pm$01.86 & REVE & \textcolor{blue}{\textbf{40.20$\pm$00.32}} & \textcolor{red}{\textbf{48.02$\pm$00.50}} & \textcolor{blue}{\textbf{12.38$\pm$00.64}} \\
NSP & \textcolor{blue}{\textbf{49.70$\pm$00.91}} & 51.33$\pm$00.78 & 30.03$\pm$00.89 & NSP & \textcolor{red}{\textbf{40.47$\pm$00.26}} & 44.67$\pm$00.93 & \textcolor{red}{\textbf{13.80$\pm$00.42}} \\
\midrule
Model & \multicolumn{3}{c}{Multi-task} & Model & \multicolumn{3}{c}{LoRA} \\
 & BAC & F1 & $\kappa$ & & BAC & F1 & $\kappa$ \\
\midrule
BENDR & \textcolor{red}{\textbf{52.55$\pm$01.52}} & \textcolor{red}{\textbf{59.61$\pm$02.80}} & \textcolor{red}{\textbf{38.35$\pm$04.93}} & BENDR & 33.30$\pm$00.00 & 37.90$\pm$00.00 & 00.00$\pm$00.00 \\
BIOT & 40.43$\pm$00.63 & 49.02$\pm$00.73 & 14.63$\pm$01.88 & BIOT & 40.57$\pm$00.37 & 49.53$\pm$00.31 & 12.40$\pm$00.57 \\
LaBraM & 45.05$\pm$00.66 & 53.42$\pm$01.98 & 22.60$\pm$02.20 & LaBraM & 34.87$\pm$00.24 & 41.67$\pm$00.45 & 03.63$\pm$00.40 \\
EEGPT & 50.77$\pm$00.61 & \textcolor{blue}{\textbf{57.44$\pm$01.11}} & \textcolor{blue}{\textbf{31.16$\pm$02.43}} & EEGPT & 43.00$\pm$00.36 & \textcolor{red}{\textbf{52.00$\pm$00.50}} & \textcolor{red}{\textbf{18.90$\pm$00.29}} \\
CBraMod & 44.53$\pm$00.19 & 52.80$\pm$00.57 & 19.23$\pm$00.77 & CBraMod & 42.90$\pm$00.92 & 50.00$\pm$01.22 & 15.07$\pm$01.62 \\
CSBrain & 41.64$\pm$00.54 & 50.70$\pm$00.61 & 15.94$\pm$01.33 & CSBrain & 39.10$\pm$00.24 & 48.23$\pm$00.26 & 12.13$\pm$00.37 \\
REVE & 45.22$\pm$00.13 & 53.73$\pm$00.65 & 21.53$\pm$01.16 & REVE & \textcolor{blue}{\textbf{43.77$\pm$00.82}} & \textcolor{blue}{\textbf{50.33$\pm$01.98}} & 17.90$\pm$02.01 \\
NSP & \textcolor{blue}{\textbf{51.66$\pm$00.60}} & 55.26$\pm$00.81 & 23.44$\pm$00.88 & NSP & \textcolor{red}{\textbf{44.83$\pm$00.80}} & 50.20$\pm$00.96 & \textcolor{blue}{\textbf{18.13$\pm$00.59}} \\
\bottomrule
\end{tabular}
\rowcolors{2}{}{}
\renewcommand{\arraystretch}{1}
\end{table}

\paragraph{BCIC-2a.} BCIC-2a provides a standard four-class motor-imagery task with 22-channel, 4-s trials, totaling 5,088 samples from nine participants~\citep{tangermann2012bcicomp}. NSP reaches $46.50\pm00.52$ BAC with single-task full fine-tuning, $46.58\pm01.48$ with multi-task full fine-tuning, $36.50\pm01.21$ with frozen transfer, and $37.03\pm00.83$ with LoRA. Its best result comes from multi-task full fine-tuning and exceeds the strongest baseline in that setting by 4.08 points.

\begin{table}[H]
\centering
\caption{Complete BCIC-2a results across four adaptation settings (\%, mean $\pm$ sample SD). \textcolor{red}{\textbf{Bold red}} and \textcolor{blue}{\textbf{bold blue}} mark the best and second-best mean for each metric and protocol.}
\label{tab:eegfm_bcic_2a}
\vskip -0.04in
\fontsize{7.2}{7.8}\selectfont
\setlength{\tabcolsep}{3.2pt}
\renewcommand{\arraystretch}{1.05}
\rowcolors{3}{black!3}{white}
\begin{tabular}{@{}lccc@{\hspace{10pt}}lccc@{}}
\toprule
\multicolumn{4}{c}{\textbf{Full Fine-Tuning}} & \multicolumn{4}{c}{\textbf{Parameter-Efficient Multi-Task}} \\
\cmidrule(lr){1-4}\cmidrule(lr){5-8}
Model & \multicolumn{3}{c}{Single-task} & Model & \multicolumn{3}{c}{Frozen} \\
 & BAC & F1 & $\kappa$ & & BAC & F1 & $\kappa$ \\
\midrule
BENDR & 35.21$\pm$00.54 & 33.25$\pm$00.57 & 13.62$\pm$00.72 & BENDR & 25.17$\pm$00.24 & 10.65$\pm$00.92 & 00.22$\pm$00.31 \\
BIOT & 33.33$\pm$00.79 & 29.85$\pm$00.27 & 11.10$\pm$01.04 & BIOT & 29.73$\pm$00.54 & 25.10$\pm$00.43 & 06.97$\pm$00.79 \\
LaBraM & 32.37$\pm$00.21 & 30.73$\pm$00.39 & 09.83$\pm$00.30 & LaBraM & 28.40$\pm$00.24 & 22.83$\pm$00.45 & 04.57$\pm$00.33 \\
EEGPT & \textcolor{blue}{\textbf{44.07$\pm$03.27}} & 38.61$\pm$04.23 & \textcolor{blue}{\textbf{25.42$\pm$04.36}} & EEGPT & \textcolor{blue}{\textbf{32.02$\pm$00.06}} & \textcolor{blue}{\textbf{25.95$\pm$01.13}} & \textcolor{blue}{\textbf{09.40$\pm$00.08}} \\
CBraMod & 33.71$\pm$00.86 & 28.99$\pm$01.35 & 11.61$\pm$01.14 & CBraMod & 29.17$\pm$01.03 & 24.94$\pm$01.72 & 05.56$\pm$01.37 \\
CSBrain & 42.10$\pm$00.87 & \textcolor{blue}{\textbf{40.64$\pm$00.62}} & 22.83$\pm$01.13 & CSBrain & 28.13$\pm$00.17 & 18.57$\pm$00.12 & 04.17$\pm$00.26 \\
REVE & 42.73$\pm$01.10 & \textcolor{red}{\textbf{42.60$\pm$01.77}} & \textcolor{red}{\textbf{27.08$\pm$01.46}} & REVE & 29.18$\pm$00.83 & 23.37$\pm$01.36 & 05.57$\pm$01.12 \\
NSP & \textcolor{red}{\textbf{46.50$\pm$00.52}} & 37.13$\pm$00.97 & 23.63$\pm$00.95 & NSP & \textcolor{red}{\textbf{36.50$\pm$01.21}} & \textcolor{red}{\textbf{32.10$\pm$02.98}} & \textcolor{red}{\textbf{15.33$\pm$01.95}} \\
\midrule
Model & \multicolumn{3}{c}{Multi-task} & Model & \multicolumn{3}{c}{LoRA} \\
 & BAC & F1 & $\kappa$ & & BAC & F1 & $\kappa$ \\
\midrule
BENDR & 34.98$\pm$00.25 & 30.01$\pm$02.27 & 13.31$\pm$00.33 & BENDR & 25.00$\pm$00.00 & 10.00$\pm$00.00 & 00.00$\pm$00.00 \\
BIOT & 29.66$\pm$00.39 & 23.61$\pm$04.64 & 05.99$\pm$00.45 & BIOT & 29.00$\pm$00.14 & 19.40$\pm$00.14 & 05.33$\pm$00.19 \\
LaBraM & 34.58$\pm$01.64 & 31.24$\pm$02.71 & 12.60$\pm$02.00 & LaBraM & 28.30$\pm$00.73 & 17.87$\pm$01.01 & 04.43$\pm$00.93 \\
EEGPT & 39.23$\pm$01.48 & 33.08$\pm$01.86 & 16.95$\pm$01.97 & EEGPT & 36.70$\pm$01.15 & 32.67$\pm$03.16 & 15.60$\pm$01.53 \\
CBraMod & 35.50$\pm$00.58 & 23.97$\pm$00.25 & 14.00$\pm$00.77 & CBraMod & 31.00$\pm$00.70 & 22.20$\pm$02.19 & 08.03$\pm$00.98 \\
CSBrain & \textcolor{blue}{\textbf{42.50$\pm$01.08}} & \textcolor{blue}{\textbf{39.97$\pm$02.34}} & \textcolor{red}{\textbf{23.32$\pm$01.48}} & CSBrain & 32.00$\pm$00.99 & 20.77$\pm$01.11 & 09.30$\pm$01.35 \\
REVE & 41.87$\pm$03.34 & 38.68$\pm$02.25 & 22.48$\pm$03.47 & REVE & \textcolor{red}{\textbf{38.63$\pm$00.19}} & \textcolor{red}{\textbf{38.43$\pm$00.17}} & \textcolor{red}{\textbf{18.17$\pm$00.25}} \\
NSP & \textcolor{red}{\textbf{46.58$\pm$01.48}} & \textcolor{red}{\textbf{40.64$\pm$02.07}} & \textcolor{blue}{\textbf{22.76$\pm$02.28}} & NSP & \textcolor{blue}{\textbf{37.03$\pm$00.83}} & \textcolor{blue}{\textbf{34.50$\pm$01.80}} & \textcolor{blue}{\textbf{16.03$\pm$01.10}} \\
\bottomrule
\end{tabular}
\rowcolors{2}{}{}
\renewcommand{\arraystretch}{1}
\end{table}

\subsubsection{Clinical EEG Analysis}
This domain spans event-level and recording-level clinical interpretation. TUEV and TUSL require localization of transient abnormal events, whereas TUAB evaluates whole-record abnormality, jointly testing robustness to class imbalance and substantial recording heterogeneity.

\paragraph{TUEV.} TUEV evaluates six-class abnormal-event classification from 21-channel, 5-s windows, comprising 113,353 samples from 370 participants~\citep{harati2015tuev}. NSP obtains $73.17\pm01.70$ BAC with single-task full fine-tuning, $74.34\pm01.03$ with multi-task full fine-tuning, $52.90\pm02.16$ with frozen transfer, and $69.60\pm01.25$ with LoRA. Multi-task full fine-tuning is strongest and narrowly exceeds the best baseline in that setting by 0.07 points.

\begin{table}[H]
\centering
\caption{Complete TUEV results across four adaptation settings (\%, mean $\pm$ sample SD). \textcolor{red}{\textbf{Bold red}} and \textcolor{blue}{\textbf{bold blue}} mark the best and second-best mean for each metric and protocol.}
\label{tab:eegfm_tuev}
\vskip -0.04in
\fontsize{7.2}{7.8}\selectfont
\setlength{\tabcolsep}{3.2pt}
\renewcommand{\arraystretch}{1.05}
\rowcolors{3}{black!3}{white}
\begin{tabular}{@{}lccc@{\hspace{10pt}}lccc@{}}
\toprule
\multicolumn{4}{c}{\textbf{Full Fine-Tuning}} & \multicolumn{4}{c}{\textbf{Parameter-Efficient Multi-Task}} \\
\cmidrule(lr){1-4}\cmidrule(lr){5-8}
Model & \multicolumn{3}{c}{Single-task} & Model & \multicolumn{3}{c}{Frozen} \\
 & BAC & F1 & $\kappa$ & & BAC & F1 & $\kappa$ \\
\midrule
BENDR & 65.53$\pm$02.40 & 80.95$\pm$02.18 & 66.62$\pm$04.07 & BENDR & 16.70$\pm$00.00 & 44.10$\pm$00.00 & 00.00$\pm$00.00 \\
BIOT & 57.07$\pm$04.29 & 74.34$\pm$04.00 & 39.55$\pm$04.15 & BIOT & 52.12$\pm$00.55 & 80.93$\pm$00.20 & 68.85$\pm$00.28 \\
LaBraM & 59.58$\pm$04.31 & 77.72$\pm$03.25 & 64.06$\pm$05.03 & LaBraM & 41.27$\pm$00.12 & 73.37$\pm$00.39 & 54.97$\pm$00.53 \\
EEGPT & 66.86$\pm$02.54 & 83.95$\pm$00.14 & 74.45$\pm$00.09 & EEGPT & \textcolor{blue}{\textbf{63.08$\pm$00.66}} & \textcolor{blue}{\textbf{86.22$\pm$00.31}} & \textcolor{blue}{\textbf{76.83$\pm$00.71}} \\
CBraMod & 65.42$\pm$00.90 & 79.03$\pm$01.36 & 64.13$\pm$02.48 & CBraMod & 32.50$\pm$00.04 & 65.97$\pm$00.20 & 41.58$\pm$00.49 \\
CSBrain & 71.20$\pm$01.34 & 85.07$\pm$00.85 & 74.67$\pm$01.44 & CSBrain & 38.63$\pm$00.17 & 72.87$\pm$00.19 & 53.40$\pm$00.28 \\
REVE & \textcolor{red}{\textbf{73.98$\pm$01.21}} & \textcolor{red}{\textbf{90.25$\pm$01.42}} & \textcolor{red}{\textbf{85.07$\pm$01.80}} & REVE & \textcolor{red}{\textbf{69.03$\pm$00.17}} & \textcolor{red}{\textbf{89.53$\pm$00.30}} & \textcolor{red}{\textbf{83.22$\pm$00.52}} \\
NSP & \textcolor{blue}{\textbf{73.17$\pm$01.70}} & \textcolor{blue}{\textbf{88.90$\pm$01.25}} & \textcolor{blue}{\textbf{81.03$\pm$01.00}} & NSP & 52.90$\pm$02.16 & 81.87$\pm$01.40 & 70.33$\pm$02.27 \\
\midrule
Model & \multicolumn{3}{c}{Multi-task} & Model & \multicolumn{3}{c}{LoRA} \\
 & BAC & F1 & $\kappa$ & & BAC & F1 & $\kappa$ \\
\midrule
BENDR & 67.46$\pm$02.59 & 84.56$\pm$01.82 & 72.85$\pm$02.87 & BENDR & 16.70$\pm$00.00 & 44.10$\pm$00.00 & 00.00$\pm$00.00 \\
BIOT & 61.35$\pm$01.35 & 80.55$\pm$00.46 & 68.37$\pm$00.61 & BIOT & 59.68$\pm$00.17 & 81.28$\pm$00.31 & 68.33$\pm$00.59 \\
LaBraM & 71.53$\pm$00.19 & 86.22$\pm$00.73 & 77.42$\pm$01.20 & LaBraM & 47.13$\pm$00.59 & 72.33$\pm$01.32 & 54.33$\pm$02.28 \\
EEGPT & 70.85$\pm$01.16 & 86.35$\pm$00.47 & 77.69$\pm$01.06 & EEGPT & 61.07$\pm$00.98 & \textcolor{blue}{\textbf{86.83$\pm$00.24}} & \textcolor{blue}{\textbf{79.60$\pm$01.07}} \\
CBraMod & 69.41$\pm$01.86 & 83.44$\pm$00.31 & 72.02$\pm$00.65 & CBraMod & 57.80$\pm$02.06 & 77.60$\pm$03.19 & 63.07$\pm$04.76 \\
CSBrain & 71.82$\pm$00.69 & 87.00$\pm$00.64 & 77.84$\pm$01.55 & CSBrain & 64.02$\pm$00.61 & 79.05$\pm$01.32 & 65.80$\pm$02.08 \\
REVE & \textcolor{blue}{\textbf{74.27$\pm$00.87}} & \textcolor{blue}{\textbf{89.37$\pm$00.33}} & \textcolor{red}{\textbf{82.12$\pm$00.76}} & REVE & \textcolor{blue}{\textbf{68.53$\pm$01.18}} & 77.83$\pm$00.86 & 58.30$\pm$00.99 \\
NSP & \textcolor{red}{\textbf{74.34$\pm$01.03}} & \textcolor{red}{\textbf{90.48$\pm$00.94}} & \textcolor{blue}{\textbf{81.92$\pm$01.80}} & NSP & \textcolor{red}{\textbf{69.60$\pm$01.25}} & \textcolor{red}{\textbf{90.57$\pm$00.22}} & \textcolor{red}{\textbf{84.83$\pm$01.37}} \\
\bottomrule
\end{tabular}
\rowcolors{2}{}{}
\renewcommand{\arraystretch}{1}
\end{table}

\paragraph{TUSL.} TUSL classifies three types of EEG slowing events from 21- or 22-channel, 10-s windows and contains 290 annotated samples from 28 participants~\citep{vonweltin2017tusl}. NSP records $81.17\pm01.12$ BAC with single-task full fine-tuning, $81.83\pm01.25$ with multi-task full fine-tuning, $67.07\pm01.25$ with frozen transfer, and $80.33\pm00.70$ with LoRA. Multi-task full fine-tuning is the best NSP configuration and leads the strongest baseline under that protocol by 1.27 points.

\begin{table}[H]
\centering
\caption{Complete TUSL results across four adaptation settings (\%, mean $\pm$ sample SD). \textcolor{red}{\textbf{Bold red}} and \textcolor{blue}{\textbf{bold blue}} mark the best and second-best mean for each metric and protocol.}
\label{tab:eegfm_tusl}
\vskip -0.04in
\fontsize{7.2}{7.8}\selectfont
\setlength{\tabcolsep}{3.2pt}
\renewcommand{\arraystretch}{1.05}
\rowcolors{3}{black!3}{white}
\begin{tabular}{@{}lccc@{\hspace{10pt}}lccc@{}}
\toprule
\multicolumn{4}{c}{\textbf{Full Fine-Tuning}} & \multicolumn{4}{c}{\textbf{Parameter-Efficient Multi-Task}} \\
\cmidrule(lr){1-4}\cmidrule(lr){5-8}
Model & \multicolumn{3}{c}{Single-task} & Model & \multicolumn{3}{c}{Frozen} \\
 & BAC & F1 & $\kappa$ & & BAC & F1 & $\kappa$ \\
\midrule
BENDR & 47.22$\pm$01.48 & 34.17$\pm$02.80 & 17.88$\pm$01.95 & BENDR & 33.30$\pm$00.00 & 17.50$\pm$00.00 & 00.00$\pm$00.00 \\
BIOT & 60.05$\pm$04.47 & 56.83$\pm$04.49 & 37.52$\pm$06.77 & BIOT & 47.62$\pm$00.05 & 44.17$\pm$00.56 & 18.75$\pm$00.19 \\
LaBraM & 66.98$\pm$00.44 & 62.43$\pm$01.73 & 45.08$\pm$01.57 & LaBraM & 63.90$\pm$02.52 & 54.57$\pm$02.18 & 37.63$\pm$04.29 \\
EEGPT & 62.41$\pm$03.10 & 70.63$\pm$02.34 & 58.92$\pm$02.63 & EEGPT & \textcolor{red}{\textbf{69.13$\pm$01.65}} & \textcolor{red}{\textbf{69.27$\pm$02.15}} & \textcolor{red}{\textbf{59.70$\pm$03.25}} \\
CBraMod & 74.18$\pm$02.69 & 73.15$\pm$02.74 & 63.67$\pm$03.23 & CBraMod & 33.00$\pm$00.00 & 09.35$\pm$01.20 & 00.00$\pm$00.00 \\
CSBrain & 71.60$\pm$02.63 & 72.60$\pm$02.69 & 61.91$\pm$03.85 & CSBrain & \textcolor{blue}{\textbf{68.17$\pm$01.04}} & \textcolor{blue}{\textbf{62.93$\pm$01.79}} & \textcolor{blue}{\textbf{47.63$\pm$01.93}} \\
REVE & \textcolor{blue}{\textbf{79.58$\pm$01.32}} & \textcolor{blue}{\textbf{74.58$\pm$01.25}} & \textcolor{blue}{\textbf{65.07$\pm$01.86}} & REVE & 63.88$\pm$00.54 & 55.37$\pm$00.87 & 38.93$\pm$00.97 \\
NSP & \textcolor{red}{\textbf{81.17$\pm$01.12}} & \textcolor{red}{\textbf{76.37$\pm$01.32}} & \textcolor{red}{\textbf{67.50$\pm$01.10}} & NSP & 67.07$\pm$01.25 & 59.27$\pm$01.15 & 42.70$\pm$00.26 \\
\midrule
Model & \multicolumn{3}{c}{Multi-task} & Model & \multicolumn{3}{c}{LoRA} \\
 & BAC & F1 & $\kappa$ & & BAC & F1 & $\kappa$ \\
\midrule
BENDR & 73.35$\pm$00.98 & 70.90$\pm$01.73 & 56.65$\pm$02.55 & BENDR & 33.30$\pm$00.00 & 25.70$\pm$00.00 & 00.00$\pm$00.00 \\
BIOT & 69.90$\pm$04.10 & 71.03$\pm$05.19 & 56.00$\pm$07.50 & BIOT & 62.90$\pm$00.28 & 54.53$\pm$00.47 & 36.93$\pm$00.33 \\
LaBraM & 75.86$\pm$01.15 & 72.74$\pm$01.33 & 59.60$\pm$01.81 & LaBraM & 54.68$\pm$00.47 & 37.08$\pm$01.33 & 25.47$\pm$00.82 \\
EEGPT & 76.12$\pm$02.05 & \textcolor{red}{\textbf{84.19$\pm$01.04}} & \textcolor{red}{\textbf{84.30$\pm$01.27}} & EEGPT & 72.59$\pm$00.81 & 71.10$\pm$00.03 & 56.74$\pm$00.23 \\
CBraMod & 79.56$\pm$03.24 & 76.24$\pm$04.45 & 64.72$\pm$05.96 & CBraMod & 73.90$\pm$00.74 & 67.57$\pm$01.41 & 53.93$\pm$01.47 \\
CSBrain & \textcolor{blue}{\textbf{80.56$\pm$00.91}} & 76.72$\pm$00.38 & 65.57$\pm$00.41 & CSBrain & 75.03$\pm$02.53 & 71.53$\pm$03.20 & 58.12$\pm$04.65 \\
REVE & 79.85$\pm$01.27 & 77.81$\pm$01.89 & 66.62$\pm$02.35 & REVE & \textcolor{red}{\textbf{80.68$\pm$02.81}} & \textcolor{red}{\textbf{80.78$\pm$02.41}} & \textcolor{red}{\textbf{71.43$\pm$03.54}} \\
NSP & \textcolor{red}{\textbf{81.83$\pm$01.25}} & \textcolor{blue}{\textbf{79.07$\pm$01.06}} & \textcolor{blue}{\textbf{68.57$\pm$02.06}} & NSP & \textcolor{blue}{\textbf{80.33$\pm$00.70}} & \textcolor{blue}{\textbf{80.47$\pm$00.48}} & \textcolor{blue}{\textbf{71.27$\pm$00.75}} \\
\bottomrule
\end{tabular}
\rowcolors{2}{}{}
\renewcommand{\arraystretch}{1}
\end{table}

\paragraph{TUAB.} TUAB is a binary normal-versus-abnormal recording task represented by 23-channel, 30-s windows, totaling 272,320 samples from 2,383 participants~\citep{lopez2015tuab}. NSP reaches $83.83\pm00.50$ BAC with single-task full fine-tuning, $82.58\pm00.48$ with multi-task full fine-tuning, $81.80\pm00.26$ with frozen transfer, and $80.10\pm00.29$ with LoRA. Single-task full fine-tuning is strongest and improves on the leading baseline in that setting by 1.11 points.

\begin{table}[H]
\centering
\caption{Complete TUAB results across four adaptation settings (\%, mean $\pm$ sample SD). \textcolor{red}{\textbf{Bold red}} and \textcolor{blue}{\textbf{bold blue}} mark the best and second-best mean for each metric and protocol.}
\label{tab:eegfm_tuab}
\vskip -0.04in
\fontsize{7.2}{7.8}\selectfont
\setlength{\tabcolsep}{3.2pt}
\renewcommand{\arraystretch}{1.05}
\rowcolors{3}{black!3}{white}
\begin{tabular}{@{}lccc@{\hspace{10pt}}lccc@{}}
\toprule
\multicolumn{4}{c}{\textbf{Full Fine-Tuning}} & \multicolumn{4}{c}{\textbf{Parameter-Efficient Multi-Task}} \\
\cmidrule(lr){1-4}\cmidrule(lr){5-8}
Model & \multicolumn{3}{c}{Single-task} & Model & \multicolumn{3}{c}{Frozen} \\
 & BAC & AUROC & AUCPR & & BAC & AUROC & AUCPR \\
\midrule
BENDR & \textcolor{blue}{\textbf{82.72$\pm$00.12}} & \textcolor{red}{\textbf{89.52$\pm$00.14}} & \textcolor{red}{\textbf{90.68$\pm$00.13}} & BENDR & 68.30$\pm$05.85 & 73.17$\pm$00.82 & 65.93$\pm$00.45 \\
BIOT & 79.52$\pm$00.59 & 88.27$\pm$00.16 & 88.45$\pm$00.07 & BIOT & 80.30$\pm$00.22 & 87.50$\pm$00.14 & 87.27$\pm$00.57 \\
LaBraM & 79.50$\pm$00.81 & 86.91$\pm$00.51 & 83.86$\pm$01.96 & LaBraM & 75.87$\pm$00.05 & 84.07$\pm$00.05 & 85.11$\pm$00.08 \\
EEGPT & 78.66$\pm$00.24 & 87.52$\pm$01.01 & 86.94$\pm$02.15 & EEGPT & 79.53$\pm$00.05 & 88.63$\pm$00.12 & \textcolor{blue}{\textbf{88.47$\pm$00.05}} \\
CBraMod & 81.40$\pm$00.33 & 89.02$\pm$00.16 & \textcolor{blue}{\textbf{89.93$\pm$00.31}} & CBraMod & 73.15$\pm$00.19 & 80.41$\pm$00.02 & 79.79$\pm$00.15 \\
CSBrain & 78.97$\pm$00.17 & 85.33$\pm$01.51 & 82.73$\pm$04.69 & CSBrain & 78.20$\pm$00.00 & 87.00$\pm$00.08 & 87.23$\pm$00.12 \\
REVE & 81.32$\pm$00.35 & 87.93$\pm$01.76 & 87.38$\pm$01.56 & REVE & \textcolor{blue}{\textbf{80.80$\pm$00.32}} & \textcolor{blue}{\textbf{88.77$\pm$00.41}} & 88.33$\pm$00.55 \\
NSP & \textcolor{red}{\textbf{83.83$\pm$00.50}} & \textcolor{blue}{\textbf{89.40$\pm$01.00}} & 87.97$\pm$01.01 & NSP & \textcolor{red}{\textbf{81.80$\pm$00.26}} & \textcolor{red}{\textbf{90.80$\pm$00.20}} & \textcolor{red}{\textbf{91.17$\pm$00.23}} \\
\midrule
Model & \multicolumn{3}{c}{Multi-task} & Model & \multicolumn{3}{c}{LoRA} \\
 & BAC & AUROC & AUCPR & & BAC & AUROC & AUCPR \\
\midrule
BENDR & \textcolor{blue}{\textbf{81.72$\pm$00.26}} & \textcolor{red}{\textbf{90.38$\pm$00.27}} & \textcolor{blue}{\textbf{90.22$\pm$00.15}} & BENDR & 68.37$\pm$00.12 & 76.77$\pm$00.21 & 71.93$\pm$00.41 \\
BIOT & 80.85$\pm$00.36 & 88.67$\pm$00.62 & 89.01$\pm$00.47 & BIOT & \textcolor{blue}{\textbf{80.67$\pm$00.28}} & \textcolor{blue}{\textbf{88.10$\pm$00.29}} & 88.78$\pm$00.32 \\
LaBraM & 80.47$\pm$00.12 & 88.33$\pm$00.59 & 88.65$\pm$00.28 & LaBraM & 73.93$\pm$00.65 & 82.57$\pm$00.48 & 81.67$\pm$00.74 \\
EEGPT & 80.68$\pm$00.37 & \textcolor{blue}{\textbf{90.10$\pm$00.52}} & \textcolor{red}{\textbf{90.35$\pm$00.54}} & EEGPT & \textcolor{red}{\textbf{83.93$\pm$00.50}} & \textcolor{red}{\textbf{91.93$\pm$00.39}} & \textcolor{red}{\textbf{91.93$\pm$00.21}} \\
CBraMod & 81.55$\pm$00.14 & 89.57$\pm$00.26 & 89.67$\pm$00.02 & CBraMod & 78.40$\pm$00.93 & 86.07$\pm$00.97 & 87.07$\pm$00.95 \\
CSBrain & 79.18$\pm$00.48 & 86.32$\pm$00.28 & 85.78$\pm$00.25 & CSBrain & 78.97$\pm$00.71 & 87.07$\pm$01.01 & 87.23$\pm$01.09 \\
REVE & 81.18$\pm$00.46 & 88.32$\pm$00.37 & 89.33$\pm$00.61 & REVE & 80.43$\pm$01.04 & 88.03$\pm$00.54 & 88.07$\pm$01.16 \\
NSP & \textcolor{red}{\textbf{82.58$\pm$00.48}} & 88.18$\pm$00.40 & 86.06$\pm$01.09 & NSP & 80.10$\pm$00.29 & 88.00$\pm$00.06 & \textcolor{blue}{\textbf{88.93$\pm$01.29}} \\
\bottomrule
\end{tabular}
\rowcolors{2}{}{}
\renewcommand{\arraystretch}{1}
\end{table}

\subsubsection{Sleep Staging}
Sleep staging evaluates whether the learned representation preserves long-duration physiological structure from a sparse montage. HMC uses standard 30-s epochs and five sleep stages, complementing the shorter event-centered tasks elsewhere in the benchmark.

\paragraph{HMC.} HMC evaluates five-stage sleep classification from a sparse four-channel montage using 30-s epochs, comprising 136,925 samples from 151 participants~\citep{alvarezestevez2021hmc}. NSP reaches $75.23\pm00.60$ BAC with single-task full fine-tuning, $75.66\pm00.60$ with multi-task full fine-tuning, $72.97\pm00.97$ with frozen transfer, and $75.53\pm00.25$ with LoRA. Multi-task full fine-tuning performs best and exceeds the strongest baseline under that protocol by 3.48 points.

\begin{table}[H]
\centering
\caption{Complete HMC results across four adaptation settings (\%, mean $\pm$ sample SD). \textcolor{red}{\textbf{Bold red}} and \textcolor{blue}{\textbf{bold blue}} mark the best and second-best mean for each metric and protocol.}
\label{tab:eegfm_hmc}
\vskip -0.04in
\fontsize{7.2}{7.8}\selectfont
\setlength{\tabcolsep}{3.2pt}
\renewcommand{\arraystretch}{1.05}
\rowcolors{3}{black!3}{white}
\begin{tabular}{@{}lccc@{\hspace{10pt}}lccc@{}}
\toprule
\multicolumn{4}{c}{\textbf{Full Fine-Tuning}} & \multicolumn{4}{c}{\textbf{Parameter-Efficient Multi-Task}} \\
\cmidrule(lr){1-4}\cmidrule(lr){5-8}
Model & \multicolumn{3}{c}{Single-task} & Model & \multicolumn{3}{c}{Frozen} \\
 & BAC & F1 & $\kappa$ & & BAC & F1 & $\kappa$ \\
\midrule
BENDR & \textcolor{blue}{\textbf{72.63$\pm$00.13}} & 73.67$\pm$00.53 & 65.86$\pm$00.70 & BENDR & 24.18$\pm$00.02 & 27.02$\pm$00.02 & 07.50$\pm$00.05 \\
BIOT & 71.01$\pm$00.07 & 72.27$\pm$00.11 & 64.40$\pm$00.35 & BIOT & 66.13$\pm$00.21 & 70.27$\pm$00.29 & 61.03$\pm$00.17 \\
LaBraM & 70.85$\pm$00.44 & 71.52$\pm$00.55 & 64.47$\pm$00.70 & LaBraM & 59.80$\pm$00.00 & 64.10$\pm$00.37 & 53.40$\pm$00.16 \\
EEGPT & 69.67$\pm$01.24 & 73.03$\pm$00.32 & 64.64$\pm$00.92 & EEGPT & \textcolor{blue}{\textbf{67.83$\pm$00.12}} & \textcolor{blue}{\textbf{71.63$\pm$00.05}} & \textcolor{blue}{\textbf{62.90$\pm$00.08}} \\
CBraMod & 71.14$\pm$00.14 & 72.76$\pm$00.55 & 64.86$\pm$00.41 & CBraMod & 51.81$\pm$00.55 & 58.11$\pm$00.82 & 44.95$\pm$00.68 \\
CSBrain & 71.13$\pm$00.26 & 73.10$\pm$00.22 & 65.20$\pm$00.29 & CSBrain & 65.60$\pm$00.00 & 69.73$\pm$00.21 & 62.00$\pm$00.28 \\
REVE & 71.82$\pm$00.82 & \textcolor{blue}{\textbf{75.08$\pm$00.74}} & \textcolor{blue}{\textbf{66.49$\pm$00.83}} & REVE & 63.80$\pm$00.08 & 65.33$\pm$00.09 & 56.27$\pm$00.05 \\
NSP & \textcolor{red}{\textbf{75.23$\pm$00.60}} & \textcolor{red}{\textbf{78.03$\pm$00.67}} & \textcolor{red}{\textbf{71.57$\pm$00.60}} & NSP & \textcolor{red}{\textbf{72.97$\pm$00.97}} & \textcolor{red}{\textbf{74.93$\pm$00.67}} & \textcolor{red}{\textbf{67.87$\pm$00.91}} \\
\midrule
Model & \multicolumn{3}{c}{Multi-task} & Model & \multicolumn{3}{c}{LoRA} \\
 & BAC & F1 & $\kappa$ & & BAC & F1 & $\kappa$ \\
\midrule
BENDR & 70.21$\pm$00.72 & 72.44$\pm$00.65 & 64.57$\pm$00.90 & BENDR & 20.00$\pm$00.00 & 20.30$\pm$00.00 & 00.00$\pm$00.00 \\
BIOT & 71.45$\pm$01.05 & 72.89$\pm$00.34 & 64.89$\pm$00.71 & BIOT & 70.43$\pm$00.19 & \textcolor{blue}{\textbf{74.03$\pm$00.19}} & \textcolor{blue}{\textbf{65.90$\pm$00.24}} \\
LaBraM & 71.63$\pm$01.04 & 72.28$\pm$01.08 & 65.22$\pm$01.23 & LaBraM & 44.10$\pm$01.31 & 47.10$\pm$01.56 & 32.67$\pm$01.39 \\
EEGPT & 71.30$\pm$00.91 & 72.46$\pm$00.96 & 64.86$\pm$00.77 & EEGPT & \textcolor{blue}{\textbf{71.97$\pm$00.12}} & 72.27$\pm$00.39 & 64.63$\pm$00.21 \\
CBraMod & 72.03$\pm$00.14 & \textcolor{blue}{\textbf{75.20$\pm$00.29}} & \textcolor{blue}{\textbf{67.43$\pm$00.12}} & CBraMod & 69.13$\pm$00.17 & 71.60$\pm$00.73 & 63.00$\pm$00.28 \\
CSBrain & \textcolor{blue}{\textbf{72.18$\pm$00.24}} & 74.16$\pm$00.27 & 66.43$\pm$00.16 & CSBrain & 69.93$\pm$00.09 & 73.13$\pm$00.17 & 64.70$\pm$00.14 \\
REVE & 71.93$\pm$00.45 & 72.07$\pm$00.73 & 64.40$\pm$00.54 & REVE & 71.03$\pm$00.25 & 71.77$\pm$00.25 & 63.83$\pm$00.26 \\
NSP & \textcolor{red}{\textbf{75.66$\pm$00.60}} & \textcolor{red}{\textbf{77.02$\pm$00.76}} & \textcolor{red}{\textbf{70.24$\pm$00.87}} & NSP & \textcolor{red}{\textbf{75.53$\pm$00.25}} & \textcolor{red}{\textbf{76.77$\pm$00.67}} & \textcolor{red}{\textbf{70.10$\pm$00.70}} \\
\bottomrule
\end{tabular}
\rowcolors{2}{}{}
\renewcommand{\arraystretch}{1}
\end{table}

\subsubsection{Seizure Detection}
Seizure detection tests rare-event discrimination under marked class imbalance. Siena is therefore reported with AUROC and AUCPR in addition to balanced accuracy so that both ranking quality and positive-class retrieval remain visible.

\paragraph{Siena.} Siena is a binary seizure-detection dataset with 29-channel, 10-s windows, totaling 50,830 samples from 14 participants~\citep{detti2020siena}. NSP obtains $82.67\pm01.27$ BAC with single-task full fine-tuning, $86.90\pm00.78$ with multi-task full fine-tuning, $69.93\pm00.06$ with frozen transfer, and $77.37\pm02.11$ with LoRA. Multi-task full fine-tuning is strongest and leads the best baseline in that setting by 3.62 points; AUROC and AUCPR are included because seizure events are rare.

\begin{table}[H]
\centering
\caption{Complete Siena results across four adaptation settings (\%, mean $\pm$ sample SD). \textcolor{red}{\textbf{Bold red}} and \textcolor{blue}{\textbf{bold blue}} mark the best and second-best mean for each metric and protocol.}
\label{tab:eegfm_siena}
\vskip -0.04in
\fontsize{7.2}{7.8}\selectfont
\setlength{\tabcolsep}{3.2pt}
\renewcommand{\arraystretch}{1.05}
\rowcolors{3}{black!3}{white}
\begin{tabular}{@{}lccc@{\hspace{10pt}}lccc@{}}
\toprule
\multicolumn{4}{c}{\textbf{Full Fine-Tuning}} & \multicolumn{4}{c}{\textbf{Parameter-Efficient Multi-Task}} \\
\cmidrule(lr){1-4}\cmidrule(lr){5-8}
Model & \multicolumn{3}{c}{Single-task} & Model & \multicolumn{3}{c}{Frozen} \\
 & BAC & AUROC & AUCPR & & BAC & AUROC & AUCPR \\
\midrule
BENDR & 74.93$\pm$01.76 & 90.90$\pm$03.60 & 99.81$\pm$00.08 & BENDR & 50.00$\pm$00.00 & 61.20$\pm$04.16 & 99.27$\pm$00.12 \\
BIOT & 72.05$\pm$01.15 & 84.23$\pm$03.42 & 99.72$\pm$00.09 & BIOT & 56.42$\pm$00.02 & 81.20$\pm$00.13 & \textcolor{blue}{\textbf{99.70$\pm$00.00}} \\
LaBraM & 70.97$\pm$02.68 & 91.33$\pm$02.00 & \textcolor{red}{\textbf{99.87$\pm$00.05}} & LaBraM & 50.17$\pm$00.24 & 84.48$\pm$01.45 & \textcolor{blue}{\textbf{99.70$\pm$00.05}} \\
EEGPT & 76.25$\pm$01.02 & \textcolor{blue}{\textbf{92.08$\pm$01.15}} & 99.81$\pm$00.02 & EEGPT & \textcolor{red}{\textbf{73.90$\pm$01.28}} & \textcolor{red}{\textbf{97.28$\pm$00.43}} & \textcolor{red}{\textbf{99.93$\pm$00.02}} \\
CBraMod & 80.64$\pm$01.40 & \textcolor{red}{\textbf{93.86$\pm$00.25}} & \textcolor{red}{\textbf{99.87$\pm$00.01}} & CBraMod & 64.17$\pm$01.18 & \textcolor{blue}{\textbf{90.16$\pm$00.41}} & 99.68$\pm$00.02 \\
CSBrain & 79.25$\pm$01.68 & 91.52$\pm$02.96 & 99.83$\pm$00.13 & CSBrain & 50.00$\pm$00.00 & 56.53$\pm$12.17 & 98.10$\pm$00.62 \\
REVE & \textcolor{blue}{\textbf{80.95$\pm$00.32}} & 91.77$\pm$03.26 & \textcolor{blue}{\textbf{99.85$\pm$00.07}} & REVE & 68.60$\pm$00.08 & 77.87$\pm$00.37 & 99.27$\pm$00.05 \\
NSP & \textcolor{red}{\textbf{82.67$\pm$01.27}} & 91.77$\pm$00.76 & 99.57$\pm$00.06 & NSP & \textcolor{blue}{\textbf{69.93$\pm$00.06}} & 75.70$\pm$00.44 & 99.40$\pm$00.00 \\
\midrule
Model & \multicolumn{3}{c}{Multi-task} & Model & \multicolumn{3}{c}{LoRA} \\
 & BAC & AUROC & AUCPR & & BAC & AUROC & AUCPR \\
\midrule
BENDR & 78.96$\pm$03.08 & 91.31$\pm$04.74 & 99.81$\pm$00.11 & BENDR & 50.00$\pm$00.00 & 80.70$\pm$05.21 & 99.67$\pm$00.19 \\
BIOT & 74.35$\pm$00.71 & 81.20$\pm$00.92 & 99.26$\pm$00.05 & BIOT & 68.83$\pm$00.47 & 88.20$\pm$01.00 & 99.80$\pm$00.00 \\
LaBraM & 73.62$\pm$00.62 & 89.43$\pm$03.59 & 99.82$\pm$00.07 & LaBraM & 50.80$\pm$00.57 & 68.57$\pm$17.66 & 99.20$\pm$00.65 \\
EEGPT & \textcolor{blue}{\textbf{83.28$\pm$01.56}} & \textcolor{red}{\textbf{93.44$\pm$00.12}} & 99.86$\pm$00.00 & EEGPT & \textcolor{red}{\textbf{84.57$\pm$00.61}} & \textcolor{red}{\textbf{94.50$\pm$00.45}} & \textcolor{red}{\textbf{99.90$\pm$00.00}} \\
CBraMod & 82.75$\pm$02.10 & 92.03$\pm$01.90 & 99.83$\pm$00.04 & CBraMod & \textcolor{blue}{\textbf{82.13$\pm$00.78}} & \textcolor{blue}{\textbf{93.20$\pm$01.35}} & \textcolor{blue}{\textbf{99.83$\pm$00.05}} \\
CSBrain & 76.42$\pm$01.43 & 92.70$\pm$00.67 & \textcolor{blue}{\textbf{99.87$\pm$00.05}} & CSBrain & 72.87$\pm$01.11 & 87.80$\pm$03.18 & 99.77$\pm$00.11 \\
REVE & 81.87$\pm$02.03 & 90.70$\pm$01.25 & \textcolor{red}{\textbf{99.88$\pm$00.03}} & REVE & 75.33$\pm$01.23 & 88.73$\pm$01.05 & \textcolor{blue}{\textbf{99.83$\pm$00.05}} \\
NSP & \textcolor{red}{\textbf{86.90$\pm$00.78}} & \textcolor{blue}{\textbf{93.10$\pm$01.55}} & 99.54$\pm$00.09 & NSP & 77.37$\pm$02.11 & 81.97$\pm$03.36 & 99.53$\pm$00.06 \\
\bottomrule
\end{tabular}
\rowcolors{2}{}{}
\renewcommand{\arraystretch}{1}
\end{table}

\subsubsection{Mental Workload}
Mental workload recognition tests state-level cognitive decoding rather than stimulus-locked or motor responses. The Workload dataset provides a binary stress/workload setting with a conventional 19-channel montage.

\paragraph{Workload.} Workload is a binary cognitive-load task based on 19-channel, 10-s windows, comprising 2,134 samples from 36 participants~\citep{lim2018stew}. NSP achieves $82.53\pm00.30$ BAC with single-task full fine-tuning, $74.28\pm01.80$ with multi-task full fine-tuning, $75.77\pm00.99$ with frozen transfer, and $76.13\pm01.70$ with LoRA. Single-task full fine-tuning is strongest and improves over the best listed baseline under that protocol by 8.41 points.

\begin{table}[H]
\centering
\caption{Complete Workload results across four adaptation settings (\%, mean $\pm$ sample SD). \textcolor{red}{\textbf{Bold red}} and \textcolor{blue}{\textbf{bold blue}} mark the best and second-best mean for each metric and protocol.}
\label{tab:eegfm_workload}
\vskip -0.04in
\fontsize{7.2}{7.8}\selectfont
\setlength{\tabcolsep}{3.2pt}
\renewcommand{\arraystretch}{1.05}
\rowcolors{3}{black!3}{white}
\begin{tabular}{@{}lccc@{\hspace{10pt}}lccc@{}}
\toprule
\multicolumn{4}{c}{\textbf{Full Fine-Tuning}} & \multicolumn{4}{c}{\textbf{Parameter-Efficient Multi-Task}} \\
\cmidrule(lr){1-4}\cmidrule(lr){5-8}
Model & \multicolumn{3}{c}{Single-task} & Model & \multicolumn{3}{c}{Frozen} \\
 & BAC & AUROC & AUCPR & & BAC & AUROC & AUCPR \\
\midrule
BENDR & 50.00$\pm$00.00 & 52.36$\pm$01.25 & 30.70$\pm$00.55 & BENDR & 50.00$\pm$00.00 & 48.83$\pm$01.68 & 24.83$\pm$00.87 \\
BIOT & 63.98$\pm$00.66 & 77.68$\pm$01.62 & \textcolor{blue}{\textbf{69.30$\pm$02.27}} & BIOT & 58.25$\pm$00.50 & 76.48$\pm$00.48 & 54.17$\pm$00.30 \\
LaBraM & 61.17$\pm$01.01 & 67.78$\pm$01.48 & 45.83$\pm$02.21 & LaBraM & 57.45$\pm$02.83 & 73.43$\pm$00.56 & 46.72$\pm$00.74 \\
EEGPT & 62.31$\pm$03.98 & 70.46$\pm$04.00 & 48.13$\pm$06.90 & EEGPT & 61.05$\pm$03.33 & 75.68$\pm$00.59 & 58.23$\pm$01.57 \\
CBraMod & 71.94$\pm$00.60 & 81.63$\pm$01.63 & 59.73$\pm$01.79 & CBraMod & 50.00$\pm$00.00 & 66.87$\pm$00.17 & 42.70$\pm$00.07 \\
CSBrain & 69.53$\pm$02.65 & 77.07$\pm$04.26 & 64.67$\pm$03.20 & CSBrain & 55.07$\pm$00.94 & 76.83$\pm$00.12 & 54.73$\pm$00.12 \\
REVE & \textcolor{blue}{\textbf{74.12$\pm$01.01}} & \textcolor{blue}{\textbf{83.47$\pm$01.74}} & 64.43$\pm$03.71 & REVE & \textcolor{blue}{\textbf{67.10$\pm$00.00}} & \textcolor{red}{\textbf{79.83$\pm$00.17}} & \textcolor{blue}{\textbf{61.20$\pm$00.85}} \\
NSP & \textcolor{red}{\textbf{82.53$\pm$00.30}} & \textcolor{red}{\textbf{87.80$\pm$01.68}} & \textcolor{red}{\textbf{81.73$\pm$03.71}} & NSP & \textcolor{red}{\textbf{75.77$\pm$00.99}} & \textcolor{blue}{\textbf{79.73$\pm$00.12}} & \textcolor{red}{\textbf{68.50$\pm$01.78}} \\
\midrule
Model & \multicolumn{3}{c}{Multi-task} & Model & \multicolumn{3}{c}{LoRA} \\
 & BAC & AUROC & AUCPR & & BAC & AUROC & AUCPR \\
\midrule
BENDR & 63.32$\pm$02.29 & 74.09$\pm$01.04 & 54.78$\pm$01.84 & BENDR & 50.00$\pm$00.00 & 49.20$\pm$00.64 & 25.47$\pm$01.06 \\
BIOT & 71.37$\pm$02.19 & 72.43$\pm$01.94 & 69.52$\pm$00.23 & BIOT & 58.30$\pm$00.71 & 78.83$\pm$00.58 & 58.60$\pm$00.57 \\
LaBraM & 67.77$\pm$00.95 & 77.32$\pm$02.88 & 61.13$\pm$03.03 & LaBraM & 55.10$\pm$07.21 & 72.07$\pm$02.26 & 43.80$\pm$03.72 \\
EEGPT & 72.66$\pm$00.90 & 81.16$\pm$03.42 & 64.30$\pm$06.48 & EEGPT & \textcolor{red}{\textbf{76.87$\pm$00.37}} & \textcolor{red}{\textbf{87.63$\pm$01.09}} & \textcolor{red}{\textbf{76.43$\pm$01.52}} \\
CBraMod & 74.12$\pm$00.79 & \textcolor{red}{\textbf{83.89$\pm$00.93}} & \textcolor{red}{\textbf{72.58$\pm$01.55}} & CBraMod & 62.80$\pm$00.57 & \textcolor{blue}{\textbf{81.63$\pm$00.90}} & 62.80$\pm$01.93 \\
CSBrain & 71.27$\pm$00.75 & 79.63$\pm$01.20 & 55.10$\pm$02.21 & CSBrain & 67.33$\pm$01.42 & 74.53$\pm$00.31 & 45.48$\pm$01.25 \\
REVE & \textcolor{red}{\textbf{74.67$\pm$01.46}} & \textcolor{blue}{\textbf{82.32$\pm$01.47}} & 68.22$\pm$03.92 & REVE & 71.55$\pm$02.13 & 81.03$\pm$01.48 & 61.33$\pm$03.53 \\
NSP & \textcolor{blue}{\textbf{74.28$\pm$01.80}} & 80.20$\pm$01.72 & \textcolor{blue}{\textbf{71.62$\pm$01.73}} & NSP & \textcolor{blue}{\textbf{76.13$\pm$01.70}} & 79.40$\pm$01.37 & \textcolor{blue}{\textbf{72.87$\pm$01.97}} \\
\bottomrule
\end{tabular}
\rowcolors{2}{}{}
\renewcommand{\arraystretch}{1}
\end{table}

\subsubsection{Visual Target Detection}
Visual target detection evaluates stimulus-evoked EEG decoding. Things-EEG-2 measures whether the representation can retain short-latency visual information across a high-density 63-channel montage.

\paragraph{Things-EEG-2.} Things-EEG-2 evaluates binary visual-target decoding from 63-channel, 5-s windows, with 41,570 samples from 10 participants~\citep{gifford2022things}. NSP records $72.77\pm01.06$ BAC with single-task full fine-tuning, $62.50\pm01.00$ with multi-task full fine-tuning, $50.23\pm00.06$ with frozen transfer, and $60.13\pm00.50$ with LoRA. Single-task full fine-tuning performs best and exceeds the strongest baseline in that setting by 1.68 points.

\begin{table}[H]
\centering
\caption{Complete Things-EEG-2 results across four adaptation settings (\%, mean $\pm$ sample SD). \textcolor{red}{\textbf{Bold red}} and \textcolor{blue}{\textbf{bold blue}} mark the best and second-best mean for each metric and protocol.}
\label{tab:eegfm_things_eeg_2}
\vskip -0.04in
\fontsize{7.2}{7.8}\selectfont
\setlength{\tabcolsep}{3.2pt}
\renewcommand{\arraystretch}{1.05}
\rowcolors{3}{black!3}{white}
\begin{tabular}{@{}lccc@{\hspace{10pt}}lccc@{}}
\toprule
\multicolumn{4}{c}{\textbf{Full Fine-Tuning}} & \multicolumn{4}{c}{\textbf{Parameter-Efficient Multi-Task}} \\
\cmidrule(lr){1-4}\cmidrule(lr){5-8}
Model & \multicolumn{3}{c}{Single-task} & Model & \multicolumn{3}{c}{Frozen} \\
 & BAC & AUROC & AUCPR & & BAC & AUROC & AUCPR \\
\midrule
BENDR & \textcolor{blue}{\textbf{71.09$\pm$02.02}} & \textcolor{blue}{\textbf{83.09$\pm$00.69}} & \textcolor{blue}{\textbf{52.68$\pm$00.45}} & BENDR & 50.00$\pm$00.00 & 51.17$\pm$01.46 & 11.23$\pm$00.39 \\
BIOT & 51.10$\pm$00.30 & 62.05$\pm$01.53 & 18.05$\pm$01.06 & BIOT & 50.05$\pm$00.00 & 53.85$\pm$00.56 & 12.43$\pm$00.33 \\
LaBraM & 53.38$\pm$00.39 & 56.38$\pm$01.08 & 13.65$\pm$00.45 & LaBraM & 50.00$\pm$00.00 & 53.77$\pm$00.45 & 12.57$\pm$00.17 \\
EEGPT & 62.08$\pm$00.58 & 75.69$\pm$01.02 & 33.32$\pm$01.46 & EEGPT & 50.00$\pm$00.00 & \textcolor{red}{\textbf{59.57$\pm$00.42}} & \textcolor{blue}{\textbf{14.18$\pm$00.15}} \\
CBraMod & 62.40$\pm$01.56 & 77.61$\pm$00.32 & 37.48$\pm$00.78 & CBraMod & 50.00$\pm$00.00 & 53.85$\pm$00.07 & 12.29$\pm$00.13 \\
CSBrain & 57.20$\pm$00.29 & 62.67$\pm$00.53 & 21.53$\pm$00.42 & CSBrain & 50.00$\pm$00.00 & 51.63$\pm$02.31 & 12.10$\pm$00.94 \\
REVE & 61.03$\pm$00.90 & 66.77$\pm$00.52 & 22.87$\pm$01.72 & REVE & \textcolor{red}{\textbf{50.47$\pm$00.13}} & \textcolor{blue}{\textbf{58.18$\pm$00.06}} & \textcolor{red}{\textbf{14.92$\pm$00.06}} \\
NSP & \textcolor{red}{\textbf{72.77$\pm$01.06}} & \textcolor{red}{\textbf{89.53$\pm$01.72}} & \textcolor{red}{\textbf{55.67$\pm$00.06}} & NSP & \textcolor{blue}{\textbf{50.23$\pm$00.06}} & 57.47$\pm$01.27 & 13.57$\pm$00.40 \\
\midrule
Model & \multicolumn{3}{c}{Multi-task} & Model & \multicolumn{3}{c}{LoRA} \\
 & BAC & AUROC & AUCPR & & BAC & AUROC & AUCPR \\
\midrule
BENDR & \textcolor{red}{\textbf{63.62$\pm$01.38}} & \textcolor{red}{\textbf{77.23$\pm$00.88}} & \textcolor{red}{\textbf{42.29$\pm$01.75}} & BENDR & 50.00$\pm$00.00 & 50.27$\pm$00.12 & 11.03$\pm$00.05 \\
BIOT & 54.15$\pm$00.25 & 63.01$\pm$00.52 & 20.12$\pm$00.33 & BIOT & 50.67$\pm$00.05 & \textcolor{blue}{\textbf{64.13$\pm$00.17}} & \textcolor{blue}{\textbf{19.47$\pm$00.05}} \\
LaBraM & 56.47$\pm$00.32 & 62.82$\pm$00.84 & 22.23$\pm$00.64 & LaBraM & 50.00$\pm$00.00 & 55.27$\pm$02.33 & 13.13$\pm$01.13 \\
EEGPT & 51.54$\pm$00.76 & 66.64$\pm$00.04 & 20.30$\pm$00.65 & EEGPT & 51.70$\pm$00.37 & 61.83$\pm$00.21 & 14.97$\pm$00.31 \\
CBraMod & 59.13$\pm$00.72 & 69.58$\pm$01.41 & 31.85$\pm$01.21 & CBraMod & 50.23$\pm$00.05 & 57.60$\pm$00.79 & 14.17$\pm$00.31 \\
CSBrain & 57.49$\pm$00.27 & 62.47$\pm$01.27 & 24.66$\pm$01.93 & CSBrain & 50.00$\pm$00.00 & 54.43$\pm$00.62 & 13.03$\pm$00.21 \\
REVE & 59.43$\pm$01.20 & 69.00$\pm$00.53 & 32.07$\pm$01.85 & REVE & \textcolor{blue}{\textbf{53.77$\pm$00.12}} & 60.70$\pm$00.24 & 15.70$\pm$00.22 \\
NSP & \textcolor{blue}{\textbf{62.50$\pm$01.00}} & \textcolor{blue}{\textbf{76.68$\pm$00.61}} & \textcolor{blue}{\textbf{41.80$\pm$01.69}} & NSP & \textcolor{red}{\textbf{60.13$\pm$00.50}} & \textcolor{red}{\textbf{77.17$\pm$01.05}} & \textcolor{red}{\textbf{41.33$\pm$01.40}} \\
\bottomrule
\end{tabular}
\rowcolors{2}{}{}
\renewcommand{\arraystretch}{1}
\end{table}

\subsubsection{Neurodegenerative Disease Identification}
Neurodegenerative disease identification tests clinically relevant subject-level discrimination. ADFTD separates Alzheimer's disease, frontotemporal dementia, and healthy controls, requiring the representation to capture disease-related differences beyond transient event structure.

\paragraph{ADFTD.} ADFTD is a three-class diagnostic task distinguishing Alzheimer's disease, frontotemporal dementia, and healthy controls from 19-channel, 10-s windows; it contains 7,013 samples from 88 participants~\citep{miltiadous2023adftd}. NSP reaches $52.17\pm00.24$ BAC with single-task full fine-tuning, $56.66\pm01.66$ with multi-task full fine-tuning, $54.10\pm00.95$ with frozen transfer, and $50.83\pm02.12$ with LoRA. Multi-task full fine-tuning is strongest and leads the best listed baseline under that setting by 0.95 points.

\begin{table}[H]
\centering
\caption{Complete ADFTD results across four adaptation settings (\%, mean $\pm$ sample SD). \textcolor{red}{\textbf{Bold red}} and \textcolor{blue}{\textbf{bold blue}} mark the best and second-best mean for each metric and protocol.}
\label{tab:eegfm_adftd}
\vskip -0.04in
\fontsize{7.2}{7.8}\selectfont
\setlength{\tabcolsep}{3.2pt}
\renewcommand{\arraystretch}{1.05}
\rowcolors{3}{black!3}{white}
\begin{tabular}{@{}lccc@{\hspace{10pt}}lccc@{}}
\toprule
\multicolumn{4}{c}{\textbf{Full Fine-Tuning}} & \multicolumn{4}{c}{\textbf{Parameter-Efficient Multi-Task}} \\
\cmidrule(lr){1-4}\cmidrule(lr){5-8}
Model & \multicolumn{3}{c}{Single-task} & Model & \multicolumn{3}{c}{Frozen} \\
 & BAC & F1 & $\kappa$ & & BAC & F1 & $\kappa$ \\
\midrule
BENDR & 37.16$\pm$02.62 & 40.38$\pm$01.69 & 07.27$\pm$04.67 & BENDR & 33.30$\pm$00.00 & 22.17$\pm$00.75 & 00.00$\pm$00.00 \\
BIOT & 47.43$\pm$01.03 & 48.17$\pm$01.38 & 23.77$\pm$01.52 & BIOT & \textcolor{blue}{\textbf{51.37$\pm$00.09}} & \textcolor{blue}{\textbf{50.47$\pm$00.05}} & \textcolor{blue}{\textbf{29.47$\pm$00.21}} \\
LaBraM & 35.75$\pm$02.76 & 34.47$\pm$03.44 & 11.40$\pm$03.67 & LaBraM & 38.20$\pm$00.77 & 38.40$\pm$01.05 & 09.70$\pm$01.27 \\
EEGPT & 46.50$\pm$02.28 & 48.77$\pm$02.56 & 21.81$\pm$03.72 & EEGPT & 39.03$\pm$02.48 & 32.02$\pm$00.53 & 07.72$\pm$03.37 \\
CBraMod & 48.47$\pm$01.12 & \textcolor{blue}{\textbf{51.68$\pm$01.11}} & \textcolor{red}{\textbf{31.17$\pm$01.74}} & CBraMod & 38.17$\pm$00.13 & 36.17$\pm$00.35 & 09.12$\pm$00.26 \\
CSBrain & 46.08$\pm$02.61 & 46.63$\pm$03.41 & 24.27$\pm$03.85 & CSBrain & 41.30$\pm$00.28 & 42.23$\pm$00.25 & 12.40$\pm$00.45 \\
REVE & \textcolor{blue}{\textbf{50.22$\pm$00.41}} & 50.72$\pm$01.85 & \textcolor{blue}{\textbf{29.21$\pm$02.08}} & REVE & 46.10$\pm$00.24 & 49.70$\pm$00.14 & 23.20$\pm$00.29 \\
NSP & \textcolor{red}{\textbf{52.17$\pm$00.24}} & \textcolor{red}{\textbf{52.67$\pm$01.97}} & 27.87$\pm$03.25 & NSP & \textcolor{red}{\textbf{54.10$\pm$00.95}} & \textcolor{red}{\textbf{57.53$\pm$02.08}} & \textcolor{red}{\textbf{37.57$\pm$00.55}} \\
\midrule
Model & \multicolumn{3}{c}{Multi-task} & Model & \multicolumn{3}{c}{LoRA} \\
 & BAC & F1 & $\kappa$ & & BAC & F1 & $\kappa$ \\
\midrule
BENDR & \textcolor{blue}{\textbf{55.71$\pm$03.76}} & \textcolor{blue}{\textbf{56.97$\pm$02.75}} & \textcolor{blue}{\textbf{33.98$\pm$04.63}} & BENDR & 33.87$\pm$00.80 & 18.77$\pm$07.78 & 00.83$\pm$01.18 \\
BIOT & 52.40$\pm$01.98 & 53.48$\pm$02.71 & \textcolor{red}{\textbf{36.27$\pm$03.65}} & BIOT & 48.77$\pm$01.40 & 51.33$\pm$01.88 & 25.27$\pm$01.93 \\
LaBraM & 50.27$\pm$01.74 & 52.36$\pm$02.23 & 26.17$\pm$03.15 & LaBraM & 43.07$\pm$00.66 & 45.77$\pm$00.76 & 18.40$\pm$01.28 \\
EEGPT & 52.91$\pm$01.57 & 54.82$\pm$01.30 & 31.14$\pm$01.85 & EEGPT & 44.93$\pm$01.18 & 45.80$\pm$01.31 & 17.73$\pm$01.70 \\
CBraMod & 51.95$\pm$02.84 & 54.41$\pm$03.13 & 30.06$\pm$04.35 & CBraMod & 48.83$\pm$00.50 & \textcolor{blue}{\textbf{51.57$\pm$00.25}} & 26.27$\pm$01.86 \\
CSBrain & 48.96$\pm$02.65 & 50.17$\pm$03.76 & 24.96$\pm$04.13 & CSBrain & 48.60$\pm$00.45 & 47.20$\pm$00.57 & 27.33$\pm$00.74 \\
REVE & 52.12$\pm$03.78 & 51.10$\pm$03.90 & 30.70$\pm$06.41 & REVE & \textcolor{blue}{\textbf{50.73$\pm$01.28}} & 24.73$\pm$02.80 & \textcolor{red}{\textbf{56.03$\pm$00.40}} \\
NSP & \textcolor{red}{\textbf{56.66$\pm$01.66}} & \textcolor{red}{\textbf{60.46$\pm$01.88}} & 32.00$\pm$03.66 & NSP & \textcolor{red}{\textbf{50.83$\pm$02.12}} & \textcolor{red}{\textbf{54.40$\pm$02.10}} & \textcolor{blue}{\textbf{31.67$\pm$00.63}} \\
\bottomrule
\end{tabular}
\rowcolors{2}{}{}
\renewcommand{\arraystretch}{1}
\end{table}